\documentclass{CVM}
\pdfoutput=1

\usepackage{tikz}
\usepackage[edges]{forest}
\definecolor{hidden-draw}{RGB}{20,68,106}
\definecolor{hidden-pink}{RGB}{255,245,247}

\definecolor{fig1blue}{RGB}{47,85,151}
\definecolor{fig1green}{RGB}{84,130,53}
\definecolor{fig1orange}{RGB}{197,90,17}

\usepackage{amssymb}
\usepackage{placeins}
\usepackage{mathrsfs}

\makeatletter
\DeclareRobustCommand\onedot{\futurelet\@let@token\@onedot}
\def\@onedot{\ifx\@let@token.\else.\null\fi\xspace}

\def\eg{\emph{e.g}\onedot} 
\def\ie{\emph{i.e}\onedot} 
 
\def\etc{\emph{etc}\onedot} 
 
\def\etal{\emph{et al}\onedot}

\makeatother

\CVMsetup{
type      = {Review Article},
doi       = {CVM.XXXX},
title     = {Personalized Image Generation with Deep Generative Models: A Decade Survey},
author    = {Yuxiang Wei$^{1, 2}$, \ Yiheng Zheng$^{1}$, \ Yabo Zhang$^{1}$, \ Ming Liu$^{1}$ \cor{}, \ Zhilong Ji$^{3}$, \ Lei Zhang$^{2}$, and Wangmeng Zuo$^{1}$ \\ },
runauthor = {Y. Wei, Y. Zheng, Y. Zhang, et al.},
abstract  = {
Recent advancements in generative models have significantly facilitated the development of personalized content creation.
Given a small set of images with user-specific concept, \emph{personalized image generation} allows to create images that incorporate the specified concept and adhere to provided text descriptions. 
Due to its wide applications in content creation, significant effort has been devoted to this field in recent years.
Nonetheless, the technologies used for personalization have evolved alongside the development of generative models, with their distinct and interrelated components.
In this survey, we present a comprehensive review of generalized personalized image generation across various generative models, including traditional GANs, contemporary text-to-image diffusion models, and emerging multi-model autoregressive (AR) models. 
We first define a unified framework that standardizes the personalization process across different generative models, encompassing three key components, \ie, inversion spaces, inversion methods, and personalization schemes.
This unified framework offers a structured approach to dissecting and comparing personalization techniques across different generative architectures. 
Building upon this unified framework, we further provide an in-depth analysis of personalization techniques within each generative model, highlighting their unique contributions and innovations.
Through comparative analysis, this survey elucidates the current landscape of personalized image generation, identifying commonalities and distinguishing features among existing methods.
Finally, we discuss the open challenges in the field and propose potential directions for future research.
We keep tracing related works at \url{https://github.com/csyxwei/Awesome-Personalized-Image-Generation}.
},
keywords  = {Personalized Image Generation, Generative Models, Generative Adversarial Networks, Text-to-Image Diffusion Models, Multi-modal AutoRegressive Models},
copyright = {The Author(s)},
}

\begin{document}

\maketitle

\enlargethispage{-3pt}
\begin{figure}[b] \vskip -4mm
\small\renewcommand\arraystretch{1.3}
\begin{tabular}{p{80.5mm}} \toprule\\ \end{tabular}
\vskip -4.5mm \noindent \setlength{\tabcolsep}{1pt}
\begin{tabular}{p{3.5mm}p{80mm}}
$1\quad $ & Faculty of Computing, Harbin Institute of Technology, Harbin 150001, China. E-mail: yuxiang.wei.cs@gmail.com;  zhengyihengCV@outlook.com; hitzhangyabo2017@gmail.com; csmliu@outlook.com\cor{}; wmzuo@hit.edu.cn\\
$2\quad $ & Department of Computing, The Hong Kong Polytechnic University, Hong Kong, China. E-mail: cslzhang@comp.polyu.edu.hk\\
$3\quad $ & Tomorrow Advancing Life, Beijing 100081, China. E-mail: jizhilong@tal.com \\
\end{tabular} \vspace {-3mm}
\end{figure}

\section{Introduction}

\begin{figure*}[t]
\centering
\includegraphics[width=1\linewidth]{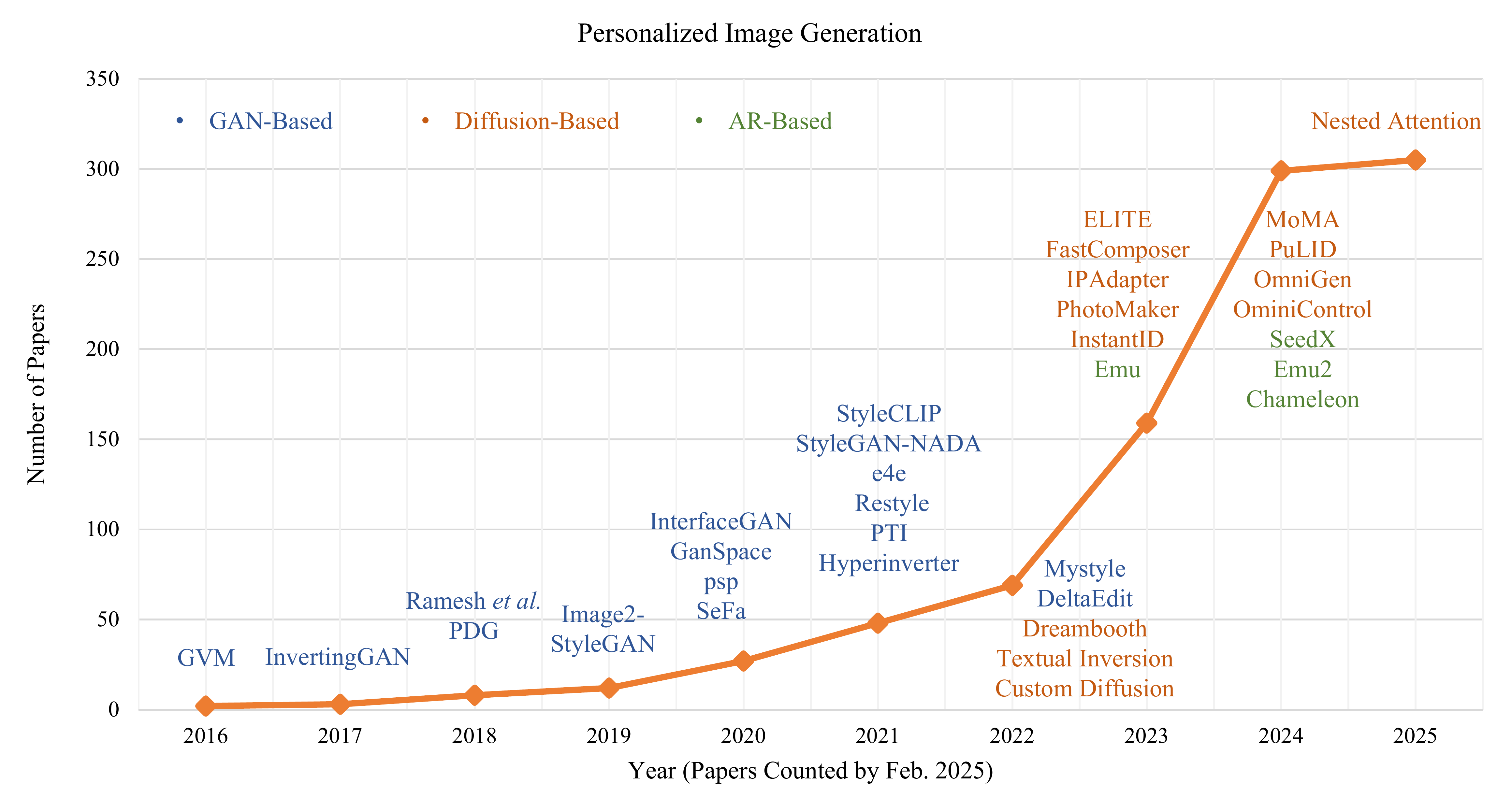}
\vspace{-2em}
\caption{\textbf{The rough number of papers on personalized image generation with deep generative models.} Representative works on the personalization task over time are shown. The GAN-based methods, diffusion-based methods, and autoregressive-based methods are highlighted in {\color{fig1blue} blue}, {\color{fig1orange} orange}, {\color{fig1green} green}, respectively}
\label{fig:paper_number}
\vspace{-1em}
\end{figure*}

In recent years, generative models have undergone rapid evolution, advancing from Generative Adversarial Networks (GANs)~\cite{GAN_goodfellow2020generative} to Diffusion Models (DMs)~\cite{DDPM_ho2020denoising} and Autoregressive (AR) Models~\cite{Emu_sun2023emu}.
These models have demonstrated superiority in generating diverse and high-quality images.
More recently, text-to-image (T2I) generation models~\cite{LDM_rombach2022high,SDXL_podell2023sdxl,Imagen_saharia2022photorealistic,DALLE2_ramesh2022hierarchical} have showcased exceptional flexibility in controlling image generation through textual inputs.
Benefiting from large-scale pretraining, these T2I models exhibit remarkable semantic understanding, enabling them to create photorealistic images that accurately reflect the given textual prompts.
These advancements have facilitated various downstream tasks, such as conditional generation~\cite{Controlnet_zhang2023adding}, image editing~\cite{prompt2prompt_hertz2022prompt,Ledits++_brack2024ledits++,Masactrl_cao2023masactrl}, and art creation~\cite{photomaker_li2024photomaker,Instantstyle_wang2024instantstyle}.
Among them, personalized image generation~\cite{TI_gal2022image,Dreambooth_ruiz2023dreambooth,ELITE_wei2023elite,IPAdapter_ye2023ip} has attracted significant attention, which focuses on creating user-specific concepts through image generation.

Contemporary personalization approaches~\cite{TI_gal2022image,Dreambooth_ruiz2023dreambooth,CD_kumari2023multi,ELITE_wei2023elite,IPAdapter_ye2023ip,photomaker_li2024photomaker} primarily leverage text-to-image diffusion models, enhancing them to generate user-specific concepts within specified context.
Specifically, the user-specific concept is indicated by a small set of images containing the target concept (\eg, subjects, faces, or styles, typically 3 $\sim$ 5 images), while the specified context is provided by the target text.
In this survey, we explore the generalized personalized image generation techniques across various generative models, including traditional GANs, current text-to-image diffusion models, and emerging multi-model autoregressive models.
For example, as illustrated in Fig.~\ref{fig:personalization_methods}, GAN Inversion~\cite{GVM_zhu2016generative,psp_richardson2021encoding,e4e_tov2021designing,liu2023survey,xia2022gan} maps real images into a GAN's latent space, allowing for subsequent manipulations to achieve generalized personalization.
These GAN Inversion techniques have significantly inspired the development of concept inversion techniques in current diffusion-based personalization methods~\cite{E4T_gal2023encoder,ELITE_wei2023elite,DAT_arar2023domain}.
Furthermore, recent advancements of autoregressive models~\cite{DALLE_ramesh2021zero,Cogview_ding2021cogview,Emu2_sun2024generative,Emu_sun2023emu} in multi-modal generation have introduced promising new avenues for personalization. 
Therefore, in this paper, we consider these techniques collectively as generalized personalized image generation and provide a comprehensive survey of personalization utilizing these generative models.
Fig.~\ref{fig:paper_number} illustrates the number of papers and representative works in this field in recent years.
Over the past two years, more than 180 methods have been proposed for diffusion-based personalization, and more than 300 methods have been developed across various generative models over the past decade.

Several surveys~\cite{cao2024controllable,zhang2024text,zhan2024conditional,shuai2024survey} have provided comprehensive reviews of state-of-the-art diffusion-based methods for conditional image synthesis. 
While these works offer valuable insights, they predominantly focus on general conditional generation techniques and do not extensively explore the field of personalized image generation.
Among them, two recent surveys~\cite{shuai2024survey, zhang2024survey} are particularly relevant to our work.
Shuai~\etal~\cite{shuai2024survey} summarizes the application of text-to-image diffusion models in image editing, categorizing personalized image generation as a form of content-free editing.
Zhang~\etal~\cite{zhang2024survey} surveys personalized image generation with diffusion models, but tends to overlook advancements introduced by other generative models.
In contrast to these existing surveys, our study focuses on personalization across various generative models, including GANs, text-to-image diffusion models, and multi-modal autoregressive models. 
We provide a comprehensive overview of the personalization techniques within these models, highlighting their commonalities and differences to clarify the current landscape of personalized image generation methods.

To systematically explore personalized image generation, in this survey, we first define a unified framework that standardizes the personalization process among different generative models.  
Specifically, we categorize personalized image generation into two main stages: \textit{concept inversion} and \textit{personalized generation}, which consist of three key components.
\textbf{Inversion Spaces}: The personalization process begins by inverting a given concept into a representation that the generative model can manipulate, with various spaces explored for this purpose.
\textbf{Inversion Methods}: Once the target space is selected, several inversion methods can be employed to learn the representation, such as optimization-based approaches, learning-based approaches, \etc.
\textbf{Personalization Schemes}: The generative model then integrates the inverted concept representation to produce the personalized images. 
This step involves various personalization methods and concept categories tailored to each generative model.
Building upon this unified framework, we provide an in-depth analysis of personalization techniques within each generative model, highlighting both commonalities and distinctions across various scenarios.
Additionally, we introduce the evaluation metrics and datasets commonly used in personalized image generation, as well as discuss the open challenges and potential directions for future research.


The rest of this paper is organized as follows. 
Section~\ref{sec:definition} defines the problem of personalized image generation and introduces the preliminaries of generative models.
Sections~\ref{sec:gans}$\sim$\ref{sec:ars} discusses the personalization techniques specific to different generative models, including GANs, diffusion models, and AR models.
Section~\ref{sec:evaluation} reviews the existing evaluation datasets and metrics used in personalized image generation.
Section~\ref{sec:challenge} identifies the open challenges and outlines potential future research directions.
Finally, Section~\ref{sec:conclusion} concludes the survey by summarizing key insights and contributions.
Fig.~\ref{fig:taxonomy} illustrates the organization of our survey and categorizes reviewed papers in each section.

\definecolor{hidden-draw}{RGB}{161,165,193}
\definecolor{hidden-pink}{RGB}{244,243,250}
\tikzstyle{my-box}=[
    rectangle,
    draw=hidden-draw,
    rounded corners,
    text opacity=1,
    minimum height=1.5em,
    minimum width=5em,
    inner sep=2pt,
    align=center,
    fill opacity=.5,
    line width=0.8pt,
]
\tikzstyle{leaf}=[my-box, minimum height=1.5em,
    fill=hidden-pink!80, text=black, align=left,font=\tiny,
    inner xsep=2pt,
    inner ysep=4pt,
    line width=0.8pt,
]
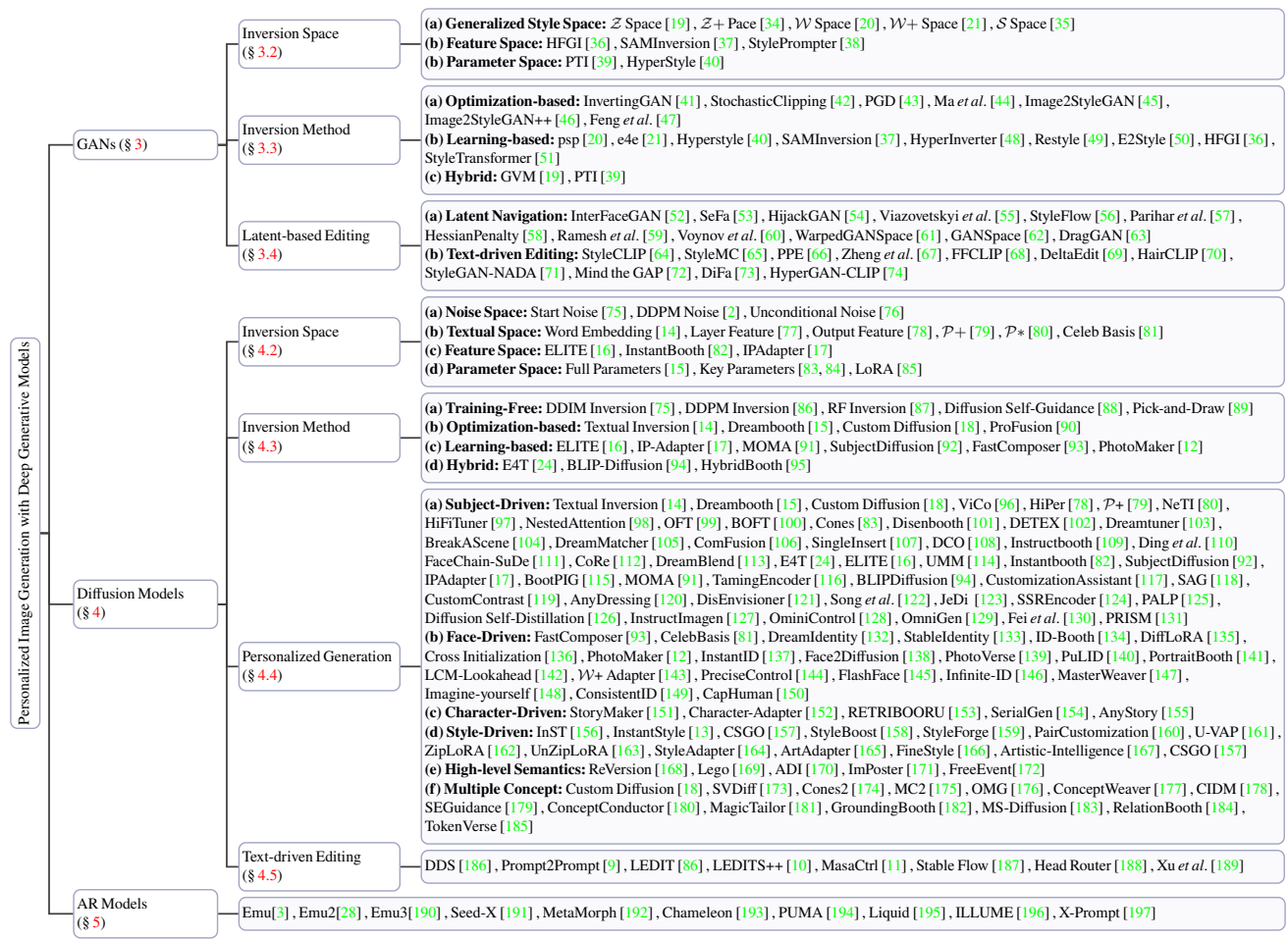
\begin{figure*}[t!]
    \centering
    \resizebox{1.0\textwidth}{!}{
        \begin{forest}
            forked edges,
            for tree={
                grow=east,
                reversed=true,
                anchor=base west,
                parent anchor=east,
                child anchor=west,
                base=left,
                font=\tiny,
                rectangle,
                draw=hidden-draw,
                rounded corners,
                align=left,
                minimum width=2em,
                edge+={darkgray, line width=1pt},
                s sep=3pt,
                inner xsep=2pt,
                inner ysep=3pt,
                line width=0.8pt,
                ver/.style={rotate=90, child anchor=north, parent anchor=south, anchor=center},
            },
            where level=1{text width=8em,font=\tiny,}{},
            where level=2{text width=9em,font=\tiny,}{},
            where level=3{text width=10em,font=\tiny,}{},
            [
                Personalized Image Generation with Deep Generative Models, ver
                [
                    GANs (\S~\ref{sec:gans})
                    [
                        Inversion Space \\ (\S~\ref{sec:gan_inversion_space})
                        [
                            \textbf{(a) Generalized Style Space:} $\mathcal{Z}$ Space~\cite{GVM_zhu2016generative} {,} 
                            $\mathcal{Z}+$ Pace~\cite{katsumata2023revisiting} {,}
                            $\mathcal{W} $ Space~\cite{psp_richardson2021encoding} {,} 
                            $\mathcal{W} + $ Space~\cite{e4e_tov2021designing} {,} $\mathcal{S} $ Space~\cite{Stylespace_wu2021stylespace} {} \\  
                            \textbf{(b) Feature Space:} HFGI~\cite{HFGI_wang2022high} {,} SAMInversion~\cite{SAMInversion_parmar2022spatially} {,} StylePrompter~\cite{StylePrompter_zhuang2023styleprompter} \\  
                            \textbf{(b) Parameter Space:} PTI~\cite{PTI_roich2022pivotal} {,} HyperStyle~\cite{Hyperstyle_alaluf2022hyperstyle} \\  
                            , leaf, text width=50em
                        ]                    
				]
                    [
                        Inversion Method \\ (\S~\ref{sec:gan_inversion_methods})
                        [
                            \textbf{(a) Optimization-based:} InvertingGAN~\cite{InvertingGAN_creswell2018inverting} {,} StochasticClipping~\cite{StochasticClipping_lipton2017precise} {,} PGD~\cite{PGD_shah2018solving} {,} Ma~\etal~\cite{ma2018invertibility} {,} Image2StyleGAN~\cite{Image2stylegan_abdal2019image2stylegan} {,} \\ Image2StyleGAN++~\cite{Image2stylegan++_abdal2020image2stylegan++} {,} Feng~\etal~\cite{feng2022near} {}  \\  
                            \textbf{(b) Learning-based:} psp~\cite{psp_richardson2021encoding} {,} e4e~\cite{e4e_tov2021designing} {,} Hyperstyle~\cite{Hyperstyle_alaluf2022hyperstyle} {,} SAMInversion~\cite{SAMInversion_parmar2022spatially} {,} HyperInverter~\cite{Hyperinverter_dinh2022hyperinverter} {,} Restyle~\cite{Restyle_alaluf2021restyle} {,} E2Style~\cite{E2Style_wei2022e2style} {,} HFGI~\cite{HFGI_wang2022high} {,} \\ StyleTransformer~\cite{StyleTransformer_hu2022style} \\  
                            \textbf{(c) Hybrid:} GVM~\cite{GVM_zhu2016generative} {,} PTI~\cite{PTI_roich2022pivotal} {} \\  
                            , leaf, text width=50em
                        ]                    
                    ]
                    [
                        Latent-based Editing \\ (\S~\ref{sec:gan_personalization})
                        [
                            \textbf{(a) Latent Navigation:} InterFaceGAN~\cite{Interfacegan_shen2020interfacegan} {,}  SeFa~\cite{SeFa_shen2021closed} {,} HijackGAN~\cite{HijackGAN_wang2021hijack} {,} Viazovetskyi~\etal~\cite{Stylegan2Distillation_viazovetskyi2020stylegan2} {,} StyleFlow~\cite{StyleFlow_abdal2021styleflow} {,} Parihar~\etal~\cite{parihar2023exploring} {,} \\ HessianPenalty~\cite{HessianPenalty_peebles2020hessian} {,} Ramesh~\etal~\cite{SpectralRegularizer_ramesh2018spectral}  {,} Voynov~\etal~\cite{voynov2020unsupervised} {,} WarpedGANSpace~\cite{WarpedGANSpace_tzelepis2021warpedganspace} {,} GANSpace~\cite{Ganspace_harkonen2020ganspace} {,} DragGAN~\cite{DragGAN_pan2023drag} {}  \\  
                            \textbf{(b) Text-driven Editing:} StyleCLIP~\cite{Styleclip_patashnik2021styleclip} {,}  StyleMC~\cite{Stylemc_kocasari2022stylemc} {,} PPE~\cite{PPE_xu2022predict} {,} Zheng~\etal~\cite{zheng2022bridging} {,} FFCLIP~\cite{FFCLIP_zhu2022one} {,} DeltaEdit~\cite{Deltaedit_lyu2023deltaedit} {,} HairCLIP~\cite{Hairclip_wei2022hairclip} {,} \\ StyleGAN-NADA~\cite{StyleganNada_gal2022stylegan} {,} Mind the GAP~\cite{MindTheGap_zhu2021mind} {,} DiFa~\cite{DiFa_zhang2022towards} {,} HyperGAN-CLIP~\cite{HyperGANCLIP_anees2024hypergan} \\  
                            , leaf, text width=50em
                        ]                    
                    ]
			]	
                [
                    Diffusion Models \\ (\S~\ref{sec:diffusion})
                    [
                        Inversion Space \\ (\S~\ref{sec:diffusion_inversion_space})
                        [
                            \textbf{(a) Noise Space:} Start Noise~\cite{DDIM_song2020denoising} {,}  DDPM Noise~\cite{DDPM_ho2020denoising} {,} Unconditional Noise~\cite{NullInversion_mokady2023null} \\  
                            \textbf{(b) Textual Space:} Word Embedding~\cite{TI_gal2022image} {,} Layer Feature~\cite{Catversion_zhao2023catversion} {,} Output Feature~\cite{HiPer_han2023highly} {,} $\mathcal{P}+$~\cite{P+_voynov2023p+} {,} $\mathcal{P}*$~\cite{NeTI_alaluf2023neural} {,} Celeb Basis~\cite{CelebBasis_yuan2023inserting} \\  
                            \textbf{(c) Feature Space:} ELITE~\cite{ELITE_wei2023elite} {,} InstantBooth~\cite{Instantbooth_shi2024instantbooth} {,}  IPAdapter~\cite{IPAdapter_ye2023ip} \\  
                            \textbf{(d) Parameter Space:} Full Parameters~\cite{Dreambooth_ruiz2023dreambooth}  {,} Key Parameters~\cite{Cones_liu2023cones,Perfusion_tewel2023key} {,}  LoRA~\cite{Lora_hu2021lora} \\  
                            , leaf, text width=50em
                        ]
                    ]
                    [
                        Inversion Method \\ (\S~\ref{sec:diffusion_inversion_methods})
                        [
                            \textbf{(a) Training-Free:} DDIM Inversion~\cite{DDIM_song2020denoising} {,}  DDPM Inversion~\cite{Ledits_tsaban2023ledits} {,} RF Inversion~\cite{rout2024semantic} {,} Diffusion Self-Guidance~\cite{DSG_epstein2023diffusion} {,} Pick-and-Draw~\cite{PickandDraw_lv2024pick} {} \\  
                            \textbf{(b) Optimization-based:} Textual Inversion~\cite{TI_gal2022image} {,}  Dreambooth~\cite{Dreambooth_ruiz2023dreambooth} {,} Custom Diffusion~\cite{CD_kumari2023multi} {,} ProFusion~\cite{ProFusion_zhou2023enhancing}{} \\
                            \textbf{(c) Learning-based:} ELITE~\cite{ELITE_wei2023elite} {,} IP-Adapter~\cite{IPAdapter_ye2023ip} {,} MOMA~\cite{MOMA_song2025moma} {,} SubjectDiffusion~\cite{SubjectDiffusion_ma2024subject} {,} FastComposer~\cite{Fastcomposer_xiao2024fastcomposer} {,} PhotoMaker~\cite{photomaker_li2024photomaker} \\  
                            \textbf{(d) Hybrid:} E4T~\cite{E4T_gal2023encoder} {,} BLIP-Diffusion~\cite{BLIPDiffusion_li2024blip} {,} HybridBooth~\cite{HybridBooth_guan2025hybridbooth} {}  \\  
                            , leaf, text width=50em
                        ]
                    ]
                    [
                        Personalized Generation \\ (\S~\ref{sec:diffusion_personalization})
                        [
                            \textbf{(a) Subject-Driven:} Textual Inversion~\cite{TI_gal2022image} {,}  Dreambooth~\cite{Dreambooth_ruiz2023dreambooth} {,} Custom Diffusion~\cite{CD_kumari2023multi} {,} ViCo~\cite{ViCo_hao2023vico} {,} HiPer~\cite{HiPer_han2023highly} {,} $\mathcal{P}$+~\cite{P+_voynov2023p+} {,} NeTI~\cite{NeTI_alaluf2023neural} {,}  \\ HiFiTuner~\cite{HiFiTuner_wang2023hifi} {,} NestedAttention~\cite{NestedAttention_patashnik2025nested} {,} OFT~\cite{OFT_qiu2023controlling} {,} BOFT~\cite{BOFT_liu2023parameter} {,} Cones~\cite{Cones_liu2023cones} {,} Disenbooth~\cite{Disenbooth_chen2023disenbooth} {,} DETEX~\cite{DETEX_cai2024decoupled} {,} Dreamtuner~\cite{Dreamtuner_hua2023dreamtuner} {,} \\ BreakAScene~\cite{BreakAScene_avrahami2023break} {,} DreamMatcher~\cite{Dreammatcher_nam2024dreammatcher} {,} ComFusion~\cite{ComFusion_hong2024comfusion} {,} SingleInsert~\cite{Singleinsert_wu2023singleinsert} {,} DCO~\cite{DCO_lee2024direct} {,} Instructbooth~\cite{Instructbooth_chae2023instructbooth} {,}  Ding~\etal~\cite{CLIPconverter_ding2024clip}  \\ FaceChain-SuDe~\cite{FaceChainSuDe_qiao2024facechain} {,} CoRe~\cite{CoRe_wu2024core} {,} DreamBlend~\cite{DreamBlend_ram2025dreamblend} {,} E4T~\cite{E4T_gal2023encoder} {,} ELITE~\cite{ELITE_wei2023elite} {,} UMM~\cite{UMM_ma2023unified} {,} Instantbooth~\cite{Instantbooth_shi2024instantbooth} {,} SubjectDiffusion~\cite{SubjectDiffusion_ma2024subject} {,} \\ IPAdapter~\cite{IPAdapter_ye2023ip} {,}  BootPIG~\cite{Bootpig_purushwalkam2024bootpig} {,} MOMA~\cite{MOMA_song2025moma} {,} TamingEncoder~\cite{Taming_jia2023taming} {,} BLIPDiffusion~\cite{BLIPDiffusion_li2024blip} {,} CustomizationAssistant~\cite{CustomizationAssistant_zhou2024customization} {,} SAG~\cite{SAG_chan2024improving} {,} \\ CustomContrast~\cite{CustomContrast_chen2024customcontrast} {,} AnyDressing~\cite{AnyDressing_li2024anydressing} {,} DisEnvisioner~\cite{DisEnvisioner_he2024disenvisioner} {,} Song~\etal~\cite{song2024harmonizing} {,}  JeDi~ \cite{JeDi_zeng2024jedi} {,} SSREncoder~\cite{SSREncoder_zhang2024ssr} {,} PALP~\cite{PALP_arar2024palp} {,} \\ Diffusion Self-Distillation~\cite{cai2024diffusion} {,} InstructImagen~\cite{InstructImagen_hu2024instruct} {,} OminiControl~\cite{OminiControl_tan2024ominicontrol} {,} OmniGen~\cite{Omnigen_xiao2024omnigen} {,} Fei~\etal~\cite{GFTI_fei2023gradient} {,} PRISM~\cite{PRISM_he2024automated} {} \\  
                            \textbf{(b) Face-Driven:} FastComposer~\cite{Fastcomposer_xiao2024fastcomposer} {,}  CelebBasis~\cite{CelebBasis_yuan2023inserting} {,} DreamIdentity~\cite{Dreamidentity_chen2023dreamidentity} {,} StableIdentity~\cite{Stableidentity_wang2024stableidentity} {,} ID-Booth~\cite{ID-Booth_tomavsevicid} {,} DiffLoRA~\cite{Difflora_wu2024difflora} {,} \\ Cross Initialization~\cite{CrossInitialization_pang2024cross} {,} PhotoMaker~\cite{photomaker_li2024photomaker} {,} InstantID~\cite{Instantid_wang2024instantid} {,} Face2Diffusion~\cite{Face2Diffusion_shiohara2024face2diffusion} {,}  PhotoVerse~\cite{Photoverse_chen2023photoverse} {,} PuLID~\cite{Pulid_guo2024pulid} {,} PortraitBooth~\cite{Portraitbooth_peng2024portraitbooth} {,} \\ LCM-Lookahead~\cite{Lcm-lookahead_gal2024lcm} {,} $\mathcal{W}$+ Adapter~\cite{wplus_li2024stylegan} {,} PreciseControl~\cite{Precisecontrol_parihar2025precisecontrol} {,}  FlashFace~\cite{FlashFace_zhang2024flashface} {,} Infinite-ID~\cite{Infinite-ID_wu2025infinite} {,} MasterWeaver~\cite{MasterWeaver_wei2025masterweaver} {,} \\ Imagine-yourself~\cite{Imagine-yourself_he2024imagine} {,} ConsistentID~\cite{Consistentid_huang2024consistentid} {,} CapHuman~\cite{Caphuman_liang2024caphuman}  \\ 
                            \textbf{(c) Character-Driven:} StoryMaker~\cite{StoryMaker_zhou2024storymaker} {,} Character-Adapter~\cite{Character-Adapter_ma2024character} {,} RETRIBOORU~\cite{RETRIBOORU_tang2023retrieving} {,} SerialGen~\cite{SerialGen_xie2024serialgen} {,} AnyStory~\cite{AnyStory_he2025anystory} \\ 
                            \textbf{(d) Style-Driven:} InST~\cite{InST_zhang2023inversion} {,} InstantStyle~\cite{Instantstyle_wang2024instantstyle} {,} CSGO~\cite{CSGO_imagecsgo} {,} StyleBoost~\cite{StyleBoost_park2023styleboost} {,} StyleForge~\cite{StyleForge_park2024text} {,}  PairCustomization~\cite{PairCustomization_jones2024customizing} {,} U-VAP~\cite{U-VAP_wu2024u} {,} \\ ZipLoRA~\cite{Ziplora_shah2025ziplora} {,} UnZipLoRA~\cite{UnZipLoRA_liu2024unziplora} {,} StyleAdapter~\cite{StyleAdapter_wang2024styleadapter} {,} ArtAdapter~\cite{ArtAdapter_chen2024artadapter} {,} FineStyle~\cite{FineStyle_zhangfinestyle} {,} Artistic-Intelligence~\cite{Artistic-Intelligence_yang2024artistic} {,} CSGO~\cite{CSGO_imagecsgo} \\ 
                            \textbf{(e) High-level Semantics:} ReVersion~\cite{ReVersion_huang2024reversion} {,}  Lego~\cite{Lego_motamed2023lego} {,} ADI~\cite{ADI_huang2024learning}  {,} ImPoster~\cite{ImPoster_kothandaraman2024imposter} {,} FreeEvent\cite{FreeEvent_wang2024event} \\ 
                            \textbf{(f) Multiple Concept:} Custom Diffusion~\cite{CD_kumari2023multi} {,} SVDiff~\cite{Svdiff_han2023svdiff} {,} Cones2~\cite{Cones2_liu2024customizable} {,} MC2~\cite{MC2_jiang2024mc} {,}  OMG~\cite{Omg_kong2025omg} {,} ConceptWeaver~\cite{ConceptWeaver_kwon2024concept} {,} 
                            CIDM~\cite{CIDM_dong2024continually} {,} \\ SEGuidance~\cite{SEGuidance_liu2024training} {,} ConceptConductor~\cite{ConceptConductor_yao2024concept} {,} MagicTailor~\cite{MagicTailor_zhou2024magictailor} {,}  GroundingBooth~\cite{GroundingBooth_xiong2024groundingbooth} {,} MS-Diffusion~\cite{MS-Diffusion_personalizationms} {,} RelationBooth~\cite{RelationBooth_shi2024relationbooth} {,} \\ TokenVerse~\cite{TokenVerse_garibi2025tokenVerse}  \\
                            , leaf, text width=50em
                        ]    
				    ]
                    [
                        Text-driven Editing \\ (\S~\ref{sec:diffusion_text_editing})
                        [
                            DDS~\cite{DDS_hertz2023delta} {,} Prompt2Prompt~\cite{prompt2prompt_hertz2022prompt} {,} LEDIT~\cite{Ledits_tsaban2023ledits} {,} LEDITS++~\cite{Ledits++_brack2024ledits++} {,} MasaCtrl~\cite{Masactrl_cao2023masactrl} {,} Stable Flow~\cite{StableFlow_avrahami2024stable} {,} Head Router~\cite{HeadRouter_xu2024headrouter} {,} Xu~\etal~\cite{xu2024unveil} \\
                             , leaf, text width=50em
                        ]
                    ]
                ]
                [
                    AR Models \\ (\S~\ref{sec:ars})
                        [
                           Emu\cite{Emu_sun2023emu} {,} Emu2\cite{Emu2_sun2024generative} {,} Emu3\cite{Emu3_wang2024emu3} {,} Seed-X~\cite{SeedX_ge2024seed} {,} MetaMorph~\cite{MetaMorph_tong2024metamorph} {,} Chameleon~\cite{Chameleon_team2024chameleon} {,} PUMA~\cite{Puma_fang2024puma} {,} Liquid~\cite{Liquid_wu2024liquid} {,} ILLUME~\cite{ILLUME_wang2024illume} {,} X-Prompt~\cite{X-Prompt_sun2024x} \\
                            , leaf, text width=60.7em
                        ]                                    
                ]
            ]
        \end{forest}
    }
    \caption{\textbf{Taxonomy of Personalized Image Generation.}}
    \label{fig:taxonomy}
    \vspace{-1em}
\end{figure*}

\begin{figure*}[t]
\centering
\includegraphics[width=1\linewidth]{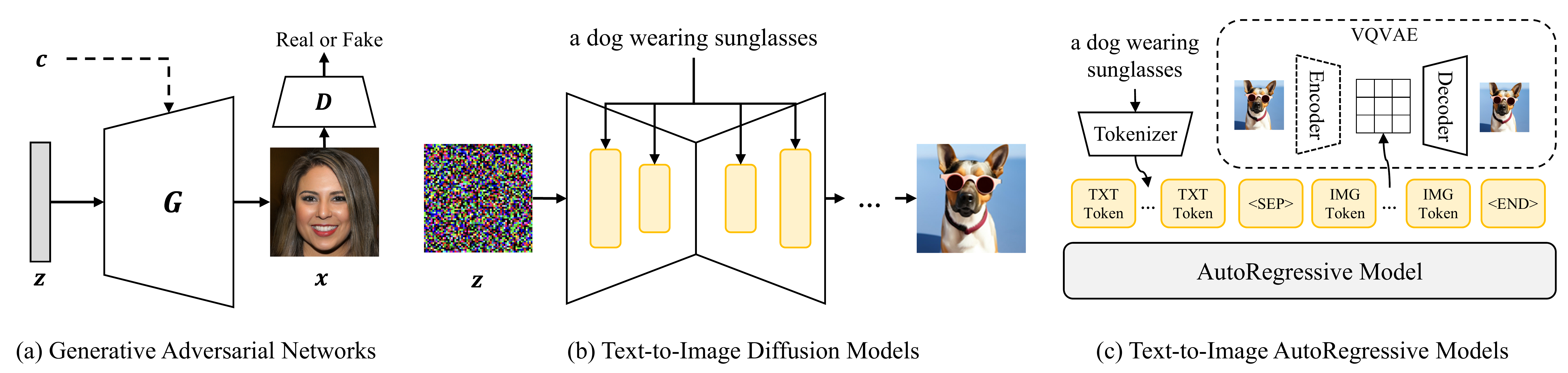}
\caption{\textbf{Illustration of different Generative Models.}}
\label{fig:generative_models}
\vspace{-1em}
\end{figure*}

\section{Problem Definition and Preliminary}
\label{sec:definition}

\subsection{Problem Definition}

Personalized image generation focuses on creating images that incorporate the user-specified concept (\eg, specific subjects, faces, or styles) and adhere to provided context. 
The visual concept is represented by a set of images, denoted as $\mathcal{X}_c$, which contain  the target concept $y_c$ (\eg, a pet corgi), while the desired context is indicated by a target text $y_t$ (\eg, ``wearing spacesuit'').
Here, we first define personalized image generation within a unified framework across different generative models. 
Formally, the generation process of a generative model $\mathit{G}$ can be described as:
\begin{equation}
    \mathbf{x} = \mathit{G}(\mathbf{z}, \mathbf{c}; \theta_\mathit{G}),
\end{equation}
where $\theta_\mathit{G}$ represents the parameters of $\mathit{G}$, and $\mathbf{z} \sim \mathcal{N}(\mathbf{0}, \mathbf{I})$ is a randomly sampled Gaussian noise. 
$\mathbf{c} = [c_0, \cdots, c_N]$ comprises a set of optional conditions, such as text or semantic maps.
Given the concept images $\mathcal{X}_c$ and target text $y_t$, personalized image generation aims to produce image $\mathbf{x}_c$ that incorporate the given visual concept $y_c$ and align with the description indicated by $y_t$,
\begin{equation}
    \mathbf{x}_c = \mathit{G}(\mathbf{z}, \mathbf{c}, y_t, \mathcal{X}_c; \theta_\mathit{G}).
\end{equation}
The generation process generally involves two main steps:
Firstly, in the \textit{concept inversion} step, the given concept image $\mathcal{X}_c$ are projected into a representation space to obtain the concept condition $\mathbf{c}_{c}$,
\begin{equation}
    \mathbf{c}_c = \phi(\mathcal{X}_c),
\end{equation}
where feeding $\mathbf{c}_c$ into $\mathit{G}$ yields an image containing the concept $y_c$.
$\phi(\cdot)$ denotes the inversion operation, and various \textbf{inversion spaces} and \textbf{inversion methods} can be employed for this projection.
Then, in the \textit{personalized generation} step, the concept condition $\mathbf{c}_c$ is integrated with the target text $y_t$ to generate the desired image:
\begin{equation}
    \mathbf{x}_c = \mathit{G}(\mathbf{z}, \mathbf{c}, y_t, \mathbf{c}_{c}; \theta_\mathit{G}).
\end{equation}
Various \textbf{personalization schemes} and concepts have been explored, tailored to specific generative models.

Based on this definition, we present a comprehensive review of personalization techniques across various generative models, including GANs, text-to-image diffusion models, and multi-model autoregressive models. 
These models, as illustrated in Fig.~\ref{fig:personalization_methods}, offer a generalized approach to personalized image generation.
In the following, we first provide an introduction to these representative generative models.

\subsection{Generative Models}
\label{sec:background}

\subsubsection{Generative Adversarial Networks}
Generative Adversarial Networks (GANs)~\cite{GAN_goodfellow2020generative} are a powerful class of generative models that consist of two neural networks: a generator and a discriminator, as shown in Fig.~\ref{fig:generative_models}(a). 
These networks are trained simultaneously through an adversarial process, where the generator aims to produce realistic images, and the discriminator strives to distinguish between real and generated (fake) images.
This adversarial training dynamic compels the generator to create increasingly realistic images over time.
In recent years, significant efforts have been dedicated to enhance the stability of training and the quality of GAN-generated images. 
These improvements focus on various areas, including network architectures~\cite{DCGAN_radford2015unsupervised,StackGAN_zhang2017stackgan,StyleGAN_karras2019style,StyleGAN2_karras2020analyzing}, loss functions~\cite{LSGAN_mao2017least,WGAN_arjovsky2017wasserstein,WGANGP_gulrajani2017improved,SNGAN_miyato2018spectral}, and training schemes~\cite{PGGAN_karras2017progressive,StyleGAN_karras2019style}.

Among these advancements, the style-based GAN series~\cite{StyleGAN_karras2019style,StyleGAN2_karras2020analyzing,StyleGAN3_karras2021alias} have garnered significant attention due to their superior ability to generate high-resolution images (\eg, 1024$\times$1024). 
Unlike traditional GANs that use latent noise directly as input, StyleGAN~\cite{StyleGAN_karras2019style} (as shown in Fig.~\ref{fig:inversion_space}) begins with a constant input and modulates the intermediate feature using projected latent styles.
A mapping network is adopted to transform the random noise into hierarchical latent styles.
This layer modulation mechanism enables the generator to control different aspects of the generated images through various layers.
Building on this foundation, StyleGAN2~\cite{StyleGAN2_karras2020analyzing} introduces weight demodulation to replace Adaptive Instance Normalization (AdaIN), further enhancing the perceptual quality and reducing artifacts in the generated images.

Beyond architectural and training improvements, several methods~\cite{BigGAN_brock2018large,StyleGANT_sauer2023stylegan,GigaGAN_kang2023scaling} have been proposed to condition image generation on specific inputs (\eg, category or text), enhancing the controllability of GANs.
Building upon these advancements, GAN-based personalization methods have been explored, which will be introduced in Sec.~\ref{sec:gans}.

\subsubsection{Text-to-Image Diffusion Models}

Denoising Diffusion Probabilistic Models (DDPMs)~\cite{DDPM_ho2020denoising} represent a novel class of generative models that have demonstrated remarkable ability in producing photo-realistic images.
Unlike GANs, Diffusion Models (DMs) produce image through a step-by-step denoising procedure that progressively converts noise into the desired output.
Specifically, they operate through two primary processes: the forward diffusion process and the reverse denoising process.
In the forward diffusion process, the model gradually adds Gaussian noise to the data $\mathbf{x}_0 \sim q(\mathbf{x}_0) $ over a series of steps (\eg, $T$ timesteps), finally converting the data into a noise distribution.
Conversely, the reverse denoising process aims to reconstruct the original data from the noisy inputs by learning to remove the added noise step-by-step.
This process begins from the randomly sampled noise $\mathbf{x}_T$ and transitions towards the original data distribution $q(\mathbf{x}_0)$. 
These processes are formulated as parameterized Markov chains, enabling DDPMs to generate highly detailed and realistic images from random noise.

In recent years, diffusion models have widely applied in the field of text-to-image (T2I) generation~\cite{ediffi_balaji2022ediff,LDM_rombach2022high,DALLE2_ramesh2022hierarchical,Imagen_saharia2022photorealistic,SDXL_podell2023sdxl,SD3_esser2024scaling,kolors}.
These models leverage pretrained text encoders, such as CLIP~\cite{CLIP_radford2021learning} or T5~\cite{T5_ni2021sentence} to transform textual information into feature representations.
This encoded text is then integrated into the image generation process through cross-attention mechanisms, allowing the models to produce images that align closely with the provided descriptions.
Benefiting from training on large-scale text-image datasets~\cite{Laion400M_schuhmann2021laion,Laion5b_schuhmann2022laion}, these T2I diffusion models are capable of generating textually coherent and high-quality images.
Among these advancements, the Stable Diffusion series~\cite{LDM_rombach2022high,SDXL_podell2023sdxl,SD3_esser2024scaling} are one of the representative open-sourced latent diffusion models, which is primarily utilized for personalized image generation.
To enhance computational efficiency, Stable Diffusion~\cite{LDM_rombach2022high} first employs an AutoEncoder to encode images into a latent space. 
A Unet-based diffusion model is then trained in the latent space, significantly reducing the computational resources required.
This approach has demonstrated the superior capacity in generating high-quality and diverse images, and facilitated a surge of recent advances in downstream tasks.
Building on the foundation of Stable Diffusion, Stable Diffusion XL~\cite{SDXL_podell2023sdxl} introduces a larger Unet architecture and an additional text encoder.
These enhancements resulted in improved image generation quality, greater textual fidelity, and support for higher resolution outputs.
More recently, Stable Diffusion 3~\cite{SD3_esser2024scaling} and FLUX~\cite{flux2023} further push the boundaries of image generation by incorporating diffusion transformer (DiT)~\cite{DIT_peebles2023scalable} based architectures and flow-matching training techniques~\cite{lipman2022flow}.
These innovations have led to higher quality and more controllable image generation capabilities.

Further discussions on integrating these models into personalized image generation are provided in Sec.~\ref{sec:diffusion}.

\subsubsection{Multi-model AutoRegressive Models}
AutoRegressive (AR) models employ the next-token prediction strategy, where each subsequent element in a sequence is predicted based on the preceding elements. 
These models have demonstrated exceptional performance in natural language processing (NLP)~\cite{radford2018improving,GPT2_radford2019language,GPT3_brown2020language}, showcasing impressive scalability, adaptability, and generalizability. 
Building on these successes, recent research has extended AR models to the domain of visual generation~\cite{DALLE_ramesh2021zero,Emu_sun2023emu,LlamaGen_sun2024autoregressive,NextGPT_wu2023next}. 
A pioneering effort in this field is DALL-E~\cite{DALLE_ramesh2021zero}, which converts both text and images into discrete tokens.
By using text tokens as a prefix condition, DALL-E learns to predict subsequent image tokens, effectively bridging the gap between textual descriptions and visual representations. 
Following it, several methods~\cite{Cogview_ding2021cogview,Cogview2_ding2022cogview2,Parti_yu2022scaling} have adopted similar framework to enhance text-to-image generation.
Beyond token-wise prediction, some approaches~\cite{VAR_tian2024visual,STAR_ma2024star,Varclip_zhang2024var} have introduced next-scale prediction, which generate images progressively from coarse to fine scales. 
Furthermore, recent advancements~\cite{NextGPT_wu2023next,Emu2_sun2024generative,SeedX_ge2024seed,MetaMorph_tong2024metamorph,Puma_fang2024puma} have explored the use of autoregressive models for multi-modal generation, combining both visual and textual information for image generation.
For example, Emu2~\cite{Emu2_sun2024generative} combines tokenized image embeddings with text tokens to predict the generated image embeddings in an autoregressive manner.
Moreover, diffusion models have been investigated as image decoders within these frameworks to refine and stabilize the quality of generated images~\cite{NextGPT_wu2023next,SeedX_ge2024seed,Emu_sun2023emu,Emu2_sun2024generative,Emu3_wang2024emu3}.

The multi-modal generation framework has also been investigated for personalized image creation. 
By integrating both visual and textual information, it can generate customized and contextually relevant images. 
The topic will be detailed explored in Sec.~\ref{sec:ars}.

\begin{figure*}[t]
\centering
\includegraphics[width=0.9\linewidth]{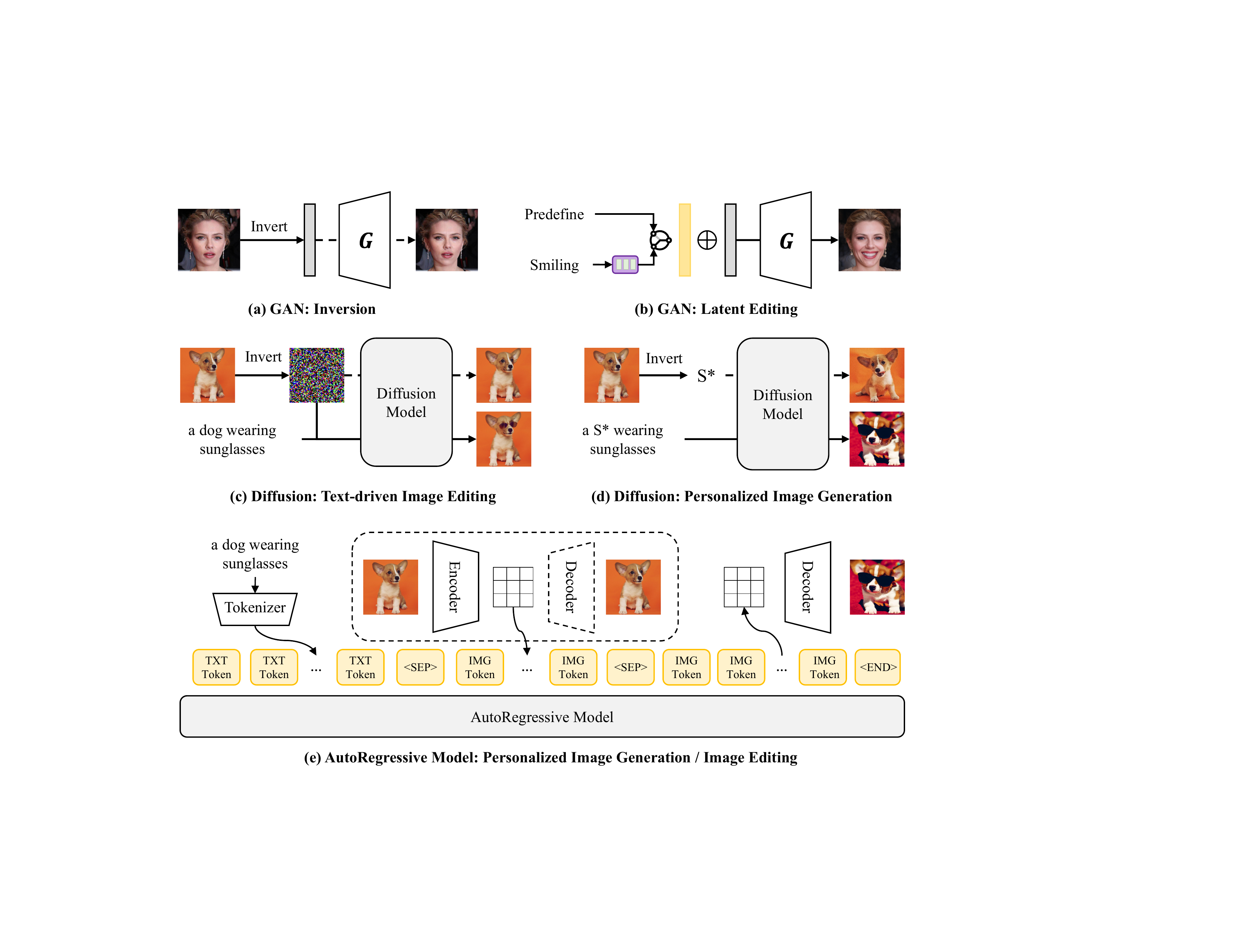}
\vspace{-0.5em}
\caption{\textbf{Generalized personalized image generation with generative models.} \textbf{(a)-(b): Generalized personalized image generation with Generative Adversarial Networks (GANs).} GAN Inversion first maps real images into a GAN's latent space, which can be used to reconstruct the input images. Then, latent editing is performed to generate personalized concepts with various attributes. The editing directions can be either predefined or derived from text. \textbf{(c) Text-driven image editing with text-to-image diffusion models.} After inverting the given image into noise space, text-driven editing techniques are applied to create target concepts with desired attributes specified by text. \textbf{(d) Personalized image generation with text-to-image diffusion models.} The target concept is inverted into the representation space of diffusion models, which is then directly combined with text prompts to generate the desired personalized images. \textbf{(e) Personalized image generation with multi-modal autoregressive models.} Images and text are encoded into a shared latent space, enabling the integration of these information to generate target personalized images. }
\label{fig:personalization_methods}
\vspace{-1em}
\end{figure*}

\begin{figure*}[t]
\centering
\includegraphics[width=1\linewidth]{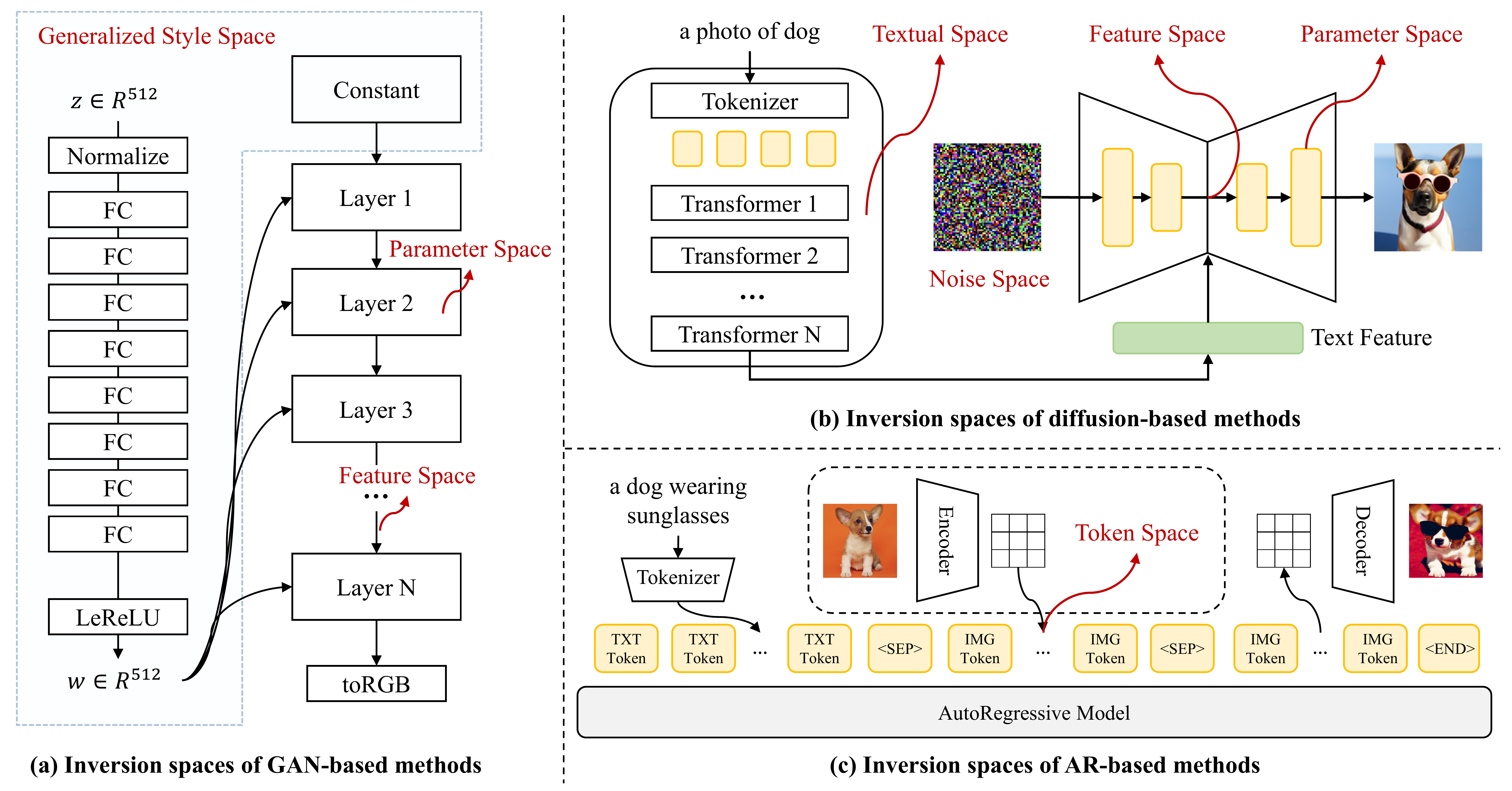}
\vspace{-2em}
\caption{\textbf{Inversion spaces of different generative models.} \textbf{(a)} For GAN-based personalization methods, the concept can be inverted into generalized style space, feature space, or parameter space. \textbf{(b)} For diffusion-based personalization methods, the concept can be inverted into noise space, textual space, feature space or parameter space. \textbf{(c)} For AR-based personalization methods, the concept is typically encoded into a shared space with text, referred to here as token space.}
\label{fig:inversion_space}
\vspace{-1em}
\end{figure*}

\section{Personalized Image Generation in GANs}
\label{sec:gans}

\subsection{Overview}

Although many GAN-based techniques can be used for image editing~\cite{liu2019stgan,Interfacegan_shen2020interfacegan,SeFa_shen2021closed}, we primarily focus on GAN inversion-based frameworks for generalized personalized image generation, as these frameworks share greater commonality with diffusion-based personalization methods.
In practice, these methods relies heavily on pretrained style-based GANs~\cite{StyleGAN_karras2019style,StyleGAN2_karras2020analyzing}.
Mathematically, the generation process can be described as: $\textbf{x}=\mathit{G}(\mathbf{z};\theta_\mathit{G})$.
Here, $\theta_\mathit{G}$ is the parameters of GAN model, and additional constants and spatial noise are omitted for simplicity.
As shown in Fig.~\ref{fig:personalization_methods}, given one concept image $\mathbf{x}_c$ (\eg, face), the personalized image generation with GANs involves two main stages.
In the first stage, the given concept image is inverted into the representation space of GAN, resulting in a concept condition $\mathbf{c}_c$.
This inversion ensures that when $\mathbf{c}_c$ is input into the $\mathit{G}$, it can accurately reconstruct the original concept image, \ie, $\mathbf{x}_c \approx \mathit{G}(\mathbf{c}_c)$.
Various inversion spaces and inversion methods have been developed to facilitate this process.
The second stage involves generating target images by editing the inverted concept condition.
In the subsequent sections, we will introduce different inversion spaces and inversion methods used in GAN Inversion, along with several editing techniques.

\subsection{Inversion Space}
\label{sec:gan_inversion_space}

As illustrated in Fig.~\ref{fig:inversion_space}, there are several potential spaces for GAN inversion.
In this discussion, we focus on StyleGAN, a widely adopted GAN architecture, as an example.
However, the definitions of these inversion spaces are general and can be extended to other GAN architectures.

\noindent \textbf{Generalized Style Space.} 
As shown in Fig.~\ref{fig:inversion_space}, StyleGANs~\cite{StyleGAN_karras2019style,StyleGAN2_karras2020analyzing} utilize a random noise vector $\mathbf{z}$ to generate the output image.
A natural choice to obtain concept condition $\mathbf{c}_c$ is to directly invert the concept image $\mathbf{x}_c$ back to a random noise $\mathbf{z}$, which spans the $\mathcal{Z}$ space.
However, the $\mathcal{Z}$ space typically follows a simple distribution (\eg, the standard Gaussian distribution), and the semantic features within this space are often entangled.
This entanglement makes it challenging to faithfully represent complex concepts.
To disentangle the semantics, the StyleGAN series~\cite{StyleGAN_karras2019style,StyleGAN2_karras2020analyzing,StyleGAN3_karras2021alias} introduce a multi-layer perceptron (MLP) to project the noise into a more disentangled $\mathcal{W}$ space.
Compared with $\mathcal{Z}$ space, $\mathcal{W}$ space offers better semantic disentanglement, allowing for a more faithful representation of concepts. 
%
%
Furthermore, benefiting from layer-wise modulation in StyleGANs, some approaches~\cite{Image2stylegan_abdal2019image2stylegan,Image2stylegan++_abdal2020image2stylegan++,psp_richardson2021encoding,katsumata2023revisiting} propose to predict an individual latent representation for each layer of the generator, resulting in $\mathcal{W}+$ space.
The $\mathcal{W}+$ space allows for more precise inversion and better controllability.
Furthermore, the $\mathcal{S}$ space~\cite{Stylespace_wu2021stylespace}, or stylespace, is introduced to further enhance controllability. 
The $\mathcal{S}$ space is defined by channel-wise style parameters $\mathbf{s}$, which serve as modulation parameters in each layer of the generator and are derived from $\mathbf{w}$.
Compared with $\mathcal{W}+$ space, $\mathcal{S}$ space~\cite{Stylespace_wu2021stylespace,StyleSpace1_liu2020style} offers superior spatial disentanglement, enabling fine-grained control over local features such as eyes, mouth, and hair. 
Overall, these various latent spaces primarily control the style of images. 
Collectively, they are referred to as generalized style spaces. 

\noindent \textbf{Feature Space.}
Although generalized style spaces demonstrate superior controllability over inverted concept, their limited capacities (\ie, 18 $\times$ 512 dimensions for $\mathcal{W}+$ space and 9088 dimensions for $\mathcal{S}$ space) hinder the faithful reconstruction of high-frequency details in the given concept.
To enhance detail consistency, numerous studies~\cite{HFGI_wang2022high,SAMInversion_parmar2022spatially} have proposed mapping the input concept image into both a $\mathbf{w}+$ vector and intermediate residual features, referred to as the feature space. 
The inverted residual feature, which has a larger capacity, is integrated with the corresponding layer's feature during generation to improve the preservation of high-frequency details.
However, these intermediate features typically maintain a fixed spatial structure, reducing their editability, particularly in the later layers of the network. 
Therefore, elaborate design~\cite{StylePrompter_zhuang2023styleprompter} is required to effectively balance the fidelity and editability.

\noindent\textbf{Parameter Space.}
To improve inversion fidelity, PTI~\cite{PTI_roich2022pivotal} introduces a novel approach that finetunes the generator's parameters to accurately reconstruct a given image. 
We refer to this as the parameter space.
The parameter space demonstrates enhanced capability to faithfully capture the detailed nuances of a given concept, even for out-of-domain details.

\begin{figure*}[t]
\centering
\includegraphics[width=1\linewidth]{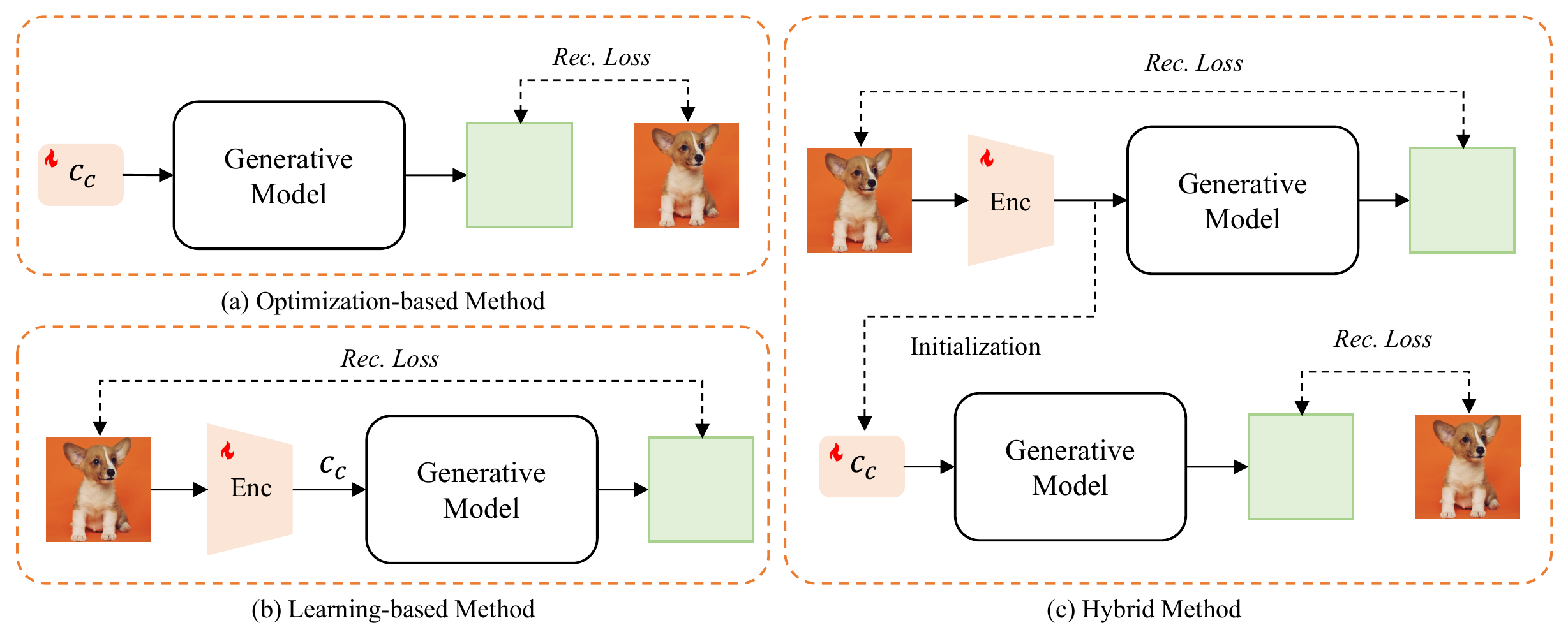}
\vspace{-2em}
\caption{\textbf{Different inversion methods for personalization.}  \textbf{(a) Optimization-based personalization methods} treat the concept condition as learnable parameters and optimize them directly. \textbf{(b) Learning-based personalization methods} employ an encoder to project the given concept into the concept condition. \textbf{(c) Hybrid methods} combine the strengths of learning-based techniques with optimization-based refinement. They use a learned encoder to obtain a coarse concept condition and then perform several optimization steps to enhance fidelity.}
\label{fig:inversion_methods}
\vspace{-1em}
\end{figure*}

\subsection{GAN Inversion Method}
\label{sec:gan_inversion_methods}

Given a pretrained GAN model $\mathit{G}$, a target inversion space, and a concept image $\mathbf{x}_c$, there are many GAN inversion methods to obtain the corresponding concept condition $\mathbf{c}_c$.
These methods can be broadly categorized into optimization-based, learning-based, and hybrid approaches.

\subsubsection{Optimization-based Method}
\label{sec:gan_inversion_optimization}

One intuitive approach to obtaining the concept condition is to optimize it directly, which we call optimization-based methods~\cite{GVM_zhu2016generative,Image2stylegan_abdal2019image2stylegan,feng2022near,InvertingGAN_creswell2018inverting,StochasticClipping_lipton2017precise,PGD_shah2018solving,bau2019inverting,ma2018invertibility}.  
The inversion process can be formulated as,
\begin{equation}
    \mathbf{c}^\ast_c = \arg\min_{{\mathbf{c}_c }}\ell({\mathit{G}(\mathbf{c}_c; \theta_\mathit{G})}, \mathbf{x}_c),
    \label{eqn:gan_inversion_opt}
\end{equation}
where $\ell$ represents the distance metric, such as $\ell_1$ or $\ell_2$ distance.
Here $\mathbf{c}_c$ can be an representation in various space, such as $\mathcal{W}+$ space or $\mathcal{S}$ space.
Given that the generator $\mathit{G}$ is differentiable, Eqn.~\ref{eqn:gan_inversion_opt} can be effectively solved using gradient descent techniques.

\noindent \textbf{Training Objective.} 
To ensure high reconstruction fidelity, several loss functions have been explored in optimization-based methods. 
For example, $\ell_2$ and perception loss are two common adopted losses~\cite{Image2stylegan_abdal2019image2stylegan,Image2stylegan++_abdal2020image2stylegan++}.
Image2StyleGAN++~\cite{Image2stylegan++_abdal2020image2stylegan++} further introduces a style loss to enhance the consistency of fine details in the reconstructed images. 
Building on observations from Zhu~\etal~\cite{PSpace_zhu2020improved}, BDInvert~\cite{BDInvert_kang2021gan} incorporates distribution regularization on the learned concept condition to improve inversion quality. 
In the domain of face images, face recognition loss is widely adopted to preserve identity fidelity, ensuring that the reconstructed image maintains the identity of the original concept image.

\noindent \textbf{Initialization Strategy.} 
A significant challenge in optimization-based inversion is the initialization of the concept condition $\mathbf{c}_c$.
Since the optimization of Eqn.~\ref{eqn:gan_inversion_opt} is highly non-convex, the quality of the reconstruction heavily depends on a good initialization.
A straightforward strategy involves starting with multiple random initializations and selecting the one that yields the minimal loss. 
However, this approach is computationally expensive and may not always provide accurate results.
To address this, Image2Stylegan~\cite{Image2stylegan_abdal2019image2stylegan} proposes using the average latent code of the pretrained generator as an initial point, thereby improving efficiency. 
Despite this improvement, such initialization methods may still fall short in accurately representing the given concept image.
Consequently, several methods~\cite{GVM_zhu2016generative,PTI_roich2022pivotal} leverage learning-based encoders (as discussed in Sec.~\ref{sec:gan_inversion_learning}) to provide more informed initializations, thereby enhancing both the accuracy and efficiency of the inversion process.

\begin{figure*}[t]
\centering
\includegraphics[width=1\linewidth]{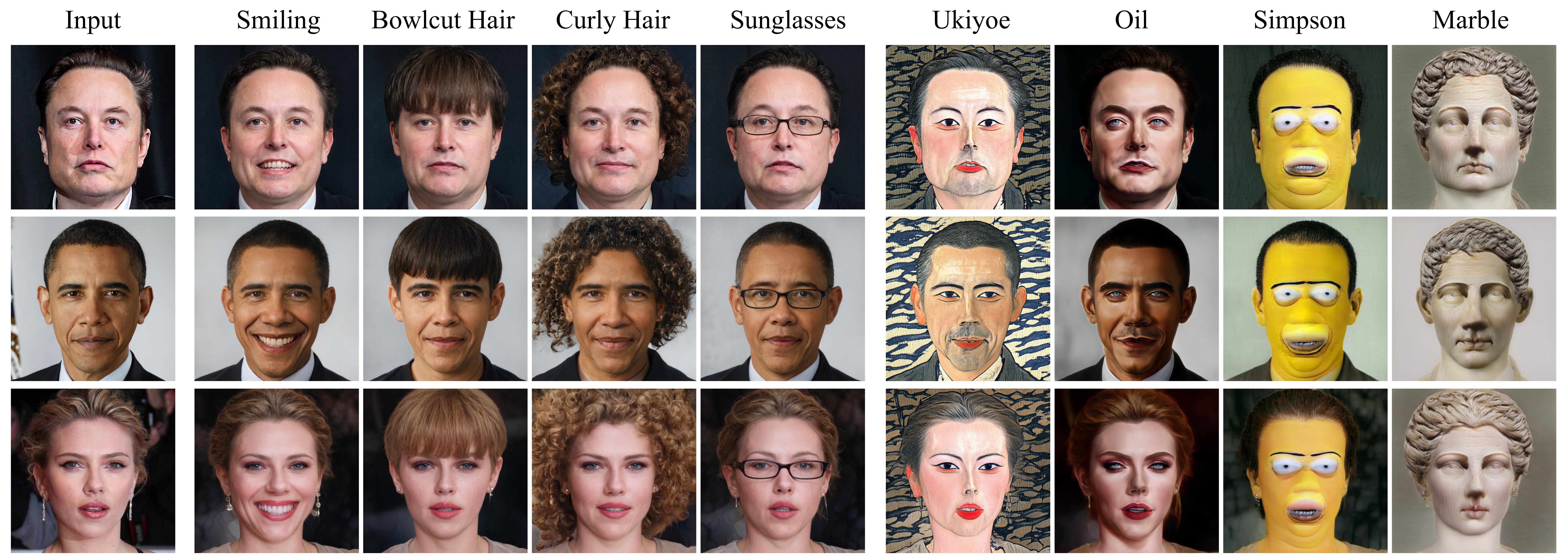}
\caption{\textbf{Visual generation results of GAN-based personalization methods.} Images generated by e4e~\cite{e4e_tov2021designing} combined with DeltaEdit~\cite{Deltaedit_lyu2023deltaedit} and StyleGAN-NADA~\cite{StyleganNada_gal2022stylegan}.}
\label{fig:gan_generation}
\vspace{-1em}
\end{figure*}

A notable drawback of optimization-based methods is their computational inefficiency.
Typically, the optimization procedure requires thousands of update iterations, resulting in prolonged inversion times.
To overcome this limitation, learning-based methods offer an alternative by predicting the latent representation in a single forward pass, significantly reducing the time required for inversion. 

\subsubsection{Learning-based Method}
\label{sec:gan_inversion_learning}

Learning-based GAN inversion methods~\cite{psp_richardson2021encoding,e4e_tov2021designing,Hyperstyle_alaluf2022hyperstyle,SAMInversion_parmar2022spatially} address efficiency challenges by incorporating an additional encoder, denoted as $\mathit{E}$, which maps input images to concept conditions in a forward pass.
The encoder is first pretrained on a dataset $\mathcal{X}$ to learn the mapping from images to their corresponding concept representations. 
This pretraining process is formulated as,
\begin{equation}
    \theta_\mathit{E}^\ast=\arg\min_{\theta_\mathit{E}}{\sum_\mathit{x_i \in \mathcal{X}}} \ell(\mathit{G}({\mathit{E}(\mathbf{x}_i; \theta_\mathit{E})}; \theta_\mathit{G}), \mathbf{x}_i),
    \label{eqn:gan_inversion_learning}
\end{equation}
where $\theta_\mathit{E}$ denotes the parameters of $\mathit{E}$, and $\mathit{G}$ is typically kept fixed during training.
The encoder's output can reside in any of the GAN's representation spaces as discussed in Sec.~\ref{sec:gan_inversion_space}.
Since the encoder is trained on comprehensive image datasets, it generally exhibits generalization capabilities to invert unseen images effectively.

\noindent \textbf{Encoder Design.} 
The design of the encoder is a pivotal aspect of learning-based GAN inversion methods.
Existing methods~\cite{psp_richardson2021encoding, StyleTransformer_hu2022style} commonly employ a ResNet backbone to extract image features, followed by a mapping network that predicts the concept condition within different spaces. 
For example, pSp~\cite{psp_richardson2021encoding} utilizes a `map2style' network to transform intermediate features into layer-wise $\mathbf{w}+$ vector.
StyleTransformer~\cite{StyleTransformer_hu2022style} adopts a transformer-based mapping to convert image features into $\mathbf{w}+$ vectors.
HyperStyle~\cite{Hyperstyle_alaluf2022hyperstyle} and HyperInverter~\cite{Hyperinverter_dinh2022hyperinverter} employ a hypernetwork to predict delta parameters of the generator.
Some methods~\cite{HFGI_wang2022high,SAMInversion_parmar2022spatially} extend to perform inversion across multiple spaces simultaneously.
For example, in addition to $\mathbf{w}+$ encoder, HFGI~\cite{HFGI_wang2022high} further includes an encoder that predicts delta features to enhance fidelity.
These methods often incorporate techniques that concatenation or subtraction between a coarse image generated by $\mathbf{w}+$ and the concept image to guide the inversion process.
To further improve inversion accuracy, ReStyle~\cite{Restyle_alaluf2021restyle} introduces an iterative refinement mechanism. 
At each step, the encoder concatenates the original image with the currently predicted image to predict the refined latent code.
Similarly, E2Style~\cite{E2Style_wei2022e2style} proposes a multi-stage refinement scheme that balances both accuracy and efficiency.

\noindent \textbf{Training Objective.}
During encoder training, learning-based GAN inversion methods typically employ a combination of loss functions to ensure high-quality inversion.
Two widely adopted loss functions are the $\ell_2$ loss and perceptual loss, which help preserve the overall structure and visual fidelity of the inverted images.
In addition to these, identity loss is commonly used to enhance identity consistency~\cite{psp_richardson2021encoding,e4e_tov2021designing,HFGI_wang2022high,SAMInversion_parmar2022spatially,StyleTransformer_hu2022style}. 
For applications in the face domain, models like ArcFace~\cite{Arcface_deng2019arcface} are typically employed as identity recognition models.
For other domains~\cite{e4e_tov2021designing}, recognition models trained with contrastive learning~\cite{MOCO_he2020momentum} are utilized.
Adversarial learning is also incorporated either in the latent space~\cite{e4e_tov2021designing} or directly on the images~\cite{e4e_tov2021designing,HFGI_wang2022high} to further improve the inversion quality.

\noindent \textbf{Training Dataset.}
To effectively train the encoder, existing approaches typically utilize datasets that are standard for GAN training, such as FFHQ~\cite{StyleGAN_karras2019style}, LSUN Church, Cars, Horses, and Cats~\cite{Lsun_yu2015lsun}.
These datasets provide a diverse set of images that help the encoder learn generalizable mapping.

\subsubsection{Hybrid Method}
\label{sec:gan_inversion_hybrid}

While learning-based GAN inversion methods offer high efficiency, they often face challenges in achieving perfect reverse mappings solely through an encoder.
This limitation stems from the encoder's inherent generalization capabilities, which may not fully capture the intricate complexities of the representation space required for accurate inversion.
To address this issue and enhance both efficiency and accuracy, hybrid methods have been proposed~\cite{GVM_zhu2016generative,bau2019inverting,IDInvert_zhu2020domain,PTI_roich2022pivotal,GANEnsembling_chai2021ensembling}.
These approaches integrate the strengths of learning-based techniques with optimization-based refinement, resulting in more precise and reliable GAN inversion.
One of the pioneering hybrid approaches was introduced by Zhu~\etal~\cite{GVM_zhu2016generative}.
This approach involves training an encoder network to provide a coarse initialization of the concept condition, which is then refined through optimization.
Building upon this foundation, IDInvert~\cite{IDInvert_zhu2020domain} further enhances the inversion process by incorporating discriminator regularization during training.
%
%
In contrast to methods that focus solely on optimizing latent codes, the PTI~\cite{PTI_roich2022pivotal} introduces a generator-tuning technique.
It begins with an initial latent code as a pivot and makes subtle adjustments to the pretrained generator. 
This tuning allows for the faithful reconstruction of input images, including those details that are out-of-domain, effectively mapping them into an in-domain space.

\subsection{Latent-based Image Editing}
\label{sec:gan_personalization}

With the inverted concept condition $c_c$, users further employ it to create personalized images. 
In GAN-based methods, personalization scheme is primarily achieved through latent editing. 
This process can be described as:
\begin{equation}
    \mathbf{x}_c = \mathit{G}(\mathbf{c} + \mathbf{n}; \theta_\mathit{G}),
\end{equation}
where $\mathbf{n}$ represents the direction associated with a specific edit, such as changing attributes like age, gender, or expression.  
The editing direction can be obtained either through latent navigation methods or derived from text inputs, providing flexible control over the generated image.

\subsubsection{Latent Navigation}

To find the editing directions of some attributes, several latent navigation methods have been proposed~\cite{Interfacegan_shen2020interfacegan,HijackGAN_wang2021hijack,HyperEditor_zhang2024hypereditor,yildirim2024warping,StyleFlow_abdal2021styleflow,parihar2023exploring}.
For example, InterFaceGAN~\cite{Interfacegan_shen2020interfacegan} employs the support vector machine (SVM) to learn a hyperplane that separates two binary attributes within the latent space (\eg, male and female). 
The normal vector of this hyperplane is then used as the editing direction, allowing for the manipulation of the corresponding attribute in the generated image.
HijackGAN~\cite{HijackGAN_wang2021hijack} trains a proxy model that maps input noise vectors to the attribute space. 
By computing the gradient of the proxy model with respect to the input noise, HijackGAN derives non-linear editing directions that facilitate attribute manipulation.
Viazovetskyi~\etal~\cite{Stylegan2Distillation_viazovetskyi2020stylegan2} utilizes a pre-trained image attribute classifier to determine the class centers of different attributes in the latent space. 
The direction between these class centers serves as the editing direction for altering specific attributes.
Instead of learning editing directions,  StyleFlow~\cite{StyleFlow_abdal2021styleflow} introduces a conditional normalizing flow model in the $\mathcal{W}$ space of StyleGAN.
This model maps latent vectors back into the noise space based on attribute conditions, enabling flexible attribute editing by generating edited latent vectors from new attribute conditions.
Parihar~\etal~\cite{parihar2023exploring} explore a new perspective for attribute editing and propose to learn the distribution over plausible attribute edits.
They train diffusion model~\cite{DDPM_ho2020denoising} within the latent space to capture the edit directions for each attribute.
This allows users to generate multiple edit variations for a given attribute and select the most suitable one.

In addition to supervised methods, several unsupervised approaches~\cite{SpectralRegularizer_ramesh2018spectral,wang2021geometry,HessianPenalty_peebles2020hessian,voynov2020unsupervised,WarpedGANSpace_tzelepis2021warpedganspace,Ganspace_harkonen2020ganspace,SeFa_shen2021closed} have been explored to discover editing directions without the need for manual attribute annotations.
For example, Ramesh~\etal~\cite{SpectralRegularizer_ramesh2018spectral} identifies directions corresponding to the principal eigenvectors of the latent space's covariance matrix, which is extended by Wang~\etal~\cite{wang2021geometry} via further introducing the Riemannian geometry metrics.
HessianPenalty~\cite{HessianPenalty_peebles2020hessian} proposes a Hessian-based regularization term to identify interpretable directions in the latent space that correspond to meaningful image transformations.
Voynov~\etal~\cite{voynov2020unsupervised} learn a set of editing directions by ensuring that perturbations along these directions are predictable by models.
WarpedGANSpace~\cite{WarpedGANSpace_tzelepis2021warpedganspace} introduces a Radial Basis Function (RBF) kernel to map latent representations into a non-linear space, enabling the learning of non-linear editing paths.
By performing Principal Component Analysis (PCA) on the features of early layers in the GAN, GANSpace~\cite{Ganspace_harkonen2020ganspace} identifies principal components that correspond to important factors of variation, and uses these components as semantic editing directions.
SeFa~\cite{SeFa_shen2021closed} derives a closed-form factorization method for latent semantic discovery, which utilizes the eigenvectors of the weights of fully connected layers as directions.
Despite the effectiveness of these methods in discovering latent editing directions, the semantic meaning of the learned directions is not inherently determined.
Human intervention is often required to interpret and label these directions based on their visual impact.
Recently, several methods~\cite{nguyen2024edit,DragGAN_pan2023drag} have explored drag-based image editing techniques, which can optimize the inverted latents to follow structural deformations based on the user's point instructions.

\subsubsection{Text-driven Editing}

In addition to latent navigation methods, editing directions can also be derived from textual inputs~\cite{Styleclip_patashnik2021styleclip,CLIP_radford2021learning,Stylemc_kocasari2022stylemc,Hairclip_wei2022hairclip,PPE_xu2022predict,clip2latent_pinkney2022clip2latent,zheng2022bridging,CLIPPAE_zhou2023clip,Contraclip_tzelepis2022contraclip,CLIPInverter_baykal2023clip,FFCLIP_zhu2022one,Deltaedit_lyu2023deltaedit}.
By leveraging natural language descriptions, users can specify desired changes in generated images through text prompts, allowing for more intuitive and flexible image generation.
One of the most influential approaches in this domain is StyleCLIP~\cite{Styleclip_patashnik2021styleclip}.
With the aid of CLIP~\cite{CLIP_radford2021learning}, it trains mappers that directly predict the editing direction corresponding to the given text input.
StyleMC~\cite{Stylemc_kocasari2022stylemc} further extends the framework by training multiple mappers to produce directions within the $\mathcal{S}$ space.
HairCLIP~\cite{Hairclip_wei2022hairclip} focuses specifically on hair style manipulation and trains a direction mapper that enables fine-grained text-based control over hairstyles.
To address the issue of attribute entanglement, PPE~\cite{PPE_xu2022predict} introduces a method to predict entangled attributes from synthesized images and incorporates an entanglement loss to prevent such entanglements. 
This ensures that changes in one attribute do not undesirably affect others.
Zheng~\etal~\cite{zheng2022bridging} propose to predict editing directions from the delta features of a CLIP encoder, achieving arbitrary text-driven manipulation without additional computation during inference. 
Similarly, DeltaEdit~\cite{Deltaedit_lyu2023deltaedit} learns to map the changes in CLIP image features to directions in the StyleGAN's $\mathcal{S}$ space, enabling seamless generalization to predict editing directions from changes in text features.
%

In addition to attribute editing, numerous methods have explored style editing within GANs.
StyleGAN-NADA~\cite{StyleganNada_gal2022stylegan} finetunes the GAN model using CLIP text prompts, enabling adaptation to specific style domains. 
This approach allows for the generation of images that embody target styles effectively.
Following it, Mind the GAP~\cite{MindTheGap_zhu2021mind} and DiFa~\cite{DiFa_zhang2022towards} further enhance style-driven generation capabilities by learning styles from user-specific reference images.
Several regularization techniques have been introduced to maintain diversity in the generated images.
Furthermore, HyperGAN-CLIP~\cite{HyperGANCLIP_anees2024hypergan} incorporates a hypernetwork to predict the modulated parameters of the generator.
The modulated weights are blended seamlessly with the original generator to enable the production of images that align with specified domains or tasks, such as reference-guided synthesis and text-guided manipulation.

\section{Personalized Image Generation in DMs}
\label{sec:diffusion}

\subsection{Overview}

Text-to-image (T2I) diffusion models, as illustrated in Fig.~\ref{fig:generative_models} (b), utilize a text prompt $y$ as condition and iteratively denoise randomly sampled noise $\mathbf{z}$ to generate corresponding image.
Benefiting from large-scale pretraining, these models can produce photo-realistic images based on textual descriptions.
Building upon these advancements, personalized image generation with diffusion models further enhances their capabilities to generate user-specific concepts, such as particular subjects, faces, or styles.
Typically, the concept is indicated by a set of images $\mathcal{X}_c$ containing the target concept (usually 3 $\sim$ 5 images), and the desired context for generation is specified via target text $y_t$.
As shown in Fig.~\ref{fig:personalization_methods} (d), similar to GANs, personalized image generation with diffusion models begins by inverting the given concept into the model's representation space to obtain a concept representation $\mathbf{c}_c$ (S*).
In contrast to GANs, the text-to-image framework allows these models to directly integrate the inverted concept condition with target text prompts to generate personalized images, offering superior controllability and flexibility.

In the subsequent sections, we will introduce various inversion spaces and inversion methods used in existing diffusion-based personalization methods.
Besides, the advanced generation capabilities of T2I diffusion models allow for the customization of a wide range of concepts (\eg, subject, face, \etc).
Therefore, we further introduce the different approaches categorized by their concept types.
Additionally, as depicted in Fig.~\ref{fig:personalization_methods} (c), text-driven image editing is considered a generalized form of personalization.
This approach allows users to modify existing images through textual descriptions, offering another layer of customization.
Thus, we also provide a brief overview of editing-based methods to complement the discussion on diffusion-based personalization.

\subsection{Inversion Space}
\label{sec:diffusion_inversion_space}

Text-to-image diffusion models provide flexible inversion spaces for personalized image generation.
In this discussion, we use Stable Diffusion~\cite{LDM_rombach2022high}, a widely used T2I model for personalized image generation, as an example.
As illustrated in Fig.~\ref{fig:inversion_space}, these spaces can be categorized into four types, \ie, noise space, textual space, feature space, and parameter space.
The definitions of these inversion spaces are general and can be extended to other models.

\noindent \textbf{Noise Space.}
Analogous to GANs, the noise space in diffusion models serves as a natural choice for representing individual concept images.
However, unlike GANs, the mathematical foundation of diffusion processes enables the direct inversion of a given image into the noise space through techniques such as DDIM inversion~\cite{DDIM_song2020denoising}. 
This inversion process does not require additional training, making it a straightforward and efficient method for image reconstruction.
Beyond the start noise, some approaches have explored the DDPM noise space~\cite{DDPMInversion_huberman2024edit,Ledits++_brack2024ledits++}, which encompasses the noise space across all diffusion steps.
This time-wise noise space enables a more precise representation of target images.
Additionally, several methods~\cite{NullInversion_mokady2023null} have investigated the use of unconditional noise within Classifier-Free Guidance (CFG)~\cite{CFG_ho2022classifier} to improve the inversion quality.
Collectively, these variations are referred to as the \textit{Noise Space}.
The inverted noise condition has same similar spatial dimensions with original image, and contains rich structure information.
Therefore, directly composing it with target text can not generate desired images~\cite{prompt2prompt_hertz2022prompt}, and elaborately designed image editing techniques are required to achieve personalized image generation, which are discussed in Sec.~\ref{sec:diffusion_text_editing}.
Furthermore, the editing flexibility of the noise space is constrained, particularly for non-rigid editing, limiting its applicability in more dynamic image generation.

\noindent \textbf{Textual Space.}
In text-to-image models, another intuitive and effective approach for personalization is representing the target concept using word, such as ``S*''. 
This allows the inverted concept to be seamlessly integrated with other textual descriptions to generate personalized images, such as ``Photo of S* wearing sunglasses''.
Researchers have explored various textual-related spaces~\cite{TI_gal2022image,CelebBasis_yuan2023inserting,Catversion_zhao2023catversion,ProFusion_zhou2023enhancing,UMM_ma2023unified,P+_voynov2023p+,NeTI_alaluf2023neural}.
Taking Stable Diffusion~\cite{LDM_rombach2022high} as an example.
As illustrated in Fig.~\ref{fig:inversion_space}, it employs a pre-trained CLIP text encoder~\cite{CLIP_radford2021learning} to transform text prompts into textual features.
During this process, each word or sub-word in the text prompt is first converted into a word embedding through an index-based lookup.
Then, the concatenated embeddings are sent to a transformer model to project as textual features, which are further used to steer the image generation.
Based on this, various methods~\cite{TI_gal2022image,CD_kumari2023multi,Catversion_zhao2023catversion,Cones2_liu2024customizable,UMM_ma2023unified} have been proposed to invert the concept into different layers of the text encoder.
For example, Textual Inversion~\cite{TI_gal2022image} proposes to learn a new word embedding to represent the target concept, spanning the \textit{Textual (Word) Space}.
This space is widely adopted because the learned embedding is plug-and-play, offering greater flexibility to be combined with other textual embeddings.
Instead, Cones2~\cite{Cones2_liu2024customizable} directly inverts the concept as the output feature of the text encoder, which is referred to as the \textit{Textual (Output) Space}.
CatVersion~\cite{Catversion_zhao2023catversion} represents the concept as intermediate features within the layers of the text transformer, spanning the \textit{Textual (Layer) Space}.
Furthermore, several extended textual spaces~\cite{P+_voynov2023p+,NeTI_alaluf2023neural,CelebBasis_yuan2023inserting} have been investigated to improve the identity fidelity and text controllability of the inverted concept.
For example, inspired by the $\mathcal{W}+$ space~\cite{Image2stylegan++_abdal2020image2stylegan++}, where an image is represented with layer-wise latents, $\mathcal{P}$+~\cite{P+_voynov2023p+} utilizes different word embeddings across various Unet layers to represent the target concept. 
This \textit{Textual ($\mathcal{P}+$) Space} allows for a more granular and detailed representation of the concept within the textual space.
NeTI~\cite{NeTI_alaluf2023neural} further expands the space as the time-wise \textit{Textual ($\mathcal{P}$*) Space}.
In the domain of human faces, CelebBasis~\cite{CelebBasis_yuan2023inserting} explores the \textit{Textual (Celeb) Space} by projecting the word embeddings of celebrities, which demonstrates superior editability. 
Collectively, these methods aim to generate text features that encapsulate both the target concept and the desired contextual information, and we refer to these spaces as Textual Spaces.

\noindent \textbf{Feature Space.}
In addition to representing concepts within textual-related spaces, several approaches~\cite{ELITE_wei2023elite,IPAdapter_ye2023ip,Instantbooth_shi2024instantbooth,UMM_ma2023unified,Fastcomposer_xiao2024fastcomposer} project the target concept into the intermediate feature space of diffusion models, commonly known as the \textit{Feature Space}.
The inverted concept features are then integrated with the text information to generate target images.
Unlike GANs, which typically inject inverted features into intermediate layers through addition, current diffusion-based personalization methods utilize adaptive adapters to inject the features, providing superior flexibility in generating concepts across various contexts.
Various types of adapters have been explored for integrating features into layers of diffusion models, including cross attention~\cite{IPAdapter_ye2023ip,ELITE_wei2023elite,wplus_li2024stylegan}, self attention~\cite{Instantid_wang2024instantid,SubjectDiffusion_ma2024subject}, and concatenation~\cite{Bootpig_purushwalkam2024bootpig}.
Compared to textual spaces, the feature space captures much finer details of the concept, thereby enhancing the fidelity of the generated concept.  
Additionally, the feature space usually incorporates with learning-based concept learning framework for personalization, which will be discussed in Sec.~\ref{sec:diffusion_inversion_methods}.

\noindent \textbf{Parameter Space.}
Similar to GANs, several methods~\cite{Dreambooth_ruiz2023dreambooth,Cones_liu2023cones,CD_kumari2023multi,OFT_qiu2023controlling,Perfusion_tewel2023key,Svdiff_han2023svdiff,Lora_hu2021lora} finetune the parameters of the model to learn the new concept, spanning the \textit{Parameter Space}.
For example, DreamBooth~\cite{Dreambooth_ruiz2023dreambooth} finetunes the parameters of diffusion Unet to align the user-specific concept with a unique identifier (\eg, S*). 
To reduce the storage burden of full parameter finetuning, techniques like LoRA~\cite{Lora_hu2021lora} are commonly used.
Additionally, instead of adjusting all parameters, some methods~\cite{Cones_liu2023cones,CD_kumari2023multi,Perfusion_tewel2023key,Svdiff_han2023svdiff} focus on identifying and modifying only key parameters essential for personalized generation.
This not only minimizes computational overhead but also helps preserve the model's original priors.
Compared to other inversion spaces, parameter space demonstrates a greater capacity to faithfully represent the given concept while offering enhanced text controllability.
Meanwhile, optimization-based concept learning framework is typically employed to learn new parameters, which will be discussed in Sec.~\ref{sec:diffusion_inversion_methods}.

It is worth noting that these spaces are not mutually exclusive.
A single concept can be inverted into multiple spaces simultaneously. 
For example, Custom Diffusion~\cite{CD_kumari2023multi} inverts a concept into both the textual (word) space and the parameter space, leveraging the strengths of each space to achieve more robust personalized generation.

\subsection{Concept Inversion Method}
\label{sec:diffusion_inversion_methods}

Given a target representation space within a diffusion model, several methods can be utilized to obtain the concept condition of a given concept.
Specifically, we categorize them into four types, including training-free methods, optimization-based methods, learning-based methods, and hybrid methods.

\subsubsection{Training-Free Method}

Training-free inversion methods~\cite{DDIM_song2020denoising,DDPMInversion_huberman2024edit,DSG_epstein2023diffusion,PickandDraw_lv2024pick,rout2024semantic} leverage the inherent properties of diffusion models to invert concepts without the need for additional training. 
For instance, DDIM Inversion~\cite{DDIM_song2020denoising} provides a technique to invert the given image as the start noise that can be used to reconstruct the target image.
To improve the reconstruction quality, DDPM Inversion~\cite{DDPMInversion_huberman2024edit} projects the image into DDPM noise space, which encompasses noise maps across all diffusion steps.
Besides the noise representation, some methods~\cite{DSG_epstein2023diffusion,PickandDraw_lv2024pick} explore intermediate representations within the diffusion models to achieve training-free personalization.
For example, Diffusion Self Guidance~\cite{DSG_epstein2023diffusion} leverages activations from Unet at each timestep to represent the coarse appearance of the image, while using self and cross-attention maps to capture structural information.
During inference, it aligns the structure and appearance features of generated image with that of the reference image through gradient guidance, allowing for personalized image generation without additional training.
Similarly, Pick-and-Draw~\cite{PickandDraw_lv2024pick} adopts earth movers distance (EMD) to calculate the spatial adaptive distance among the appearance features, which helps generate subjects consistent with the reference image.
These training-free inversion methods offer a straightforward approach to achieving personalization without requiring extra training.

\subsubsection{Optimization-based Method}

Optimization-based concept learning methods~\cite{TI_gal2022image,Dreambooth_ruiz2023dreambooth,CD_kumari2023multi,ViCo_hao2023vico,Catversion_zhao2023catversion,UnZipLoRA_liu2024unziplora,Disenbooth_chen2023disenbooth,DCO_lee2024direct,Dreamtuner_hua2023dreamtuner,ComFusion_hong2024comfusion} treat the concept condition $\mathbf{c}_c$ as learnable parameters and optimize them based on the given concept images $\mathcal{X}_c$.
The optimization process can be formulated as,
\begin{equation}
\mathbf{c}^\ast_c = \arg \min_{\mathbf{c}_c}  \mathbb{E}_{\mathbf{x}_c\in\mathcal{X}_c, y, \epsilon, t}\Big[ \Vert \epsilon - \epsilon_\theta(\mathbf{z}_{t}, t, \tau(y), \mathbf{c}_c) \Vert_{2}^{2}\Big] \, ,
\label{eq:diffusion_opt}
\end{equation}
where $\epsilon_\theta(\cdot)$ denotes pretrained diffusion models, and here we take Stable Diffusion~\cite{LDM_rombach2022high} as an example.
$\epsilon$ represents the unscaled noise, $t$ is the time step. s
$\mathbf{z}_t$ is encoded latent noise of image $\mathbf{x}$ at time $t$.
$\tau(\cdot)$ represents the pretrained CLIP text encoder~\cite{CLIP_radford2021learning}, and $y$ is the input text.

\noindent \textbf{Training Prompt Construction.} 
As shown in Eqn.~\ref{eq:diffusion_opt}, a text $y$ related to concept image $\mathbf{x}_c$ is required during optimization.
The choice of text prompts significantly influences the learning of the concept condition~\cite{he2023data,DCO_lee2024direct}.
Textual Inversion~\cite{TI_gal2022image} uses simple text templates, such as ``a photo of S*'', as training prompts.
Here S* denotes the pseudo word representing the target concept.
Dreambooth~\cite{Dreambooth_ruiz2023dreambooth} further incorporates a class word in the text prompt, for example, ``a photo of S* dog''.
The inclusion of the class word provides the category prior and facilitates the learning of target concept.
To disentangle target concept from irrelevant context (\eg, pose and background), several methods~\cite{he2023data,DCO_lee2024direct} also include additional contextual information of input image in the text, such as ``a photo of S* running on the beach''.
Some approaches~\cite{DETEX_cai2024decoupled,Disenbooth_chen2023disenbooth} further introduce learnable embeddings into text prompts to explicitly capture irrelevant information.
For example, DETEX~\cite{DETEX_cai2024decoupled} employs prompts like ``a photo of S* dog with [P] pose and [B] background'' during training, where [P] and [B] are learnable embeddings designed to capture irrelevant pose and background information, respectively.

\noindent \textbf{Data Augmentation.}
During optimization, only a small number of training samples (typically 3$\sim$5 images) is adopted.
This limited dataset size usually leads to overfitting of the learned concept conditions, resulting in poor editability.
To address this issue, several methods~\cite{CD_kumari2023multi,ProFusion_zhou2023enhancing,Dreambooth_ruiz2023dreambooth,CelebBasis_yuan2023inserting} employ data augmentation techniques to enhance the generalization. 
For example, Custom Diffusion~\cite{CD_kumari2023multi} utilizes random cropping, flipping, and resizing to increase the diversity of the training data. 
Similarly, ProFusion~\cite{ProFusion_zhou2023enhancing} adopts an inpainting approach to augment the background of images, further improving dataset variability.

\noindent \textbf{Training Objective.}
To effectively learn the target concept while preserving the prior of T2I models, various losses have been introduced~\cite{Dreambooth_ruiz2023dreambooth,Singleinsert_wu2023singleinsert,FaceChainSuDe_qiao2024facechain,Disenbooth_chen2023disenbooth,DETEX_cai2024decoupled,BreakAScene_avrahami2023break}.
For example, during finetuning, Dreambooth~\cite{Dreambooth_ruiz2023dreambooth} introduces a prior preservation regularization to maintain the prior of original T2I model and prevent language drift.
A predefined dataset consisting of images from a specific category is required to calculate the regularization.
Some methods~\cite{Singleinsert_wu2023singleinsert,DCO_lee2024direct,CoRe_wu2024core} regularize the finetuned model to remain consistent with the original model across various representations, thereby improving prior preservation.
Besides, several losses~\cite{ComFusion_hong2024comfusion,Imagereward_xu2024imagereward,FaceChainSuDe_qiao2024facechain} have been proposed to improve the text controllability of context and actions.
Additionally, attention loss and masked diffusion loss~\cite{Disenbooth_chen2023disenbooth,DETEX_cai2024decoupled,BreakAScene_avrahami2023break} are commonly used to improve the fidelity of learned concepts.

\noindent \textbf{Initialization.}
Similar to GANs, initialization plays an important role to affect the performance of personalization in diffusion models.
Existing initialization methods~\cite{TI_gal2022image,NeTI_alaluf2023neural,P+_voynov2023p+} are primarily designed for the textual (word) space and typically initialize the textual embedding with a super-category token (\eg, ``dog'' or ``cat'').
To speed up the optimization process, Cross Initialization~\cite{CrossInitialization_pang2024cross} employs the mean textual embedding of 691 well-known names as the initialization, which provides the prior of human identities.

\subsubsection{Learning-based Method}
\label{sec:diffusion_learning_methods}

Optimization-based methods usually suffer from low efficiency, requiring several or tens of minutes to learn a new concept.
To address this, learning-based concept inversion methods~\cite{ELITE_wei2023elite,IPAdapter_ye2023ip,Instantbooth_shi2024instantbooth,E4T_gal2023encoder,Fastcomposer_xiao2024fastcomposer} have been developed that employ encoders to directly invert given images as concept condition. 
The learning process of encoder can be formulated as follow,
\begin{equation}
\theta_\mathit{E}^* = \arg \min_{\theta_\mathit{E}}  \mathbb{E}_{\mathbf{x}\in\mathcal{X}, y, \epsilon, t}\Big[ \Vert \epsilon - \epsilon_\theta(\mathbf{z}_{t}, t, \tau(y), \mathit{E}(\mathbf{x}; \theta_\mathit{E})) \Vert_{2}^{2}\Big] \, ,
\label{eq:diffusion_learn}
\end{equation}
where $\mathit{E}$ is the encoder to project the given concept image as the concept condition, that $\mathbf{c}_c = \mathit{E}(\mathbf{x}; \theta_{\mathit{E}})$, and $\theta_{\mathit{E}}$ is the parameter of $\mathit{E}$.
It is worth noting that learning-based inversion methods usually use a single concept image as reference for personalization, but they can be easily extended to accommodate multiple reference images.

\noindent \textbf{Encoder Design.}
In learning-based methods, the design of the encoder is crucial for achieving effective and faithful concept inversion.
Existing approaches~\cite{ELITE_wei2023elite,IPAdapter_ye2023ip,UMM_ma2023unified,Anydoor_chen2024anydoor,Instantid_wang2024instantid,Bootpig_purushwalkam2024bootpig,BLIPDiffusion_li2024blip,MOMA_song2025moma,photomaker_li2024photomaker,Fastcomposer_xiao2024fastcomposer} typically utilize an image encoder followed by a mapping model to project the input image into a concept condition.
Several image encoders have been explored to extract semantically rich features, including CLIP image encoder~\cite{ELITE_wei2023elite,IPAdapter_ye2023ip,UMM_ma2023unified}, DINO encoder~\cite{Anydoor_chen2024anydoor}, reference Unet~\cite{Instantid_wang2024instantid,Bootpig_purushwalkam2024bootpig}, and multi-modal encoder~\cite{BLIPDiffusion_li2024blip,MOMA_song2025moma}.
For example, ELITE~\cite{ELITE_wei2023elite} employs the pretrained CLIP image encoder to extract the multi-layer features, and utilizes global and local mapping to project them into the textual word space and feature space, respectively.
This dual mapping approach allows ELITE to achieve flexible text controllability while maintaining rich details of the target concept.
Similarly, IP-Adapter~\cite{IPAdapter_ye2023ip} uses a simple linear projection to encode the projected CLIP image feature as image embeddings.
It also alternatively explores the use of Resampler to capture finer details from extracted features, enhancing the fidelity of the generated concepts.
To improve the concept fidelity, various methods~\cite{Instantid_wang2024instantid,Bootpig_purushwalkam2024bootpig} adopt reference Unet to extract time- and layer-wise features from the given concept image, which captures intricate details of the given concept.
Moreover, some approaches ~\cite{BLIPDiffusion_li2024blip,MOMA_song2025moma} leverage multi-modal models to extract concept information alongside text input, allowing for more precise and flexible control over the generated images.
Face customization methods~\cite{photomaker_li2024photomaker, CelebBasis_yuan2023inserting} also employ ID encoders~\cite{Arcface_deng2019arcface,TFace_huang2020curricularface} to extract domain-specific information.

To integrate the extracted features into diffusion model to guide the concept generation, several types of adapters have been explored, such as cross attention~\cite{IPAdapter_ye2023ip,ELITE_wei2023elite,wplus_li2024stylegan,Pulid_guo2024pulid}, self attention~\cite{Instantid_wang2024instantid,SubjectDiffusion_ma2024subject}, and concatenation~\cite{Bootpig_purushwalkam2024bootpig} mechanisms.
For example, IP-Adapter~\cite{IPAdapter_ye2023ip} employs dual cross attention mechanism that calculates text and image attention in parallel and subsequently fuses them for personalization.
Additional key and value projections are introduced into the cross attention layers for image attention calculation.
For recent DiT-based diffusion models, such as FLUX~\cite{flux2023} and Stable Diffusion 3~\cite{SD3_esser2024scaling}, a new cross attention is usually introduced after each attention block to inject the concept information~\cite{Pulid_guo2024pulid}.

\noindent \textbf{Dataset Construction.}
Existing learning-based methods pretrain the encoders and adapters using self-constructed datasets.
The training datasets~\cite{ELITE_wei2023elite,IPAdapter_ye2023ip,SubjectDiffusion_ma2024subject} are typically composed of triplet image pairs, \ie, \{reference image, target text, target image\}, where the reference image provides concept information.
Additional information such as bounding boxes~\cite{SubjectDiffusion_ma2024subject,MS-Diffusion_personalizationms} and masks~\cite{ELITE_wei2023elite,Fastcomposer_xiao2024fastcomposer,photomaker_li2024photomaker} may also be incorporated to facilitate the model training.
In practice, the training datasets used are typically constructed from existing corpus, such as LAION 5B~\cite{Laion5b_schuhmann2022laion} and LAION 400M~\cite{Laion400M_schuhmann2021laion}, where the reference image and the target image are derived from the same source image~\cite{SubjectDiffusion_ma2024subject,IPAdapter_ye2023ip}.
However, training model on these datasets usually leads to copy-paste issues, resulting in poor text editability~\cite{MasterWeaver_wei2025masterweaver,Pulid_guo2024pulid,IPAdapter_ye2023ip}. 
To address this issue, several methods~\cite{photomaker_li2024photomaker,Dreamidentity_chen2023dreamidentity,SUTI_chen2024subject,MS-Diffusion_personalizationms,Imagine-yourself_he2024imagine,RelationBooth_shi2024relationbooth,JeDi_zeng2024jedi,cai2024diffusion,Bootpig_purushwalkam2024bootpig,DreamCache_aiello2024dreamcache,CustomizationAssistant_zhou2024customization} propose to create the unpaired datasets, where the reference image and the target image originate from different sources.
For example, PhotoMaker~\cite{photomaker_li2024photomaker} collects multiple images of the same celebrities to form an unpaired dataset, while MS-Diffusion~\cite{MS-Diffusion_personalizationms} construct such an unpaired dataset by leveraging existing video datasets.
Beyond collecting images from real-world sources, some methods~\cite{SUTI_chen2024subject,CustomizationAssistant_zhou2024customization} propose to generate datasets synthetically.
SUTI~\cite{SUTI_chen2024subject} trains millions of optimization-based models~\cite{Dreambooth_ruiz2023dreambooth}, and uses them to produce unpaired target images based on given text.
CustomizationAssistant~\cite{CustomizationAssistant_zhou2024customization} adopts a self-distillation approach, where a previously trained personalized model is used to generate unpaired images.
Additionally, several methods~\cite{Dreamidentity_chen2023dreamidentity,Imagine-yourself_he2024imagine,RelationBooth_shi2024relationbooth,JeDi_zeng2024jedi,cai2024diffusion,Bootpig_purushwalkam2024bootpig,DreamCache_aiello2024dreamcache} exploit the advanced generative capabilities of pretrained text-to-image models to construct unpaired datasets.
For instance, JeDi~\cite{JeDi_zeng2024jedi} uses the pretrained SDXL model to generate a dataset of same-subject photo collages by appending the text ``photos of the same'' to each of the text prompts.
Table~\ref{tab:data_source} provides a detailed overview of the training datasets used in learning-based methods.

\noindent \textbf{Training Objective.}
In addition to Eqn.~\ref{eq:diffusion_learn}, several loss functions~\cite{Fastcomposer_xiao2024fastcomposer,wplus_li2024stylegan,Pulid_guo2024pulid,MasterWeaver_wei2025masterweaver} have been proposed to facilitate the learning of concept identities while disentangling irrelevant information such as pose and background.
Similar to optimization-based methods, attention loss and masked diffusion loss~\cite{Fastcomposer_xiao2024fastcomposer,wplus_li2024stylegan} are commonly used to improve the fidelity and quality of learned concepts.
To further enhance text controllability, several methods~\cite{Pulid_guo2024pulid,MasterWeaver_wei2025masterweaver} propose to regularize the intermediate features of personalization models to align them with those of the original T2I model. 
The identity loss is also employed for face personalization~\cite{Pulid_guo2024pulid}.

\subsubsection{Hybrid Method}

Similar to GANs, while learning-based methods offer high efficiency, they often encounter challenges in generalization and struggle to achieve adequate identity fidelity through an encoder.
To address this, hybrid methods~\cite{E4T_gal2023encoder,BLIPDiffusion_li2024blip,DAT_arar2023domain,Hyperdreambooth_ruiz2024hyperdreambooth,HybridBooth_guan2025hybridbooth} have been adopted, which integrates the strengths of learning-based techniques with optimization-based refinement.
For example, BLIP-Diffusion~\cite{BLIPDiffusion_li2024blip} employs a learned encoder to initially obtain the concept condition, and optimizes this condition over several steps to improve identity consistency. 
After obtaining the concept condition, HybridBooth~\cite{HybridBooth_guan2025hybridbooth} further optimizes the model parameters to enhance identity fidelity. 
Compared with existing learning-based and optimization-based methods, hybrid methods demonstrate a comparable learning speed while maintaining superior fidelity.

\begin{figure*}[t]
\centering
\includegraphics[width=1\linewidth]{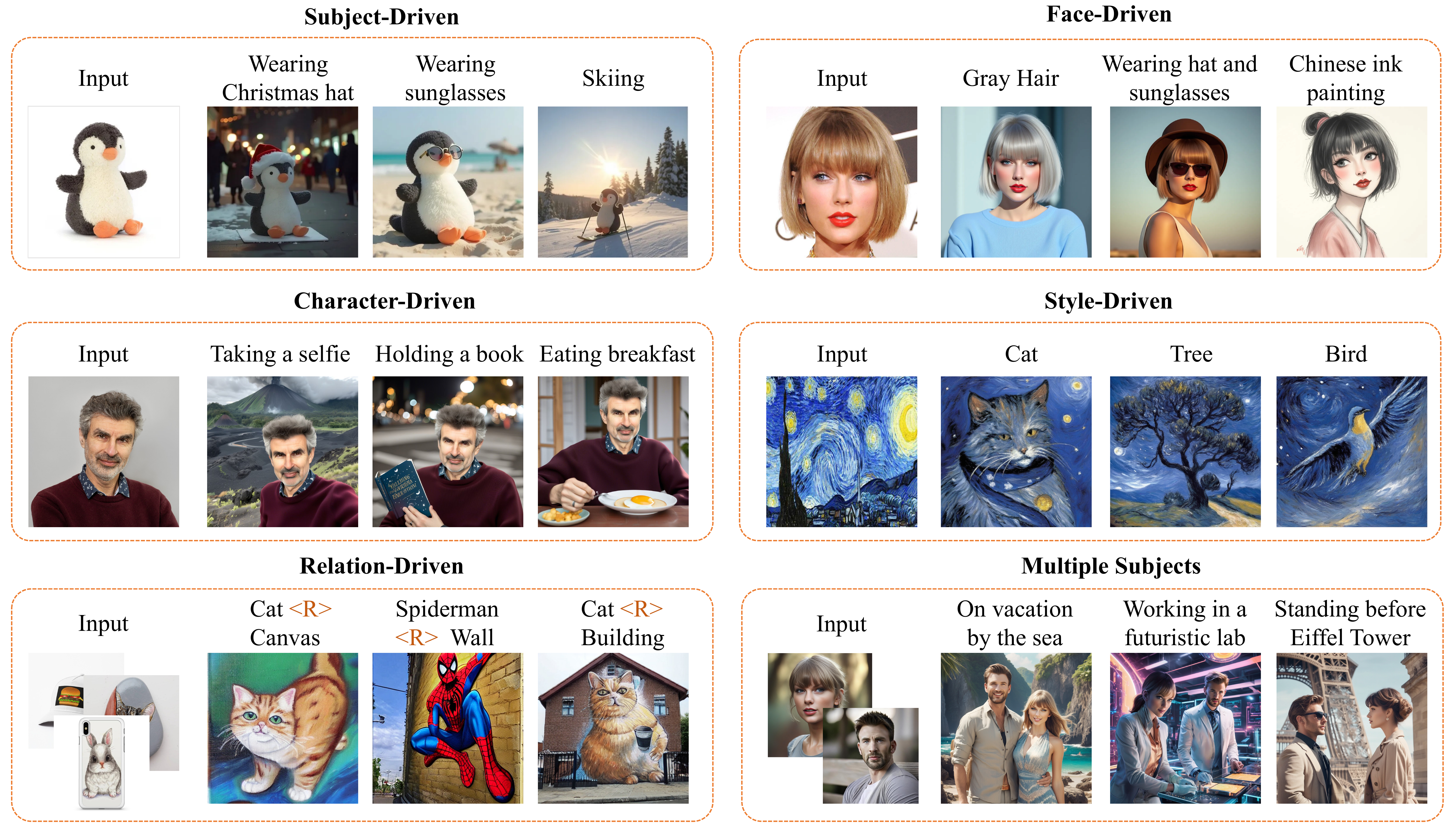}
\caption{\textbf{Visual generation results of diffusion-based personalization methods.} Images generated by OminiControl~\cite{OminiControl_tan2024ominicontrol}, PuLID~\cite{Pulid_guo2024pulid}, StoryMaker~\cite{StoryMaker_zhou2024storymaker}, InstantStyle~\cite{Instantstyle_wang2024instantstyle}, Reversion~\cite{ReVersion_huang2024reversion}, and OMG~\cite{Omg_kong2025omg}.}
\label{fig:diffusion_generation}
\vspace{-1em}
\end{figure*}

\subsection{Personalized Image Generation}
\label{sec:diffusion_personalization}

Benefiting from the text-to-image framework, diffusion-based personalization methods enable the direct integration of inverted conditions with target text prompts to generate personalized images, offering superior controllability and flexibility.
In this section, we provide an overview of personalized image generation across various categories, including subject, face, and style, \etc.

\subsubsection{Subject-Driven Personalization}
\label{sec:diffusion_personalization_subject}

Subject-driven personalization is a widely researched area that aims to generate images containing the subject concept from reference images. 
For better understanding, we will introduce them sequentially based on their inversion methods, \ie, optimization-based, learning-based, \etc.

Given a small set of images containing target subject, Textual Inversion~\cite{TI_gal2022image} learns a new word embedding to represent the given subject.
After optimization, it can be used for personalized subject generation in a plug-and-play manner without affecting the priors of T2I models.
Similar to Textual Inversion, DreamBooth~\cite{Dreambooth_ruiz2023dreambooth} introduces a rare word as a unique identifier to represent the target subject. 
A prior-preserved parameter finetuning is adopted to align the unique identifier with the given subject.
These two methods give the the foundation for numerous subsequent optimization-based methods~\cite{CD_kumari2023multi,DCI-ICO_jin2025customized,ViCo_hao2023vico,HiPer_han2023highly,P+_voynov2023p+,Catversion_zhao2023catversion,NeTI_alaluf2023neural,Dreamartist_dong2022dreamartist,HiFiTuner_wang2023hifi,NestedAttention_patashnik2025nested}.

Identity fidelity is one of key factors in subject-driven personalization.
To improve the identity fidelity of learned concept, various inversion spaces~\cite{ViCo_hao2023vico,HiPer_han2023highly,P+_voynov2023p+,Catversion_zhao2023catversion,NeTI_alaluf2023neural,Dreamartist_dong2022dreamartist,HiFiTuner_wang2023hifi,TextBoost_park2024textboost,Dreambooth_ruiz2023dreambooth,Lora_hu2021lora,OFT_qiu2023controlling,P3S-Diffusion_hu2024p3s} have been explored, as discussed in Sec.~\ref{sec:diffusion_inversion_space}.
For example, inspired by the $\mathcal{W}+$ space in StyleGANs, $\mathcal{P}+$~\cite{P+_voynov2023p+} proposed to use the different word embeddings for different layers to represent the target concept.
A similar layer-wised disentanglement is found in stable diffusion Unet, where color is determined by the fine outer layers and content is determined by the coarse inner layers.
NeTI~\cite{NeTI_alaluf2023neural} further extents $\mathcal{P}+$ space across time dimension, resulting in time and layer independent embeddings.
Besides, ViCo~\cite{ViCo_hao2023vico} incorporates the image information to enhance the subject details.

Parameter tuning methods demonstrate enhanced representational capacity to capture the intricate details of a given subject. 
However, it brings increased computational demands and potential prior degradation due to limited dataset scale~\cite{CD_kumari2023multi,Cones_liu2023cones,Perfusion_tewel2023key}.
To address this, Custom Diffusion~\cite{CD_kumari2023multi} sidentifies critical parameters for personalization by exploiting several existing DreamBooth models. 
Specifically, it targets the key and value projections within cross-attention layers for updating.
Following it, several methods~\cite{Perfusion_tewel2023key,Svdiff_han2023svdiff,PersonalizedResiduals_ham2024personalized,Cones_liu2023cones} have been proposed to explore different key parameters for personalization, such as neurons~\cite{Cones_liu2023cones}, value projection weights~\cite{Perfusion_tewel2023key}, the singular values of model weights~\cite{Svdiff_han2023svdiff}, and output projection weights~\cite{PersonalizedResiduals_ham2024personalized}.
Moreover, parameter-efficient tuning (PEFT) methods, such as LoRA~\cite{Lora_hu2021lora}, also play an important role in personalization.
Orthogonal finetuning methods~\cite{OFT_qiu2023controlling,BOFT_liu2023parameter} have also been proposed to improve the generalization capabilities.

Text alignment is another critical challenge in personalization tasks.
Numerous studies~\cite{Disenbooth_chen2023disenbooth,DETEX_cai2024decoupled,Dreamtuner_hua2023dreamtuner,BreakAScene_avrahami2023break,CompositionalInversion_zhang2024compositional,DreamBlend_ram2025dreamblend,Dreammatcher_nam2024dreammatcher} have been conducted to address this issue through disentanglement learning.
Among them, masks are widely adopted to minimize the effects of the background~\cite{BreakAScene_avrahami2023break,Disenbooth_chen2023disenbooth}.
Several methods~\cite{Disenbooth_chen2023disenbooth,DETEX_cai2024decoupled} introduce learnable embeddings to capture the irrelevant information, such as background and pose. 
By optimizing these embeddings alongside the subject embedding with designed losses, these methods can achieve more flexible pose and context control through text prompts. 
DreamTuner~\cite{Dreamtuner_hua2023dreamtuner} introduces ControlNet~\cite{Controlnet_zhang2023adding} to provide additional pose information, which helps decouple the pose from learn subject concept.
Additionally, several loss functions~\cite{Dreambooth_ruiz2023dreambooth,ComFusion_hong2024comfusion,Singleinsert_wu2023singleinsert,DCO_lee2024direct,CoRe_wu2024core,kim2024learning,Instructbooth_chae2023instructbooth,PALP_arar2024palp,FaceChainSuDe_qiao2024facechain} are employed to improve text alignment.
Dreambooth~\cite{Dreambooth_ruiz2023dreambooth} introduces a prior preservation loss that regularizes the class prior to match that of the pretrained model.
This approach employs a regularization dataset comprising images from a specific category to compute the loss.
ComFusion~\cite{ComFusion_hong2024comfusion} further extends it by preserving both class-specific and scene-specific knowledge from pretrained models through a class-scene prior loss.
FaceChain-SuDe~\cite{FaceChainSuDe_qiao2024facechain} introduces a subject-derived regularization, which helps the subject inherit the public attributes of its super-category while learning its private attributes.
InstructBooth~\cite{Instructbooth_chae2023instructbooth} incorporates reinforcement learning (RL) finetuning to enhance text alignment.
ImageReward~\cite{Imagereward_xu2024imagereward} is utilized to measure the alignment between generated images and text prompts.
PALP~\cite{PALP_arar2024palp} adopts a prompt-aligned score sampling to encourage the personalized model to keep the ability to generate images aligned with a specific prompt.

To improve the efficiency, learning-based methods~\cite{E4T_gal2023encoder,ELITE_wei2023elite,SubjectDiffusion_ma2024subject,UMM_ma2023unified,SUTI_chen2024subject,Taming_jia2023taming,IPAdapter_ye2023ip,Instantbooth_shi2024instantbooth,InstructImagen_hu2024instruct,SSREncoder_zhang2024ssr} employ encoders with optional adapters to inject target concept into generation process, offering a significant speed advantage.
Some works focus on domain-aware encoders specifically designed to encode images from target domains~\cite{E4T_gal2023encoder,Instantbooth_shi2024instantbooth,AnyDressing_li2024anydressing}.
For example, E4T~\cite{E4T_gal2023encoder} employs a time-aware encoder to predict the delta word embedding from the subject image.
It also updates the parameters of Unet model to adapt it to a specific domain, such as face or cat.
AnyDressing~\cite{AnyDressing_li2024anydressing} utilizes a Unet-based GarmentsNet to extract detailed features from garments, which are then injected into the diffusion process through the proposed dressing attention.
It also supports the customization of multiple garments simultaneously and generates the corresponding human images, demonstrating great potential for e-commerce applications.

In contrast to domain-specific personalization, other methods adopt a domain-agnostic approach, which train encoders on open-world images to extract more generalized conditions~\cite{ELITE_wei2023elite, UMM_ma2023unified,IPAdapter_ye2023ip,SubjectDiffusion_ma2024subject,Bootpig_purushwalkam2024bootpig,MOMA_song2025moma,Taming_jia2023taming}.
Among these, IP-Adapter~\cite{IPAdapter_ye2023ip} is a widely used technique.
It employs a pretrained CLIP image encoder to extract image features, subsequently projecting them into feature space as image embeddings.
To integrate these embeddings with text for personalized generation, IP-Adapter introduces additional key and value projections into the cross-attention layers for image attention calculation, which is called dual cross attention.
The text and image attentions are then combined to guide image generation.
Following a similar framework, various encoders and adapters have been explored to enhance the fidelity of the learned subject concepts, as discussed in Sec.~\ref{sec:diffusion_learning_methods}.
For example, with the development of vision language models (VLMs), several studies~\cite{BLIPDiffusion_li2024blip,UNIMOG_li2024unimo,CustomizationAssistant_zhou2024customization,CustomContrast_chen2024customcontrast,MOMA_song2025moma} leverage VLMs to encode images and text simultaneously, achieving better text controllability.
Additionally, some methods~\cite{UMM_ma2023unified,ELITE_wei2023elite,Instantbooth_shi2024instantbooth}  project images into text-related spaces to obtain textual features that can be directly utilized by T2I models without adapter training. 
Furthermore, several approaches~\cite{DAT_arar2023domain,Hyperdreambooth_ruiz2024hyperdreambooth} employ hypernetwork-like encoders, which predict weight deltas of T2I models for personalized generation.

To address the copy-paste problem, a common issue in learning-based methods, several studies~\cite{DisEnvisioner_he2024disenvisioner,song2024harmonizing,SAG_chan2024improving} have been conducted.
DisEnvisioner~\cite{DisEnvisioner_he2024disenvisioner} disentangles the features of the subject and other irrelevant components by projecting the image feature into two distinct and orthogonal tokens.
Then, the disentangled features are refined to produce identity consistent images that adhere to the input text.
Besides, several methods construct unpaired image datasets to address these challenges, as discussed in Sec.~\ref{sec:diffusion_learning_methods}.
Based on these datasets, more flexible personalization methods have been developed.
However, there is still a lack of large-scale, high-quality open-source datasets for future research.
Several approaches~\cite{song2024harmonizing,SAG_chan2024improving} enhance text editability through post-processing methods.
For example, during inference, Song~\etal~\cite{song2024harmonizing} adjust the visual embedding to be orthogonal to the textual embedding. 
This adjustment accurately guides the generation process in a direction that adheres to the text prompt.
SAG~\cite{SAG_chan2024improving} introduces dual classifier-free guidance to improve text alignment by attenuating the subject-aware condition.
Furthermore, several sampling techniques~\cite{Instantbooth_shi2024instantbooth,Fastcomposer_xiao2024fastcomposer,Infinite-ID_wu2025infinite,ELITE_wei2023elite} have been proposed to balance the text alignment and subject fidelity.

Various personalization frameworks have also been explored~\cite{JeDi_zeng2024jedi,cai2024diffusion,lambdaECLIPSE_patel2024lambda}.
Instead of encoding the subject image, JeDi~ \cite{JeDi_zeng2024jedi} treats personalized image generation as an image inpainting task.
It leverages reference subject images as examples and inpaints the output images based on target text.
Similarly, Diffusion Self-Distillation~\cite{cai2024diffusion} performs personalized image generation through image-guided video generation, where the given subject image serves as the first frame to guide the generation of subsequent frame.
Considering that the reference image may contain multiple concepts simultaneously, SSREncoder~\cite{SSREncoder_zhang2024ssr} utilizes a token-to-patch aligner to highlight selective regions in the reference image based on a given query.
During inference, users can flexibly select the target concept from the image using mask or text input.
Recently, several methods~ \cite{IPAdapterInstruct_rowles2024ipadapter, InstructImagen_hu2024instruct,OminiControl_tan2024ominicontrol,le2024diffusiongenerate,Omnigen_xiao2024omnigen} have been developed to train personalized image generation alongside other tasks, resulting in multi-conditional controllable generation techniques.
For example, OmniGen~\cite{Omnigen_xiao2024omnigen} jointly models text and images within a single framework and supports various downstream tasks, such as image editing, subject-driven generation, and visual conditional generation.

In addition to the aforementioned methods, several researchers have explored alternative training techniques.
Fei~\etal~\cite{GFTI_fei2023gradient} propose a gradient-free textual inversion method that optimizes word embeddings without accessing the gradients of the text-to-image model.
Similarly, PRISM~\cite{PRISM_he2024automated} introduces a black-box personalization approach that employs a vision-language model (VLM) to generate human-interpretable prompts for the desired concept. 
During the learning process, the VLM serves as both a prompt engineering assistant and a judge to iteratively adjust the prompts.
Ding~\etal~\cite{CLIPconverter_ding2024clip} utilize a linear projection to transform CLIP image features into text features, which can be sent to a T2I model directly to generate customized images.
Several methods~\cite{E4T_gal2023encoder,BLIPDiffusion_li2024blip,DAT_arar2023domain,Hyperdreambooth_ruiz2024hyperdreambooth,HybridBooth_guan2025hybridbooth} also employ the hybrid framework, which integrates the strengths of learning-based techniques with optimization-based refinement.

\subsubsection{Face-Driven Personalization}
\label{sec:diffusion_personalization_face}

Personalized face generation is specifically focused on creating
human-centric images that maintain the same identity as the individuals depicted in the reference images.
This task can be considered a specialized case of personalized subject generation, and the methods mentioned in Sec.~\ref{sec:diffusion_personalization_subject} are also applicable to this task.
Compared with general subject domain, face domain have been widely studied in previous researches, such as face generation~\cite{StyleGAN_karras2019style}, face editing~\cite{Deltaedit_lyu2023deltaedit}, and face analysis~\cite{EasyPortrait_kapitanov2023easyportrait,PFLD_guo2019pfld}.
In this section, we focus on techniques that are specifically designed for face-driven personalization.

Similar to subject-driven personalization, several face personalization methods~\cite{ProFusion_zhou2023enhancing,SeFi-IDE_li2024sefi,CelebBasis_yuan2023inserting,CrossInitialization_pang2024cross,Stableidentity_wang2024stableidentity,ID-Booth_tomavsevicid} employ an optimization-based framework to project a given face image into textual or parameter space.
Among them, CelebBasis~\cite{CelebBasis_yuan2023inserting} leverages the existing celebrity knowledge within pretrained T2I models, and projects the embeddings of well-known names to construct the celeb basis.
Based on the celeb basis, the model can represent a new identity with 1,024 learnable coefficients, enabling flexible controllability on learned faces.
Following it, StableIdentity~\cite{Stableidentity_wang2024stableidentity} uses an identity encoder and a multilayer perceptron to extract the identity representation and projects it into the constructed celeb embedding space.

In contrast to optimization-based personalization, learning-based framework is widely adopted in face-driven generation methods~\cite{Fastcomposer_xiao2024fastcomposer,Dense-Face_guo2024dense,PersonaMagic_li2024personamagic,Omni-ID_qian2024omni,RealisID_sun2024realisid,PersonaHOI_hu2025personahoi,UniPortrait_he2024uniportrait,IC-Portrait_yang2025ic,CharacterFactory_wang2024characterfactory}.
For example, FastComposer~\cite{Fastcomposer_xiao2024fastcomposer} employs an image encoder to extract subject embeddings and utilizes a multilayer perceptron (MLP) to fuse them with person-related text features (\eg, man or woman) to inject identity information.
During inference, a delayed identity conditioning mechanism is introduced to balance identity consistency and text controllability.
To improve face identity, several methods~\cite{Face0_valevski2023face0,Dreamidentity_chen2023dreamidentity,Facestudio_yan2023facestudio,Portraitbooth_peng2024portraitbooth,FACT_yu2024facechain} propose to employ face recognition models~\cite{Arcface_deng2019arcface,TFace_huang2020curricularface} to extract identity-related information.
Face2Diffusion~\cite{Face2Diffusion_shiohara2024face2diffusion} retrains the face recognition model to remove identity-irrelevant information from the extracted identity embeddings.
Identity loss is also widely adopted to improve identity consistency~\cite{Photoverse_chen2023photoverse,Pulid_guo2024pulid,Magicapture_hyung2024magicapture,Portraitbooth_peng2024portraitbooth,Lcm-lookahead_gal2024lcm,ID-Aligner_chen2024id}.
Additionally, inspired by the attributes controllability in StyleGANs, several methods~\cite{wplus_li2024stylegan,Precisecontrol_parihar2025precisecontrol} propose adapting the $\mathcal{W}+$ space to text-to-image models for personalized face generation.
This adaptation allows for more precise manipulation of facial attributes based on latent editing, such as eye size and age (as discussed in Sec.~\ref{sec:gan_personalization}).
Moreover, DiffLoRA~\cite{Difflora_wu2024difflora} utilizes a diffusion model to predict identity-specific LoRA weights from reference images, which are then merged into SDXL for inference.
It leverages LoRA's custom generation capabilities, while avoiding the complex optimization process.

Similar to subject-driven generation, the copy-paste problem poses a significant challenge for learning-based methods.
Several methods~\cite{photomaker_li2024photomaker,FlashFace_zhang2024flashface,IDAdapter_cui2024idadapter,Infinite-ID_wu2025infinite} tackle the copy-paste problem from the \textit{dataset perspective}.
For example, PhotoMaker~\cite{photomaker_li2024photomaker} collects multiple images of the same identity and constructs an unpaired dataset for training, resulting in superior text editability. 
Additionally, it stacks face embeddings from multiple images to enhance identity fidelity.
In addition to collecting datasets, several methods~\cite{MasterWeaver_wei2025masterweaver,Imagine-yourself_he2024imagine,Dreamidentity_chen2023dreamidentity} utilize generated unpaired images.
For example, DreamIdentity~\cite{Dreamidentity_chen2023dreamidentity} leverages the celebrity knowledge from pretrained T2I models to generate different images of celebrities, thereby forming an unpaired training dataset.
MasterWeaver~\cite{MasterWeaver_wei2025masterweaver} utilizes face editing models~\cite{Deltaedit_lyu2023deltaedit} to augment the facial attributes of the reference face image.
From the \textit{training perspective}, several approaches~\cite{MasterWeaver_wei2025masterweaver,Lcm-lookahead_gal2024lcm,Pulid_guo2024pulid} introduce specialized loss functions to address the copy-paste problem.
LCM-Lookahead~\cite{Lcm-lookahead_gal2024lcm} employs the LCM model~\cite{LCM_luo2023latent} to generate an image in several steps, which is used to calculate a text alignment loss, thereby improving text controllability.
PuLID~\cite{Pulid_guo2024pulid} introduces a semantic alignment loss between the identity condition branch and the pure text branch to encourage the text controllability of personalization model is aligned with original T2I model.
Various approaches~\cite{Face-diffuser_wang2024high,FreeCure_cai2024foundation,Infinite-ID_wu2025infinite,Fastcomposer_xiao2024fastcomposer} address the problem from the \textit{sampling perspective}. 
For example, Face-diffuser~\cite{Face-diffuser_wang2024high} employs two independent models for scene and character generation, and introduces a saliency-adaptive noise fusion mechanism to automatically blend the noise from both models during the generation process.
Infinite-ID~\cite{Infinite-ID_wu2025infinite} utilizes decoupled cross-attention and replaces self-attention with a mixed attention mechanism during the inference stage.

Facial analysis is an extensively researched area in deep learning, and numerous tools have been developed, such as face attribute classification, face detection, and face parsing.
Therefore, several methods incorporate these face conditions to provide additional controllability~\cite{Portraitbooth_peng2024portraitbooth,identity-expression_liu2024towards,EmojiDiff_jiang2024emojidiff,Consistentid_huang2024consistentid,MagicID_deng2024magicid,Diff-PC_xu2024diff,Caphuman_liang2024caphuman}.
For example, methods~\cite{Portraitbooth_peng2024portraitbooth,identity-expression_liu2024towards,EmojiDiff_jiang2024emojidiff} like PortraitBooth~\cite{Portraitbooth_peng2024portraitbooth} incorporate expression conditions into the generation process to control the facial expressions of the generated images.
Based on the face parsing map, ConsistentID~\cite{Consistentid_huang2024consistentid} employs a fine-grained feature extractor to extract detailed multimodal facial features from different parts of the face, which are used to improve identity fidelity.
Additionally, methods~\cite{Instantid_wang2024instantid,MagicID_deng2024magicid} such as InstantID~\cite{Instantid_wang2024instantid} integrate face landmarks with additional ControlNet to provide spatial information, enhancing the controllability of the generation process.
CapHuman~\cite{Caphuman_liang2024caphuman} introduces a 3D parametric face model to provide structure and pose conditions for face generation, offering a more flexible and fine-grained head control.

\subsubsection{Character-Driven Personalization}
\label{sec:diffusion_personalization_character}

Personalized character generation further ensures consistency in both the identity and the body during the personalization.

To ensure body consistency, some methods~\cite{StoryMaker_zhou2024storymaker,UniHuman_li2024unihuman,huang2024parts,Character-Adapter_ma2024character,RETRIBOORU_tang2023retrieving} employ additional encoders to process the face and body separately and optimize them simultaneously.
For instance, StoryMaker~\cite{StoryMaker_zhou2024storymaker} employs two independent encoders to extract face and body embeddings from masked face and body images.
To disentangle irrelevant pose and background information, it further extracts pose and background embeddings and incorporates them with character embeddings for disentanglement learning.
During inference, users can optionally provide reference keypoints to control the pose of the generated character.
Besides, several methods~\cite{Character-Adapter_ma2024character,RETRIBOORU_tang2023retrieving} process the upper and lower body separately, allowing for the flexible generation of images by composing parts from different sources.
Recently, some approaches~\cite{SerialGen_xie2024serialgen,AnyStory_he2025anystory} encode the character image as a single character condition for personalization, similar to subject and face personalization. 
For example, AnyStory~\cite{AnyStory_he2025anystory} employs a ReferenceNet combined with a CLIP image encoder to faithfully encode the character information. 
It further introduces an instance-aware subject router to encourage the generation of multiple characters without interfering with each other.
SerialGen~\cite{SerialGen_xie2024serialgen} first utilizes a model to project the given character image as a standardized reference.
During personalization, this standardized reference is used to provide the character information, offering high text controllability and appearance consistency.

\subsubsection{Style-Driven Personalization}
\label{sec:diffusion_personalization_style}

Style-driven personalization aims to generate images with style as indicated by reference images, while ensuring that the content aligns with the given text prompt. 
The style of an image typically encompasses a variety of artistic elements, including colors, textures, brush strokes, \etc.

Numerous studies~\cite{InST_zhang2023inversion,Prospect_zhang2023prospect,StyleBoost_park2023styleboost,StyleForge_park2024text,PairCustomization_jones2024customizing,U-VAP_wu2024u,Style-friendly_choi2024style,inspirationtree_vinker2023concept,MATTE_agarwal2023image,Ziplora_shah2025ziplora} have employed optimization-based methods for style personalization.
For example, Textual Inversion~\cite{TI_gal2022image} and subsequent methods~\cite{InST_zhang2023inversion,Dreamartist_dong2022dreamartist,Prospect_zhang2023prospect} simply utilize learned word embeddings to capture the style of reference images. 
During inference, these methods employ prompts structured like ``[content] in the style of S*'' for style-driven personalization, where S* denotes the learned style embedding and [content] is the content-related text.
%
%
To facilitate accurate style learning, U-VAP~\cite{U-VAP_wu2024u} leverages a large language model to generate sentences describing the desired attributes and unrelated attributes, separately.
Based on these sentences, it learns both target and non-target material word embeddings, which can be fused for decoupled target style generation.

Several methods~\cite{StyleBoost_park2023styleboost,StyleForge_park2024text,PairCustomization_jones2024customizing,break-for-make_xu2024break,UnZipLoRA_liu2024unziplora,Ziplora_shah2025ziplora} focus on finetuning model parameters to capture specific styles.
Since the given images contain both subject and style, the learned style concept is easily entangled with irrelevant content information, resulting unsatisfactory generation results.
To address this issue, various methods ~\cite{PairCustomization_jones2024customizing,UnZipLoRA_liu2024unziplora} adopt an additional content LoRA to capture content-specific information, thereby decoupling style from content.
For instance, PairCustomization~\cite{PairCustomization_jones2024customizing} learns the style and content LoRAs from a pair of (content, style) images, where the style image and content image share the same content but differ in style.
During training, the content LoRA is exclusively trained with the content image to capture the content information.
By jointing optimizing the style LoRA and content LoRA, it effectively disentangles content information from style learning.
UnZipLoRA~\cite{UnZipLoRA_liu2024unziplora} employs three types of prompts, \ie, mixed prompts, individual style prompts, and individual content prompts, to help learn decoupled content and style.
During training, each type of prompt is assigned its own LoRA, and an orthogonal loss is applied among the style and content LoRAs to disentangle the style from the content.

For efficiency, several approaches~\cite{StyleAdapter_wang2024styleadapter,IPAdapterInstruct_rowles2024ipadapter,ArtAdapter_chen2024artadapter,zhuoqi2024content,FineStyle_zhangfinestyle,Artistic-Intelligence_yang2024artistic,CSGO_imagecsgo} utilize learning-based frameworks that employ encoders to extract style-related conditions.
Since there is a lack of datasets consisting of content-style-stylized image pairs, these methods typically use existing style images for training. 
For example, StyleAdapter~\cite{StyleAdapter_wang2024styleadapter} leverages a CLIP image encoder to extract style information from multiple style references.
To decouple the content information from the style embeddings, it shuffles the patch-based vision embeddings and removes the original class embedding.
The extracted style embeddings are then injected through a dual cross-attention mechanism~\cite{IPAdapter_ye2023ip}.
Instead, ArtAdapter~\cite{ArtAdapter_chen2024artadapter} introduces an Auxiliary Content Adapter (ACA), which uses an augmented image to provide weak content guidance during training, enhancing the accuracy of style learning.
Based on B-LoRA~\cite{BLoRA_frenkel2025implicit}, CSGO~\cite{CSGO_imagecsgo} constructs a dataset consists of content-style-stylized image pairs. 
Using this dataset, CSGO employs the content image with ControlNet to provide structural information, effectively preventing the content image from leaking style information.
Additionally, InstantStyle~\cite{Instantstyle_wang2024instantstyle} employs the IP Adapter for personalized style generation by injecting style features selectively into specific cross-attention layers.

Several studies~\cite{Aesthetic-Gradients_gallego2022personalizing,Stylealigned_hertz2024style,SAG_pan2023towards,FreeTuner_xu2024freetuner,RB-Modulation_rout2024rb,Diptych-Prompting_shin2024large} have also explored training-free frameworks to efficiently apply desired styles from reference images to image generation.
For example, StyleAligned~\cite{Stylealigned_hertz2024style} utilizes an inversion method to extract style features from the reference image.
During the image generation process, it encourages the generated image to maintain target styles by utilizing a shared attention mechanism. 
Additionally, some methods~\cite{Aesthetic-Gradients_gallego2022personalizing,SAG_pan2023towards,FreeTuner_xu2024freetuner} adopt style-guided sampling to achieve personalized style generation.
For example, SAG~\cite{SAG_pan2023towards} calculates a style loss between the estimated clean image and the style image using a CLIP image encoder.
This loss is then used to guide the sampling process through gradient guidance, ensuring that the style of the generated image aligns with that of the reference image.

\subsubsection{High-level Semantic Personalization}
\label{sec:diffusion_personalization_semantic}

Personalized image generation has also been expanded to include the generation of specific semantic relations or actions, which we refer to as high-level semantic personalization.

Given several exemplar images, Reversion~\cite{ReVersion_huang2024reversion} aims to extract the common relation among them, such as ``A \textless R\textgreater \ B'' and \textless R\textgreater \ can represent actions like ``shakes hands'' or relations like ``is painted on''.
Specifically, Reversion introduces a relation steering contrastive learning to guide relation embeddings toward proper prepositions, facilitating the training process.
%
%
Lego~\cite{Lego_motamed2023lego} focuses on learning more general concepts, such as adjectives and verbs, which are often intertwined with the subject's appearance. 
To better disentangle the concept from the subject, Lego learns the subject embedding from subject-only images, while learning the concept jointly with subject in a contrastive setting.

ADI~\cite{ADI_huang2024learning} proposes a method to learn the customized action from limited data.
To disentangle action-agnostic information from the learned action,
it extracts the gradient invariance from the constructed sample triples and masks the updates of irrelevant channels.
ImPoster~\cite{ImPoster_kothandaraman2024imposter} further combines the subject-driven generation and action-driven generation.
Given a single ``source'' image and a ``driving'' image along with their corresponding text descriptions, ImPoster could generate an image of the source subject performing the driving actions. 
%
%

FreeEvent~\cite{FreeEvent_wang2024event} introduces the event-customized image generation, which aims to learn all the actions, poses, relations, or interactions between different entities in the reference image. 
To achieve this, FreeEvent incorporates two additional generation paths.
The entity switching path applies cross-attention guidance and regulation for target entity generation.
The event transferring path injects spatial features and self-attention maps from the reference image into the target image for event generation. 
This approach enables FreeEvent to accurately capture complex events and generates customized images with various target entities.

\subsubsection{Multiple Concept Personalization}
\label{sec:diffusion_personalization_multiple}

Multiple concept personalization aims to generate images that incorporate multiple customized concepts, such as  ``a dog* and a cat*'', where dog* and cat* are user-indicated concepts.

A straightforward approach for multiple concept personalization is learning these concepts jointly using images that contain all the desired concepts together. 
However, collecting such images requires human labeling.
To address this, SVDiff~\cite{Svdiff_han2023svdiff} employs a simple technique called Cut-Mix-Unmix to create training images composing multiple objects.
Using the constructed training dataset, SVDiff further trains a personalization model to learn multiple concepts simultaneously.
With the guidance of part masks, MagicTailor~\cite{MagicTailor_zhou2024magictailor} learns to compose multiple parts form different faces.
In contrast, CIDM~\cite{CIDM_dong2024continually} treats the task as a continual learning problem.
During training, it introduces a concept consolidation loss along with an elastic weight aggregation module to mitigate the forgetting of previously learned concepts.

Additionally, several methods have been proposed to combine separately trained models for multiple concept personalization.
Several \emph{weight fusion} methods have been proposed~\cite{CD_kumari2023multi,Mix-of-show_gu2024mix,Ziplora_shah2025ziplora,Cones_liu2023cones,Ortha_po2024orthogonal,LoRA.rar_shenaj2024lora,Block-wise-LoRA_li2024block,MuDI_jang2024identity}.
For example, Custom Diffusion~\cite{CD_kumari2023multi} proposes to merge finetuned key and value matrices through solving a constrained optimization method.
Po~\etal~\cite{Ortha_po2024orthogonal} propose a technique that encourages customized models, which do not have access to each other during fine-tuning, to have orthogonal residual weights.
During inference, these customized models can be summed with minimal interference, facilitating multiple concept generation.
Furthermore, several methods~\cite{MC2_jiang2024mc,Omg_kong2025omg,ConceptWeaver_kwon2024concept} propose merging multiple concepts through \emph{noise fusion}. 
OMG~\cite{Omg_kong2025omg} and ConceptWeaver~\cite{ConceptWeaver_kwon2024concept} first generate a template image to provide structural information, and then regenerate the multi-concept image by mixing multiple score estimation outputs.
Other approaches also utilize attention maps to refine the generation of multiple concepts~\cite{MC2_jiang2024mc,SEGuidance_liu2024training,Cones2_liu2024customizable,IR-Diffusion_he2024improving,DisenDiff_zhang2024attention}.
For example, SEGuidance~\cite{SEGuidance_liu2024training} employs a backward loss on the attention maps to ensure that there exists at least one patch in the attention map with a high attention value for each concept, which facilitates the generation of each concept.

To address location conflicts among different concepts, additional layout inputs (\eg, bounding boxes or keypoints) are usually employed~\cite{Cones2_liu2024customizable,LoRAComposer_yang2024lora,ConceptConductor_yao2024concept,jain2024multi,Autostory_wang2023autostory}.
For example, Cones2~\cite{Cones2_liu2024customizable} utilizes layout guidance to strengthen the signal of the target concept while weakening the signals of irrelevant concepts. 
ConceptConductor~\cite{ConceptConductor_yao2024concept} uses a reference image to provide layout information. 
During generation, the output features from the attention layers of different concept models are multiplied by their corresponding masks and summed to obtain a fused feature map.
For learning-based methods~\cite{SubjectDiffusion_ma2024subject,GroundingBooth_xiong2024groundingbooth,MS-Diffusion_personalizationms,RelationBooth_shi2024relationbooth,TokenVerse_garibi2025tokenVerse,parmar2025object}, the location information are usually encoded and injected through adapters.
For example,  GroundingBooth~\cite{GroundingBooth_xiong2024groundingbooth} employs a grounding input module that takes layout and images to extract grounding tokens, which are then injected through gated self-attention.
%

\begin{figure*}[t]
\centering
\includegraphics[width=1\linewidth]{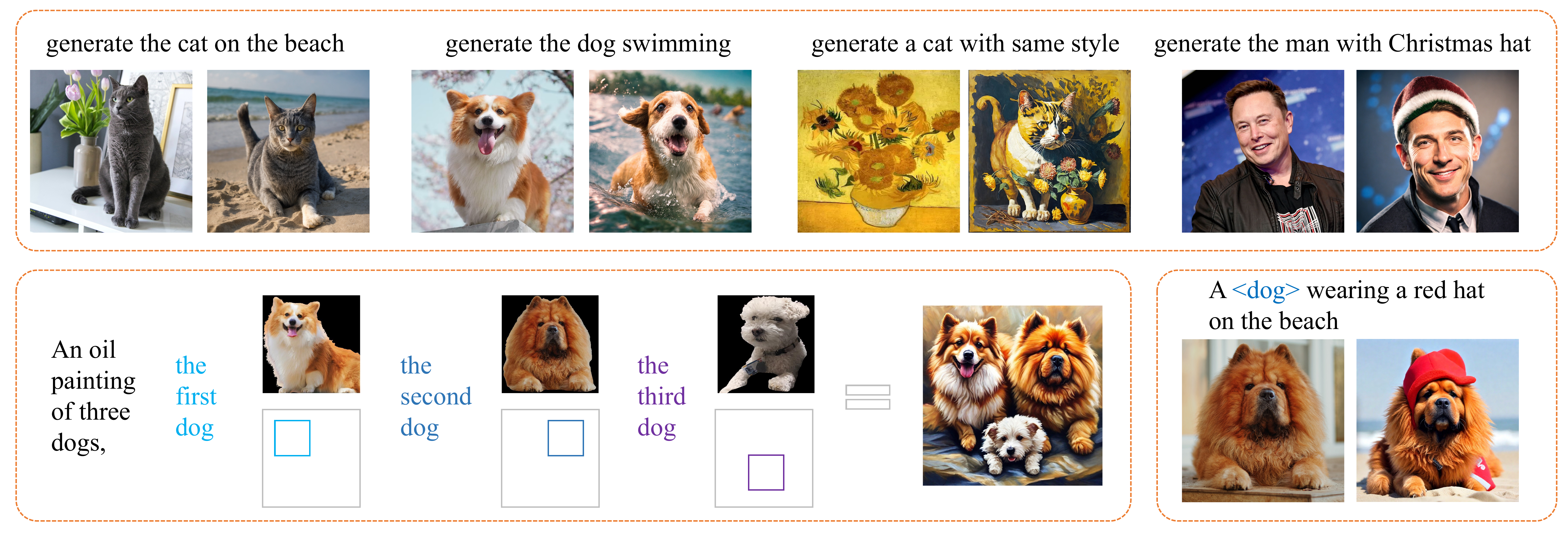}
\caption{\textbf{Visual generation results of AR-based personalization methods.} Images generated by Seed-X~\cite{SeedX_ge2024seed} and Emu2-Gen~\cite{Emu2_sun2024generative}.}
\label{fig:ar_generation}
\vspace{-1em}
\end{figure*}

\subsection{Text-driven Image Editing}
\label{sec:diffusion_text_editing}

Similar to GANs, with the inverted noise of given concept image, text-guided image editing is another possible way to create personalized content, as illustrated in Fig.~\ref{fig:personalization_methods} (c).
Here, we also give a brief introduction to several representative text-based image editing techniques~\cite{GANTASTIC_dalva2024gantastic,DDS_hertz2023delta,prompt2prompt_hertz2022prompt,Ledits++_brack2024ledits++,Masactrl_cao2023masactrl,StableFlow_avrahami2024stable}.
For example, DDS~\cite{DDS_hertz2023delta} modifies the inverted noise to perform editing.
After inverting the concept image into the noise space, it employs a text-based image editing scoring function to guide noise towards the intended direction of text descriptions.
Different from GANs, text-driven editing methods for T2I diffusion models can be performed in a training-free framework.
For instance, Prompt-to-prompt~\cite{prompt2prompt_hertz2022prompt} investigates the attention mechanism in diffusion models for image editing. 
Specifically, it inverts the given image into the noise space based on the source prompt and then uses it to generate an edited image with a new editing prompt.
To ensure accurate editing while keeping irrelevant parts unchanged, it leverages the attention maps from the reconstruction path to replace the corresponding maps in the editing path.
LEDITS++~\cite{Ledits++_brack2024ledits++} edits the image by manipulating the noise estimation during generation based on a set of edit instructions.
DDPM inversion is employed to represent the image as a sequence of noises.
However, these editing techniques often yield unsatisfactory results for non-rigid edits, such as pose editing.
To address this, MasaCtrl~\cite{Masactrl_cao2023masactrl} introduces a mutual self-attention mechanism, where the key and value in the editing path are derived from the reconstruction path.
By querying the reference context, MasaCtrl maintains the object's appearance while adhering to non-rigid guidance specified in the edit instructions, such as changes in pose or action.

Recently, based on advanced T2I models, such as Stable Diffusion 3~\cite{SD3_esser2024scaling} or FLUX~\cite{flux2023}, several methods have been proposed to achieve more flexible editing~\cite{StableFlow_avrahami2024stable,HeadRouter_xu2024headrouter,xu2024unveil}.
For example, Stable Flow~\cite{StableFlow_avrahami2024stable} identifies the vital layers in DiT-based diffusion models that are crucial for image formation during generation. 
During generation, it injects the editing instructions only into these vital layers, yielding a stable edit for both attribute and non-rigid editing.
HeadRouter~\cite{HeadRouter_xu2024headrouter} investigates the sensitivity of various attention heads to different image semantics within DiTs, and proposes to perform editing by adaptively routing text guidance to the appropriate attention heads during generation.

\begin{table*}[t]
    \centering
    \caption{Summary of optimization-based personalized image generation methods with diffusion models.}
    \label{tab:duffision_opt}
    \setlength{\tabcolsep}{12mm}
    \resizebox{1\linewidth}{!}{
    \begin{tabular}{l|ccc}
    \hline
    Method & Category & Inversion Space & Backbone  \\ \hline
    Dreambooth~\cite{Dreambooth_ruiz2023dreambooth} & Subject & Parameter & Imagen / SD \\
    TI~\cite{TI_gal2022image} & Subject & Textual (Word) & LDM / SD \\
    Aesthetic Gradients~\cite{Aesthetic-Gradients_gallego2022personalizing} & Style & Parameter & SD \\
    ARLDM~\cite{ARLDM_pan2024synthesizing} & Character & Parameter & SD \\
    DreamArtist~\cite{Dreamartist_dong2022dreamartist} & Subject & Textual (Word) & SD 1.4 \\
    InST~\cite{InST_zhang2023inversion} & Style & Textual (Word) & SD \\
    Custom Diffusion~\cite{CD_kumari2023multi} &  Multiple Subjects & Textual (Word) + Parameter & SD 1.4 \\
    Cones~\cite{Cones_liu2023cones} &  Subject & Parameter & SD 1.4 \\
    HiPer~\cite{HiPer_han2023highly} & Subject & Textual (Output) & SD 1.4 \\
    P+~\cite{P+_voynov2023p+} & Subject & Textual ($\mathcal{P}$+) & SD \\
    Reversion~\cite{ReVersion_huang2024reversion} & Relation & Textual (Word) & SD 1.4 \\
    SVDiff~\cite{Svdiff_han2023svdiff} & Multiple Subjects & Parameter & SD \\
    Fei~\etal~\cite{GFTI_fei2023gradient} & Subject & Textual (Word) & SD 1.4 \\
    Break-A-Scene~\cite{BreakAScene_avrahami2023break} & Subject & Textual (Word) + Parameter & SD 1.4 \\
    Cones2~\cite{Cones2_liu2024customizable} & Multiple Subjects & Textual (Output) & SD 1.4 \\
    Disenbooth~\cite{Disenbooth_chen2023disenbooth} & Subject & Textual (Word) & SD 1.4 \\
    Inspirationtree~\cite{inspirationtree_vinker2023concept} & Subject + Style & Textual (Word) & SD 1.4 \\
    Mix-of-show~\cite{Mix-of-show_gu2024mix} & Multiple Subjects & Parameter & SD \\
    NeTI~\cite{NeTI_alaluf2023neural} & Subject & Textual ($\mathcal{P}$*) & SD 1.4 \\
    Perfusion~\cite{Perfusion_tewel2023key} & Subject & Parameter & SD 1.5 \\
    ProFusion~\cite{ProFusion_zhou2023enhancing} & Face & Textual (Word) & SD 2.0 \\
    Prospect~\cite{Prospect_zhang2023prospect} & Style + Subject & Textual (Word) & SD 1.4 \\
    CelebBasis~\cite{CelebBasis_yuan2023inserting} & Face & Textual (Celeb Space) & SD 1.4 \\
    OFT~\cite{OFT_qiu2023controlling} & Subject & Parameter & SD 1.5 \\
    ViCo~\cite{ViCo_hao2023vico} & Subject & Textual (Word) + Feature & SD 1.4 \\
    Hyperdreambooth~\cite{Hyperdreambooth_ruiz2024hyperdreambooth} & Face & Parameter & SD 1.5 \\
    MagiCapture~\cite{Magicapture_hyung2024magicapture} & Face + Style & Textual (Word) & SD 1.5 \\
    MCPL~\cite{MCPL_jin2023image} & Multiple Subjects & Textual (Word) & LDM \\
    SingleInsert~\cite{Singleinsert_wu2023singleinsert} & Subject & Textual (Word) + Parameter & SD 1.5 \\
    StyleBoost~\cite{StyleBoost_park2023styleboost} & Style & Parameter & SD 1.5 \\
    ADI~\cite{ADI_huang2024learning} & Action & Textual (Word) & SD 2.1 \\
    Autostory~\cite{Autostory_wang2023autostory}  & Multiple Subjects & Parameter & SD \\
    BOFT~\cite{BOFT_liu2023parameter} & Subject & Parameter & SD 1.4 \\
    CatVersion~\cite{Catversion_zhao2023catversion} & Subject & Textual (Layer) & SD 1.5 \\
    He~\etal~\cite{he2023data} & Subject & Parameter & SD / SDXL \\
    HiFiTuner~\cite{HiFiTuner_wang2023hifi} & Subject & Parameter & SD 1.4 \\
    Lego~\cite{Lego_motamed2023lego} &  General Concepts (Adjectives/Verbs)  & Textual (Word) & SD \\
    MATTE~\cite{MATTE_agarwal2023image} & Subject + Style & Textual (Word) & SD \\
    ZipLoRA~\cite{Ziplora_shah2025ziplora} & Subject + Style & Parameter & SD \\
    Compositional Inversion~\cite{CompositionalInversion_zhang2024compositional} & Subject & Textual (Word) & SD 1.4 \\
    DETEX~\cite{DETEX_cai2024decoupled} & Subject & Textual (Word) + Parameter & SD 1.4 \\
    DreamTuner~\cite{Dreamtuner_hua2023dreamtuner} & Subject & Parameter+ Feature & SD 1.4 \\
    InstructBooth~\cite{Instructbooth_chae2023instructbooth} & Subject & Parameter & SD 1.5 \\
    Po~\etal~\cite{Ortha_po2024orthogonal} & Multiple Subjects & Parameter & ChilloutMix \\
    Lu~\etal~\cite{lu2024object} & Subject & Textual (Word) + Parameter & SD 1.4 \\
    PALP~\cite{PALP_arar2024palp} & Subject & Textual (Word) + Parameter & SD 1.4 \\
    Ryu~\cite{ryu2024memory} & Subject & Parameter & Quantized SD \\
    SeFi-IDE~\cite{SeFi-IDE_li2024sefi} & Face & Textual (Word) & SD 1.4 \\
    Stableidentity~\cite{Stableidentity_wang2024stableidentity} & Face & Textual (Word) & SD 2.1 \\
    ComFusion~\cite{ComFusion_hong2024comfusion} & Subject & Textual (Word) + Parameter & SD 1.5 \\
    DCO~\cite{DCO_lee2024direct} & Subject & Parameter & SDXL \\
    Block-wise LoRA~\cite{Block-wise-LoRA_li2024block} & Face + Style & Parameter & SD 1.4 \\
    B-LoRA~\cite{BLoRA_frenkel2025implicit} & Style & Parameter & SDXL \\
    Break-for-make~\cite{break-for-make_xu2024break} & Subject + Style & Parameter & SDXL \\
    DisenDiff~\cite{DisenDiff_zhang2024attention} & Multiple Subjects & Textual (Word) + Parameter & SD 1.4\\
    FaceChain-SuDe~\cite{FaceChainSuDe_qiao2024facechain} & Subject & Textual (Word) + Parameter& SD \\
    LoRAComposer~\cite{LoRAComposer_yang2024lora} & Multiple Concepts & Parameter & SD 1.4 \\
    PRISM~\cite{PRISM_he2024automated} & Subject & Text & SD 2.1 / DALLE2 / DALLE3 \\
    U-VAP~\cite{U-VAP_wu2024u} & Style & Textual (Word) & SD 1.5 \\
    ConceptWeaver~\cite{ConceptWeaver_kwon2024concept} & Multiple Subjects & Textual (Word) + Parameter & SD 2.1\\
    ID-Aligner~\cite{ID-Aligner_chen2024id} & Face & Parameter & SD 1.5 / SDXL \\
    MC2~\cite{MC2_jiang2024mc} & Multiple Subjects & Parameter & SD 1.5 \\
    MuDI~\cite{MuDI_jang2024identity} & Multiple Subjects & Parameter & SDXL \\
    StyleForge~\cite{StyleForge_park2024text} & Style & Parameter & SD 1.5 \\
    CLIF~\cite{CLIF_lin2024non} & Subject & Parameter & SD \\
    MCP~\cite{jain2024multi} & Multiple Subjects & Parameter & SD \\
    PairCustomization~\cite{PairCustomization_jones2024customizing} & Style + Subject & Parameter & SDXL \\
    Personalized Residuals~\cite{PersonalizedResiduals_ham2024personalized} & Subject & Parameter & SD 1.4 \\
    SAG~\cite{SAG_chan2024improving} & Subject & Textual (Word) & LDM \\
    Cross Initialization~\cite{CrossInitialization_pang2024cross} & Face & Textual (Word) & SD 2.1 \\
    BRAT~\cite{BRAT_baker2024brat} & Subject & Textual (Word) & SD 1.4 \\
    ConceptConductor~\cite{ConceptConductor_yao2024concept} & Multiple Subjects  & Textual ($\mathcal{P}$+) + Parameter & SD 1.5 \\
    CoRe~\cite{CoRe_wu2024core} & Subject & Textual (Word) & SD 1.4 \\
    ArtiFade~\cite{ArtiFade_yang2024artifade} & Subject & Textual (Word) + Parameter & LDM \\
    ID-Booth~\cite{ID-Booth_tomavsevicid} & Face & Parameter & SD 2.1 / SDXL \\
    ImPoster~\cite{ImPoster_kothandaraman2024imposter} & Subject + Action & Parameter & SD 2.1 \\
    TextBoost~\cite{TextBoost_park2024textboost} & Subject & Textual (Output) & SD 1.5 \\
    CIDM~\cite{CIDM_dong2024continually} & Multiple Subjects & Parameter & SD 1.5 / SDXL \\
    Cusconcept~\cite{Cusconcept_xu2024cusconcept} & Subject + Attribute & Textual (Word) & SD 2.1 \\
    Kim~\etal~\cite{kim2024learning} & Subject & - & SD  \\
    MagicTailor~\cite{MagicTailor_zhou2024magictailor} & Subject + Part &  Parameter & SD 2.1 \\
    DreamBlend~\cite{DreamBlend_ram2025dreamblend} & Subject & Parameter & SD \\
    Style-friendly~\cite{Style-friendly_choi2024style} & Style & Parameter & FLUX / SD 3.5 \\
    DCI-ICO~\cite{DCI-ICO_jin2025customized} & Subject & Textual (Word) + Parameter & SD 1.4 \\
    LoRA.rar~\cite{LoRA.rar_shenaj2024lora} & Style + Subject & Parameter & SDXL \\
    UnZipLoRA~\cite{UnZipLoRA_liu2024unziplora} & Style + Subject & Parameter & SDXL \\
    TokenVerse~\cite{TokenVerse_garibi2025tokenVerse} & Multiple Subjects & Parameter & FLUX \\
    \hline
    \end{tabular}
    }
\end{table*}

\begin{table*}[t]
    \centering
    \caption{Summary of learning-based personalized image generation methods with diffusion models.}
    \label{tab:duffision_learning}
    \setlength{\tabcolsep}{8mm}
    \resizebox{1\linewidth}{!}{
    \begin{tabular}{l|ccccc}
    \hline
    Method & Category & Inversion Space & Encoder & Adapter & Backbone  \\ \hline
    E4T~\cite{E4T_gal2023encoder} & Face / Cat & Textual (Word) & CLIP + SD Unet & - & SD 1.4  \\
    ELITE~\cite{ELITE_wei2023elite} & Subject & Textual (Word) + Feature & CLIP & DCA & SD 1.4 \\
    UMM~\cite{UMM_ma2023unified} & Subject & Textual (Word) & CLIP & - & SD 1.5 \\
    InstantBooth~\cite{Instantbooth_shi2024instantbooth} & Face / Cat & Textual (Output) + Feature & CLIP & SA & SD 1.4 \\
    SUTI~\cite{SUTI_chen2024subject} & Subject & Feature & Imagen Encoder & CA & Imagen   \\
    Taming~\cite{Taming_jia2023taming} & Subject & Feature & CLIP & CA & Imagen \\
    BLIP-Diffusion~\cite{BLIPDiffusion_li2024blip} & Subject & Textual (Word) & BLIP-2 Encoder & - & SD 1.5 \\
    FastComposer~\cite{Fastcomposer_xiao2024fastcomposer} & Face & Textual (Word) & CLIP & - & SD 1.5 \\
    Face0~\cite{Face0_valevski2023face0} & Face & Textual (Word) & Inception Resnet V1 & - & SD 1.4 \\
    AnyDoor~\cite{Anydoor_chen2024anydoor} & Subject & Textual + Feature & DINO & Concat & SD 2.1 \\
    Domain Agnostic~\cite{DAT_arar2023domain} & Subject & Textual (Word) + Parameter & CLIP + SD Unet & - & SD 1.4 \\
    DreamIdentity~\cite{Dreamidentity_chen2023dreamidentity} & Face & Textual (Word) & ID Encoder & - & SD 2.1 \\
    HyperDreambooth~\cite{Hyperdreambooth_ruiz2024hyperdreambooth} & Face & Parameter & ViT & - & SD 1.5 \\
    SubjectDiffusion~\cite{SubjectDiffusion_ma2024subject} & Multiple Subjects & Textual (Word) + Feature & CLIP & SA &  SD 2 \\
    IP-Adapter~\cite{IPAdapter_ye2023ip} & Subject & Feature & CLIP & DCA & SD 1.5 \\
    PhotoVerse~\cite{Photoverse_chen2023photoverse} & Face & Textual (Word) + Feature & CLIP & DCA & SD \\
    StyleAdapter~\cite{StyleAdapter_wang2024styleadapter} & Style & Feature & CLIP & DCA & SD 1.5 \\
    CustomNet~\cite{Customnet_yuan2023customnet} & Subject & Feature & CLIP & DCA & SD 1.5 \\
    Face-diffuser~\cite{Face-diffuser_wang2024high} & Face & Textual (Word) & CLIP & - & SD 1.5 \\
    $\mathcal{W}$+ Adapter~\cite{wplus_li2024stylegan} & Face & Feature & StyleGAN Encoder & RCA & SD 1.5 \\
    ArtAdapter~\cite{ArtAdapter_chen2024artadapter} & Style & Textual (Word) + Feature & VGG + StyleEncoder & CA & SD 1.5 \\
    CustomizationAssistant~\cite{CustomizationAssistant_zhou2024customization} & Subject & Feature & Llama2 + DINO & CA & SD 2 \\
    FaceStudio~\cite{Facestudio_yan2023facestudio} & Face + Style & Textual (Word) & CLIP + Arcface & - & SD \\
    PhotoMaker~\cite{photomaker_li2024photomaker} & Face & Textual (Word) & CLIP+InsightFace & - & SDXL \\
    PortraitBooth~\cite{Portraitbooth_peng2024portraitbooth} & Face & Textual (Word) & TFace & - & SD 1.5 \\
    RETRIBOORU~\cite{RETRIBOORU_tang2023retrieving} & Character & Feature & CLIP + Retrieval Encoder & CA & SD 1.5 \\
    SSREncoder~\cite{SSREncoder_zhang2024ssr}  & Subject & Feature & SSREncoder & DCA & SD 1.5 \\
    BootPIG~\cite{Bootpig_purushwalkam2024bootpig} & Subject & Feature & SD Unet & RSA & SD 2.1 \\
    Liu~\etal~\cite{identity-expression_liu2024towards} & Face & Feature & DLN+Face Model & CA & LDM \\
    InstantID~\cite{Instantid_wang2024instantid} & Face & Feature & SD Unet + CLIP + Face Model & DCA & SDXL \\
    InstructImagen~\cite{InstructImagen_hu2024instruct} & Subject & Feature & - & CA & Imagen \\
    UNIMO-G~\cite{UNIMOG_li2024unimo} & Subject & Feature & MLLM & CA & SD \\
    CapHuman~\cite{Caphuman_liang2024caphuman} & Face & Feature & CLIP + Face Model & DCA & SD 1.5 \\
    $\lambda$-ECLIPSE~\cite{lambdaECLIPSE_patel2024lambda} & Multiple Subjects & Textual (Output) & OpenCLIP & - & Kandinsky \\
    Face2Diffusion~\cite{Face2Diffusion_shiohara2024face2diffusion} & Face & Textual (Word) & ViT + Face Model & - & SD 1.4 \\
    FlashFace~\cite{FlashFace_zhang2024flashface} & Face & Feature & SD Unet & SA & SD 1.5 \\
    IDAdapter~\cite{IDAdapter_cui2024idadapter} & Face & Textual (Word) + Feature & CLIP + Arcface & SA & SD 2.1 \\
    Infinite-ID~\cite{Infinite-ID_wu2025infinite} & Face & Feature & CLIP + Arcface & DCA + MixA & SDXL \\
    LARGEN~\cite{LARGEN_pan2024locate} & Subject & Feature & CLIP + SD Unet & CA + SA & SD 1.5 \\
    RealCustom~\cite{RealCustom_huang2024realcustom} & Subject & Feature & CLIP & DCA & SD \\
    Song~\etal~\cite{song2024harmonizing} & Subject & - & - & - & ELITE / BLIP-Diffusion \\
    ConsistentID~\cite{Consistentid_huang2024consistentid} & Face & Feature & CLIP + InsightFace & DCA & SD 1.5 \\
    ID-Aligner~\cite{ID-Aligner_chen2024id} & Face & Feature & CLIP & DCA & SD 1.5 / SDXL \\
    InstantStyle~\cite{Instantstyle_wang2024instantstyle} & Style & Feature & CLIP & DCA & SDXL \\
    LCM-Lookahead~\cite{Lcm-lookahead_gal2024lcm} & Face & Feature & SD Unet+CLIP & SA + DCA & SDXL \\
    MoA~\cite{Moa_wang2024moa} & Face & Textual (Word) + Feature & CLIP & SA + DCA & SD 1.5 \\
    MoMA~\cite{MOMA_song2025moma} & Subject & Feature + Textual (Word) & LLaVA + SD Unet & CA & SD 1.5 \\
    Pulid~\cite{Pulid_guo2024pulid} & Face & Feature & CLIP + Antelopev2 & DCA & SDXL \\
    MasterWeaver~\cite{MasterWeaver_wei2025masterweaver} & Face & Feature & CLIP & DCA & SD 1.5 \\
    SAG~\cite{SAG_chan2024improving} & Subject & Textual (Word) + Feature & CLIP & DCA & SD 1.4 \\
    SEGuidance~\cite{SEGuidance_liu2024training} & Multiple Subjects & CLIP & DCA & SD 1.5 \\
    Character-Adapter~\cite{Character-Adapter_ma2024character} & Character & Feature & CLIP & CA & SD 1.5\\
    JeDi~\cite{JeDi_zeng2024jedi} & Subject & Image & - & - & SD 1.4 \\
    LPGen~\cite{Artistic-Intelligence_yang2024artistic} & Style & Feature & CLIP & DCA & SD \\
    PreciseControl~\cite{Precisecontrol_parihar2025precisecontrol} & Face & Textual (Word) & StyleGAN Encoder & - & SD 2.1 \\
    DiffLoRA~\cite{Difflora_wu2024difflora} & Face & Parameter & CLIP + InsightFace & - & SDXL \\
    IPAdapter-Instruct~\cite{IPAdapterInstruct_rowles2024ipadapter} & Style + Subject + Face & Feature & CLIP & DCA & SD 1.5 \\
    MagicID~\cite{MagicID_deng2024magicid} & Face & Feature & CLIP + ArcFace & - & SD 1.5 \\
    RealCustom++~\cite{RealCustom++_mao2024realcustom++} & Subject & Feature & CLIP & DCA & SD 1.5 / SDXL \\
    UniPortrait~\cite{UniPortrait_he2024uniportrait} & Face & Feature & CLIP + Face Model & DCA & SD 1.5 \\
    CSGO~\cite{CSGO_imagecsgo} & Style + Subject & Feature & ViT-H & DCA & SDXL \\
    CustomContrast~\cite{CustomContrast_chen2024customcontrast} & Subject & Feature & MFI Encoder & CA & SD 1.5 / SDXL \\
    EZIGen~\cite{EZIGen_duan2024ezigen} & Subject & Feature & SD Unet & Adapter SA & SD 2.1 \\
    FineStyle~\cite{FineStyle_zhangfinestyle} & Style & Feature & SigLip & - & SD 1.5 \\
    Mohamed~\etal~\cite{Fusion_mohamed2024fusion} & Face & - & - & - & SD 1.5 \\
    GroundingBooth~\cite{GroundingBooth_xiong2024groundingbooth} & Multiple Subjects & Feature & Dino & CA & SD 1.4 \\
    Imagine-Yourself~\cite{Imagine-yourself_he2024imagine} & Face & Feature & CLIP & CA & LDM \\
    MS-Diffusion~\cite{MS-Diffusion_personalizationms} & Multiple Subjects & Feature & CLIP & DCA & SDXL \\
    Omnigen~\cite{Omnigen_xiao2024omnigen} & Subject & Textual (Word) & VAE & - & Phi-3 \\
    StoryMaker~\cite{StoryMaker_zhou2024storymaker} & Character & Feature & CLIP + ArcFace & DCA & SDXL \\
    DisEnvisioner~\cite{DisEnvisioner_he2024disenvisioner} &  Subject & Feature & CLIP + Image Tokenizer & DCA & SD 1.5 \\
    FACT~\cite{FACT_yu2024facechain} & Face & Feature & TransFace & SA & SD 1.5 \\
    HybridBooth~\cite{HybridBooth_guan2025hybridbooth} & Subject + Face & Textual (Word) & DINO & - & SD 1.5 \\
    RelationBooth~\cite{RelationBooth_shi2024relationbooth} & Multiple Subjects & Feature & CLIP & DCA & SDXL \\
    Cai~\etal~\cite{cai2024diffusion} & Subject & Image Token & FLUX Encoder & - & FLUX \\
    DreamCache~\cite{DreamCache_aiello2024dreamcache} & Subject & Feature & SD Unet & CA & SD 1.5 / SD 2.1 \\
    ID-Patch~\cite{ID-Patch_zhang2024id} & Face & Textual (Output) + Feature & ArcFace & CA & SDXL \\
    OneDiffusion~\cite{le2024diffusiongenerate} & Subject & - & VAE & - & Next-DiT \\
    OminiControl~\cite{OminiControl_tan2024ominicontrol} & Subject & - & VAE & - & FLUX \\
    PersonaCraft~\cite{PersonaCraft_kim2024personacraft} & Character & Feature & MultiHMR + Insightface & - & SDXL \\
    AnyDressing~\cite{AnyDressing_li2024anydressing} & Subject & Feature & GFE & Dressing Attention & SD 1.5 \\
    Diff-PC~\cite{Diff-PC_xu2024diff} & Face & Feature & CLIP+Arcface+SMIRK & - & SDXL \\
    EmojiDiff~\cite{EmojiDiff_jiang2024emojidiff} & Face & Feature & CLIP & CA & SD 1.5 / SDXL \\
    LoRA.rar~\cite{LoRA.rar_shenaj2024lora} & Style + Subject & Parameter & - & - & SDXL \\
    MagicNaming~\cite{MagicNaming_zhao2024magicnaming} & Face & Textual (Word) & CLIP & - &  SDXL \\
    P3S-Diffusion~\cite{P3S-Diffusion_hu2024p3s} & Subject & Feature & VAE + CLIP + SD Unet & SA & SD 1.5 \\
    PersonaMagic~\cite{PersonaMagic_li2024personamagic} & Face & Textual (Word) & CLIP & - & SD 1.4 \\
    RealisID~\cite{RealisID_sun2024realisid} & Face & Feature & CLIP + SD Unet & - & SDXL \\
    SerialGen~\cite{SerialGen_xie2024serialgen} & Character & Feature & CLIP & - & SDXL \\
    StoryWeaver~\cite{StoryWeaver_zhang2024storyweaver} & Character & - & - & - & SD 1.5 \\
    Ma~\etal~\cite{zhuoqi2024content} & Style & Parameter & CLIP & CA & SD 1.5 \\
    AnyStory~\cite{AnyStory_he2025anystory} & Character & Feature & SD Unet + CLIP & DCA & SDXL \\
    NestedAttention~\cite{NestedAttention_patashnik2025nested} & Subject & Feature & CLIP & CA & SDXL \\
    Parmar~\etal~\cite{parmar2025object} & Multiple Subjects & Feature & CLIP & DCA & SD 1.5 / SDXL \\
    \hline
    \end{tabular}
    }
\end{table*}

\begin{table*}[t]
    \centering
    \caption{Summary of training datasets among learning-based methods. Paired Images denotes the concept image and target image are collected from same image. Unpaired Images denotes concept image and target image are collected from different images. BBOX denotes the bounding box.}
    \label{tab:data_source}
    \resizebox{1\linewidth}{!}{
    \begin{tabular}{l|cccc}
    \hline
    Method & Category & Data Source & Data Format & Data Scale \\ \hline
    ELITE~\cite{ELITE_wei2023elite} & Subject & OpenImages & (Paired Images, Text) & 13K \\
    UMM~\cite{UMM_ma2023unified} & Subject & LAION-400M & (Paired Images, Text, Mask) & $\sim$ 1.8M \\
    InstantBooth~\cite{Instantbooth_shi2024instantbooth} & Face & Self Collected & (Paired Images, Text, Mask) & 1.43M \\
    InstantBooth~\cite{Instantbooth_shi2024instantbooth} & Cat & Self Collected & (Paired Images, Text, Mask) & 0.37M \\
    SUTI~\cite{SUTI_chen2024subject} & Subject & Generated & (Unpaired Images, Text) & 2M \\
    Taming~\cite{Taming_jia2023taming} & Subject & Internal Data + CelebA + LSUN Dog & (Paired Images, Text, Mask) & - \\
    BLIP-Diffusion~\cite{BLIPDiffusion_li2024blip} & Subject & OpenImages & (Paired Images, Text) & 292K \\
    FastComposer~\cite{Fastcomposer_xiao2024fastcomposer} & Face & FFHQ & (Paired Images, Text, Mask) & 70K \\
    Face0~\cite{Face0_valevski2023face0} & Face & Laion & (Paired Images, Text, Face Embedding) & 10M \\
    AnyDoor~\cite{Anydoor_chen2024anydoor} & Subject & Video Datasets + Multi-View Image Datasets + Single Image Datasets & (Unpaired Images, Paired Images, BBOX) & 400k \\
    Domain Agnostic~\cite{DAT_arar2023domain} & Subject & ImageNet-1K + Open-Images & (Paired Images, Text) & 3M \\
    DreamIdentity~\cite{Dreamidentity_chen2023dreamidentity} & Face & FFHQ & (Unpaired Images, Paired Images, Text) & 70K \\
    Hyperdreambooth~\cite{Hyperdreambooth_ruiz2024hyperdreambooth} & Face & CelebA-HQ & (Paired Images, Text, Model Weights) & 15K \\
    SubjectDiffusion~\cite{SubjectDiffusion_ma2024subject} & Multiple Subjects & Laion-5B & (Paired Images, Text, Bbox, Mask) & 76M  \\
    IP-Adapter~\cite{IPAdapter_ye2023ip} & Subject & Laion-2B + COYO-700M  & (Paired Images, Text) & 10M   \\
    PhotoVerse~\cite{Photoverse_chen2023photoverse} & Face & Fairface + CelebA-HQ + FFHQ & (Paired Images, Text) & 108k + 30k + 70k \\
    StyleAdapter~\cite{StyleAdapter_wang2024styleadapter} & Style & LAION-AESTHETICS & (Unpaired Images, Text) & 600k \\
    CustomNet~\cite{Customnet_yuan2023customnet} & Subject & Objaverse + OpenImages & (Unpaired Images, Camera, Text, Mask) & 250K+500K \\
    Face-diffuser~\cite{Face-diffuser_wang2024high} & Face &  FFHQ & (Paired Images, Text) & 70k \\
    $\mathcal{W}$+ Adapter~\cite{wplus_li2024stylegan} & Face & FFHQ + SHHQ  & (Paired Images, Text, Mask) & 70K + 40K \\
    ArtAdapter~\cite{ArtAdapter_chen2024artadapter} & Style & LAION AESTHETICS + WikiArt & (Paired Images, Text) & - \\
    CustomizationAssistant~\cite{CustomizationAssistant_zhou2024customization} & Subject & Generated & (Unpaired Images, Text) & 1M \\
    FaceStudio~\cite{Facestudio_yan2023facestudio} & Face + Style & FFHQ + LAION & (Paired Images, Text) & - \\
    PhotoMaker~\cite{photomaker_li2024photomaker} & Face & Self Collected & (Unpaired Images, Text) & 112k \\
    PortraitBooth~\cite{Portraitbooth_peng2024portraitbooth} & Face & CelebV-T & (Unpaired Images, Text) & 70k \\
    RETRIBOORU~\cite{RETRIBOORU_tang2023retrieving} & Character & Danbooru 2019 Figures & (Unpaired Images, Text) & 116K \\
    SSREncoder~\cite{SSREncoder_zhang2024ssr} & Subject & Laion5B & (Paired Images, Mask, Text) & 10M \\
    BootPIG~\cite{Bootpig_purushwalkam2024bootpig} & Subject & Generated & (Unpaired Images, Text) & 200K \\
    Identity-Expression Control~\cite{identity-expression_liu2024towards} & Face & CelebA-HQ + FFHQ & (Unpaired Images, Scene Text, Expression Text) & 30k + 70k \\
    InstantID~\cite{Instantid_wang2024instantid} & Face & LAION-Face + Self Collected & (Paired Images, Text) & 50M + 10M \\
    InstructImagen~\cite{InstructImagen_hu2024instruct} & Subject & SUTI Dataset & (Unpaired Images, Text) & - \\
    UNIMO-G~\cite{UNIMOG_li2024unimo} & Subject & LAION-2B + COYO-700M + Self Collected & (Paired Images, Text) & 1M \\
    CapHuman~\cite{Caphuman_liang2024caphuman} & Face & CelebA & (Paired Images, Text) & 200K \\
    $\lambda$-ECLIPSE~\cite{lambdaECLIPSE_patel2024lambda} & Multiple Subjects & LAION-5B & (Paired Images, Text) & 2M \\
    Face2Diffusion~\cite{Face2Diffusion_shiohara2024face2diffusion} & Face & FFHQ & (Paired Images, Text) & 70k \\
    FlashFace~\cite{FlashFace_zhang2024flashface} & Face & Self Collected & (Unpaired Images, Text) & 1.8M \\
    IDAdapter~\cite{IDAdapter_cui2024idadapter} & Face & CelebA-HQ & (Unpaired Images, Text) & 30K \\
    Infinite-ID~\cite{Infinite-ID_wu2025infinite} & Face &  LAION-2B + LAION-Face + Self Collected & (Unpaired Images) & 10M + 50M + - \\
    LARGEN~\cite{LARGEN_pan2024locate} & Subject & Generated & (Paired Images, Text, Mask) & - \\
    RealCustom~\cite{RealCustom_huang2024realcustom} & Subject & Laion 5B & (Paired Image, Text) & - \\
    ConsistentID~\cite{Consistentid_huang2024consistentid} & Face & FFHQ + CelebA + SFHQ & (Paired Images, Text) & 500K \\
    ID-Aligner~\cite{ID-Aligner_chen2024id} & Face & LAION & (Paired Images, Text) & 200K \\
    InstantStyle~\cite{Instantstyle_wang2024instantstyle} & Style & - & (Paired Images, Text) & 4M \\
    LCM-Lookahead~\cite{Lcm-lookahead_gal2024lcm} & Face & Generated & (Unpaired Images, Text) & 500K \\
    MoA~\cite{Moa_wang2024moa} & Face & FFHQ & (Paired Images, Text) & 60K \\
    MoMA~\cite{MOMA_song2025moma} & Subject & OpenImage V7 & (Paired Images, Text, Mask) & 282k \\
    PuLID~\cite{Pulid_guo2024pulid} & Face & Self Collected & (Paired Images, Text) & 1.5M \\
    MasterWeaver~\cite{MasterWeaver_wei2025masterweaver} & Face & Generated & (Unpaired Images, Text, Mask) & 160K \\
    SAG~\cite{SAG_chan2024improving} & Subject & Generated & (Paired Images, Text) & - \\
    SEGuidance~\cite{SEGuidance_liu2024training} & Multiple Subjects & LAION-2B + COYO-700M & (Paired Images, Text) & 10M \\
    Character-Adapter~\cite{Character-Adapter_ma2024character} & Character & - & (Paired Images, Prompt Text, Image Description Text) & - \\
    JeDi~\cite{JeDi_zeng2024jedi} & Subject & Generated + WebVid10M + LAION Aesthetic & (Unpaired + Paired Images, Text) & 1.6M + X \\
    LPGen~\cite{Artistic-Intelligence_yang2024artistic} & Style & Self Collected & (Paired Images, Text) & 2K \\
    PreciseControl~\cite{Precisecontrol_parihar2025precisecontrol} & Face & Generated + FFHQ & (Paired Images, Text) & X + 70K \\
    DiffLoRA~\cite{Difflora_wu2024difflora} & Face & Generated & (Unpaired Images, Text, LoRA) & - \\
    IPAdapter-Instruct~\cite{IPAdapterInstruct_rowles2024ipadapter} & Style + Subject + Face & JourneyDB + Generated + COCO+CelebA & (Paired Images, Unpaired Images, Instruction) & 42K + 35K + 40k \\
    MagicID~\cite{MagicID_deng2024magicid} & Face & CeleB-A + FFHQ + LAION-Face + Self Collected & (Paired Images, Text, Mask) & 1M \\
    RealCustom++~\cite{RealCustom++_mao2024realcustom++} & Multiple Subjects & Laion 5B + MVImageNet & (Paired + Unpaired Image, Text) & - \\
    UniPortrait~\cite{UniPortrait_he2024uniportrait} & Face & LAION + CelebA + Self Collected & (Paired Images, Text) & 240K+100K+160K+120K \\
    CSGO~\cite{CSGO_imagecsgo} & Style + Subject & Generated & (Unpaired Images, Text) & 210K \\
    CustomContrast~\cite{CustomContrast_chen2024customcontrast} & MVImageNet + OpenImages & (Paired Images, Text) & - \\
    EZIGen~\cite{EZIGen_duan2024ezigen} & Subject & COCO2014 + Youtube VIS & (Paired + Unpaired Images, Text) & 200K \\
    FineStyle~\cite{FineStyle_zhangfinestyle} & Style & Generated & (Unpaired Images, Text) & 35k \\
    Fusion is all you need~\cite{Fusion_mohamed2024fusion} & Face & LAION-Face & (Paired Images, Text) & 80K \\
    GroundingBooth~\cite{GroundingBooth_xiong2024groundingbooth} & Multiple Subjects & MVImgNet + LVIS & (Unpaired + Paired Images, BBOX, Text) & 6.6 M \\
    Imagine yourself~\cite{Imagine-yourself_he2024imagine} & Face & Generated & (Unpaired Images, Text) & 9M \\
    MS-Diffusion~\cite{MS-Diffusion_personalizationms} & Multiple Subjects & Video dataset & (Paired + Unpaired Images, BbOX, Text) & 3.6 M \\
    Omnigen~\cite{Omnigen_xiao2024omnigen} & Subject & GRIT + Self Collected & (Paired + Unpaired Images, Text) & 6M + 533K \\
    StoryMaker~\cite{StoryMaker_zhou2024storymaker} & Character & Self Collected & (Paired Images, Text) & 500K \\
    DisEnvisioner~\cite{DisEnvisioner_he2024disenvisioner} & Subject & OpenImages V6 & (Paired Images, Text)  & 6.82 M \\
    FACT~\cite{FACT_yu2024facechain} & Face & - & (Paired Images, Text) & - \\
    HybridBooth~\cite{HybridBooth_guan2025hybridbooth}  & Subject + Face & FFHQ & (Paired Images, Text) & 70K \\
    RelationBooth~\cite{RelationBooth_shi2024relationbooth} & Multiple Subjects & Generated & (Paired + Unpaired Images, BBOX, Text) & - \\
    Diffusion Self-Distillation~\cite{cai2024diffusion} & Subject & Generated & (Unpaired Images, Text) & 400k \\
    DreamCache~\cite{DreamCache_aiello2024dreamcache} & Subject & Generated & (Unpaired Images, Text) & - \\
    ID-Patch~\cite{ID-Patch_zhang2024id} & Face & - & (Paired Images, Landmarks, Text) & 17M + 1.95 M \\
    OneDiffusion~\cite{le2024diffusiongenerate} & Subject & Self Collected & (Unpaired Images, Text) & 130K \\
    OminiControl~\cite{OminiControl_tan2024ominicontrol} & Subject & Generated & (Unpaired Images, Text) & 200K \\
    PersonaCraft~\cite{PersonaCraft_kim2024personacraft} & Character & MPII & (Paired Images, Text, SMPLx) & 6K \\
    AnyDressing~\cite{AnyDressing_li2024anydressing} & Subject & Generated & (Paired Images, Text) & 26K + 37K \\
    Diff-PC~\cite{Diff-PC_xu2024diff} & Face & Self Collected & (Unpaired Images, Text) & 650K \\
    EmojiDiff~\cite{EmojiDiff_jiang2024emojidiff} & Face & Self Collected + Generated & (Paired + UnpairedImages, Text) & 10K + 100K \\
    LoRA.rar~\cite{LoRA.rar_shenaj2024lora} & Style + Subject & Generated & (LoRA Pairs, Text) & 360 \\
    MagicNaming~\cite{MagicNaming_zhao2024magicnaming} & Face & Laion 5B & (Unpaired Images, Text, Name Embeddings) & 810K \\
    RealisID~\cite{RealisID_sun2024realisid} & Face & CosmicMan & (Paired Images, Text) & 2M \\
    SerialGen~\cite{SerialGen_xie2024serialgen} & Character & Generated + Self Collected & (Unpaired Images, Text) & 1M + 300K \\
    StoryWeaver~\cite{StoryWeaver_zhang2024storyweaver} & Character & Self Collected & (Unpaired Images, Text) & - \\
    Ma~\etal~\cite{zhuoqi2024content} & Style & Self Collected & (Paired Images, Text) & 146K \\
    AnyStory~\cite{AnyStory_he2025anystory} & Character & Self Collected + LAION & (Paired + Unpaired Images, Text) & 6.5M + 300K \\
    NestedAttention~\cite{NestedAttention_patashnik2025nested} & Subject & FFHQ + Generated + AFHQ & (Paired Images, Text) & 70K + 50K + 15K + 1K \\
    Parmar~\etal~\cite{parmar2025object} & Multiple Subjects & Coyo-700m + Generated & (Unpaired Images, Text) & - \\
    \hline
    \end{tabular}
    }
\end{table*}

\section{Personalized Image Generation in ARs}
\label{sec:ars}

With the advancement of large language models (LLMs), autoregressive (AR) models have gained significant attention for their capabilities in text-to-image and multi-modal image generation.
Although existing AR-based methods designed for personalized image generation are rare, this task can be considered a subset of multi-modal image generation.
Consequently, many critical techniques overlap between personalization and multi-modal image generation. 
In this section, we explore various AR-based multi-modal generation methods that hold potential for personalized image generation.

As shown in Fig.~\ref{fig:inversion_space}, unlike GANs and diffusion models, multi-modal AR models~\cite{MetaMorph_tong2024metamorph,Liquid_wu2024liquid,Emu2_sun2024generative,Chameleon_team2024chameleon,SeedX_ge2024seed,Puma_fang2024puma,ILLUME_wang2024illume} encode both images and text into a shared token space, enabling the generation of target images by integrating these modalities. 
For example, Emu2~\cite{Emu2_sun2024generative} tokenizes images into embeddings using a CLIP-based visual encoder. 
During generation, image embeddings are combined with text tokens to predict the generated image embeddings autoregressively. 
These visual embeddings are then decoded into images via a diffusion-based visual decoder.
Chameleon~\cite{Chameleon_team2024chameleon} leverages VQGAN as both the visual encoder and decoder, bridging image understanding and generation through discrete image tokens.
This mixed-modal approach allows the model to understand and generate images and text in arbitrary sequences.
Seed-X~\cite{SeedX_ge2024seed} introduces dynamic resolution image encoding that enables the processing of images with varying sizes and aspect ratios by dividing them into grids of sub-images. 
To retain fine-grained details for image manipulation, it adopts a diffusion-based visual de-tokenizer and finetunes it to accept an additional condition image as input.
PUMA~\cite{Puma_fang2024puma} utilizes a CLIP-based semantic image encoder to extract multi-scale features, which serve as the foundation for diverse visual tasks. 
Then, the autoregressive multimodal large language model (MLLM) processes these multi-scale image features alongside text features to predicting each token sequentially, from the coarsest to the finest granularity level. 
A set of diffusion-based decoders is then employed to generate or reconstruct images conditioned on different features.
ILLUME~\cite{ILLUME_wang2024illume} uses a pretrained vision encoder to extract semantic features and supervises the quantization process through feature reconstruction loss. 
It employs the Stable Diffusion model as a decoder to reconstruct these semantic features back into high-resolution images.

During training, these methods are typically trained using a next-token prediction framework and employ a two-stage training pipeline, \ie, multimodal pretraining followed by task-specific instruction tuning. 
During the pretraining phase, models are trained on large text-image pairs to enable comprehensive ability of image understanding and generation.
Depending on the type of image embeddings, classification loss or regression loss is adopted to regularize the predicted image tokens. 
Subsequently, models are aligned to follow specific task instructions through instruction tuning~\cite{Emu2_sun2024generative,Emu3_wang2024emu3,SeedX_ge2024seed,ILLUME_wang2024illume}.
For example, Emu2~\cite{Emu2_sun2024generative} leverages a mix of high-quality datasets to enhance controllable generation within context. 
More specific, grounded image-text pair datasets are utilized to improve the model's ability for personalized image generation.

Overall, the multi-modal capabilities of AR models demonstrates great potential for personalized image generation by leveraging the strengths of text and image understanding.
However, several challenges remain. 
For example, the identity consistency between the generated images and reference images remains limited, as illustrated in Fig.~\ref{fig:ar_generation}.

\section{Evaluation}
\label{sec:evaluation}

\subsection{Evaluation Dataset}

Several datasets have been developed to facilitate the evaluation of personalized image generation models:

\noindent \textbf{DreamBench}~\cite{Dreambooth_ruiz2023dreambooth} is a primary dataset consisting of 30 subjects, including categories such as backpacks, animals, cars, and toys. For each subject, 25 prompts are incorporated to ensure diverse evaluation, covering various contexts, accessories, and attributes.

\noindent \textbf{DreamBench-v2}~\cite{SUTI_chen2024subject} is expanded upon DreamBench by adding 220 additional test prompts for each subject, enhancing the dataset's variability.

\noindent \textbf{Dreambench++}~\cite{peng2024dreambench++} scales the DreamBench data to include 150 images and 1,350 prompts, providing a more extensive framework for evaluating model performance.

\noindent \textbf{Custom101}~\cite{CD_kumari2023multi} is a dataset of 101 concepts with 3-15 images and 20 prompts for each concept, offering a broader scope for evaluation compared to previous datasets.

For specialized evaluation tasks, additional datasets and benchmarks are employed to evaluate models' performance in specific scenarios.
For example, face-driven methods typically utilize images from the CelebA dataset, which provides a rich collection of facial images.
Unlike face personalization, style personalization lacks a standardized benchmark dataset. 
Consequently, users often rely on self-collected datasets tailored to their specific evaluation requirements.

\subsection{Evaluation Metrics}

Personalized image generation should ensure both the fidelity of the generated concept and the alignment with input texts. 
Therefore, the primary evaluation metrics are categorized into several categories, \ie, concept fidelity and text editability.

\subsubsection{Concept Fidelity}
\label{sec:fidelity}

Concept fidelity measures how accurately the generated images reflect the target concepts. 
Several metrics are used:

\noindent \textbf{Fr\bm{$\acute{e}$}chet inception distance}~\cite{FID_heusel2017gans} (FID) calculates the Fr$\acute{e}$chet distance between feature vectors of real and generated images extracted from the Inception-v3~\cite{Inception_szegedy2016rethinking} model's pool3 layer. 
A lower FID score indicates better perceptual quality and closer resemblance to real images.

\noindent \textbf{Identity score} (ID) is relevant for face generation, which measures the similarity between the generated image and target image using a pretrained face recognition model~\cite{Arcface_deng2019arcface}. 
Higher ID scores denote greater similarity in identity. 

\noindent \textbf{CLIP-I} calculates the CLIP visual similarity between the generated images and the concept images.
Higher CLIP-I values indicate greater alignment between the images.

\noindent \textbf{DINO-I} calculates cosine similarity between the ViTS/16 DINO~\cite{Dinov2_oquab2023dinov2} embeddings of the generated images and the concept images.
Similar to CLIP-I, higher DINO-I values reflect better image similarity.

To ensure that background elements do not skew the evaluation of the main subject, some studies~\cite{DETEX_cai2024decoupled} calculate CLIP-I and DINO-I exclusively on foreground objects.

\subsubsection{Text Editability}
\label{sec:editability}

Text editability assesses how well the generated images adhere to the semantic instructions provided in text prompts. Key metrics include:

\noindent \textbf{CLIP-T} evaluates the alignment between text prompts and generated images by measuring the similarity of their respective CLIP embeddings. 
Higher scores indicate better adherence to the textual instructions.

\noindent \textbf{ImageReward}~\cite{Imagereward_xu2024imagereward} is a metric that measures the alignment of generated images with text prompts, providing an additional layer of assessment for image-text compatibility.

\subsubsection{Subjective Metrics}
\label{sec:subjective}

In addition to quantitative metrics, subjective evaluations involving human raters are employed to gauge the quality and relevance of generated images. 
A common approach is user study~\cite{CD_kumari2023multi,ELITE_wei2023elite} .
Participants are presented with sets of images, typically including the source image, results from a baseline model, and results from the proposed method. 
They are asked questions such as ``Which image is more consistent with text?" or ``Which image better represents the objects in target image?".
The percentage of preferences for proposed method over baselines provides insight into its effectiveness.

While subjective metrics offer valuable insights into human perception and satisfaction, they come with drawbacks such as potential biases, variance in human judgment, and high costs associated with conducting comprehensive studies.

\section{Challenge and Future Directions}
\label{sec:challenge}

While significant advancements have been made in personalized image generation, several critical challenges remain.
In this section, we will discuss the key challenges of personalization as well as the future research directions.

\subsection{Trade off on Subject Fidelity and Text Controllability} 

Personalized image generation aims to produce images that faithfully represent the target concept while adhering to textual prompts.
Existing methods~\cite{IPAdapter_ye2023ip,SubjectDiffusion_ma2024subject,MOMA_song2025moma} have made great efforts to enhance the identity fidelity, but usually suffer from overfitting issues. 
These approaches tend to replicate the given concept in the generated image while neglecting the text instructions, especially when generating the concept in diverse poses or with varying attributes. 
To enhance text controllability, several techniques have been proposed, including disentanglement learning~\cite{Disenbooth_chen2023disenbooth,DETEX_cai2024decoupled}, advanced loss functions~\cite{Pulid_guo2024pulid,MasterWeaver_wei2025masterweaver,Lcm-lookahead_gal2024lcm}, data augmentation~\cite{photomaker_li2024photomaker,JeDi_zeng2024jedi}, and sampling strategies~\cite{ProFusion_zhou2023enhancing}.
Despite these advancements, the results remain suboptimal.
Intuitively, achieving high subject fidelity involves capturing and reproducing intricate details of user-specific concept, which may conflict with the changes (\eg, poses or attributes) suggested by textual prompts.
Therefore, more elaborate design is worth investigating to better balance identity fidelity and text controllability.

\subsection{Universal Category Personalization}

Current personalization methods typically focus on specific domains, such as subjects~\cite{ELITE_wei2023elite,MOMA_song2025moma,TI_gal2022image}, faces~\cite{CelebBasis_yuan2023inserting,photomaker_li2024photomaker,Instantid_wang2024instantid,Pulid_guo2024pulid}, clothing~\cite{AnyDressing_li2024anydressing}, or styles~\cite{InST_zhang2023inversion,Instantstyle_wang2024instantstyle}. 
However, in practice, users further expect the ability to incorporate multiple customized concepts within a single image. 
Although several methods~\cite{Ziplora_shah2025ziplora,MS-Diffusion_personalizationms,Omg_kong2025omg,PairCustomization_jones2024customizing} have explored generating multiple subjects or combining content and style, their capacities remain limited.
Therefore, there is a need to develop methods that enable universal category personalization, which provide more flexible and versatile generation capabilities.
Beyond the existing personalization categories, more fine-grained personalization, such as attribute or part, is also needed to enable users to precisely control the generated images.

\subsection{Multi-Condition Controllable Image Generation}
Multi-condition image generation is an emerging and promising field focused on developing unified models capable of handling multiple conditional inputs and supporting a variety of tasks.
For example, several recent methods~\cite{Omnigen_xiao2024omnigen,OminiControl_tan2024ominicontrol,Emu2_sun2024generative} have explored using a single model to handle diverse conditions and tasks simultaneously, including text-to-image synthesis, personalized image generation, conditional image generation, and image editing. 
Implementing such a unified framework significantly enhances the versatility and applicability of generative models across a wide range of contexts.
Despite these advancements, there remains a need to develop robust multi-condition image generation methods that ensure both scalability and high generation quality.

\subsection{Personalization with Advanced Generative Models} 

Generative models have undergone significant evolution, including recent advancements such as DiT-based text-to-image diffusion models~\cite{SD3_esser2024scaling,flux2023} and multi-model autoregressive models~\cite{Emu_sun2023emu,Emu2_sun2024generative,SeedX_ge2024seed}.
Effectively incorporating personalized concepts into these advanced models presents new challenges. 
As shown in Fig.~\ref{fig:ar_generation}, the identity fidelity of autoregressive-based personalization methods remains unsatisfactory.
Developing effective personalization techniques that are compatible with these models is an ongoing area of research. 

\subsection{Personalized Video and 3D Generation} 

In recent years, video and 3D content generation have advanced rapidly, with several methods~\cite{CustomVideoX_she2025customvideox,Pia_zhang2024pia,MovieWeaver_liang2025movie,Motrans_li2024motrans,PersonalVideo_li2024personalvideo,Customvideo_wang2024customvideo,Videobooth_jiang2024videobooth,Dreambooth3d_raj2023dreambooth3d,ouyang2023chasing} also explore the application of personalization techniques in these areas. 
These methods introduce new opportunities for content creation, but also present new challenges. 
For example, personalized video generation requires maintaining appearance and structure consistency across frames, which is more complex than single-image generation.
Similarly, 3D generation involves ensuring multi-view consistency, necessitating models that can effectively understand and depict three-dimensional structures. 
Future research can be done on these fields to extend personalized generation techniques beyond static images to dynamic and spatially complex content.

\section{Conclusion}
\label{sec:conclusion}

In this paper, we present a comprehensive survey of personalized image generation with various generative models, including Generative Adversarial Networks (GANs), text-to-image diffusion models, and multi-modal autoregressive models.
We begin by defining the scope of personalized image generation from a holistic perspective, unifying different approaches under a common framework. 
Specifically, we categorize personalized image generation into three key components: inversion spaces, inversion methods, and personalization schemes.
Building on this unified framework, we provide an in-depth analysis of techniques within each category of generative models, highlighting both the commonalities and differences across existing methods. 
Finally, we discuss the open challenges in the field and propose potential directions for future research. 
Our survey offers a current and comprehensive overview of personalized image generation, systematically tracking related studies in this rapidly evolving domain.

\section*{Acknowledgements}
The work was supported by National Key R\&D Program of China under Grant No. 2022YFA1004100.

\bibliographystyle{CVMbib}
\bibliography{arxiv}

\begin{thebibliography}{100}
\expandafter\ifx\csname urlstyle\endcsname\relax
  \providecommand{\doi}[1]{doi:\discretionary{}{}{}#1}\else
  \providecommand{\doi}{doi:\discretionary{}{}{}\begingroup
  \urlstyle{rm}\Url}\fi

\bibitem{GAN_goodfellow2020generative}
Goodfellow I, Pouget-Abadie J, Mirza M, Xu B, Warde-Farley D, Ozair S,
  Courville A, Bengio Y. Generative adversarial networks. \emph{Communications
  of the ACM}, 2020, 63(11): 139--144.

\bibitem{DDPM_ho2020denoising}
Ho J, Jain A, Abbeel P. Denoising diffusion probabilistic models.
  \emph{Advances in neural information processing systems}, 2020, 33:
  6840--6851.

\bibitem{Emu_sun2023emu}
Sun Q, Yu Q, Cui Y, Zhang F, Zhang X, Wang Y, Gao H, Liu J, Huang T, Wang X.
  Emu: Generative pretraining in multimodality. In \emph{The Twelfth
  International Conference on Learning Representations}, 2023.

\bibitem{LDM_rombach2022high}
Rombach R, Blattmann A, Lorenz D, Esser P, Ommer B. High-resolution image
  synthesis with latent diffusion models. In \emph{Proceedings of the IEEE/CVF
  conference on computer vision and pattern recognition}, 2022, 10684--10695.

\bibitem{SDXL_podell2023sdxl}
Podell D, English Z, Lacey K, Blattmann A, Dockhorn T, M{\"u}ller J, Penna J,
  Rombach R. Sdxl: Improving latent diffusion models for high-resolution image
  synthesis. \emph{arXiv preprint arXiv:2307.01952}, 2023.

\bibitem{Imagen_saharia2022photorealistic}
Saharia C, Chan W, Saxena S, Li L, Whang J, Denton EL, Ghasemipour K,
  Gontijo~Lopes R, Karagol~Ayan B, Salimans T, et~al.. Photorealistic
  text-to-image diffusion models with deep language understanding.
  \emph{Advances in neural information processing systems}, 2022, 35:
  36479--36494.

\bibitem{DALLE2_ramesh2022hierarchical}
Ramesh A, Dhariwal P, Nichol A, Chu C, Chen M. Hierarchical text-conditional
  image generation with clip latents. \emph{arXiv preprint arXiv:2204.06125},
  2022, 1(2): 3.

\bibitem{Controlnet_zhang2023adding}
Zhang L, Rao A, Agrawala M. Adding conditional control to text-to-image
  diffusion models. In \emph{Proceedings of the IEEE/CVF International
  Conference on Computer Vision}, 2023, 3836--3847.

\bibitem{prompt2prompt_hertz2022prompt}
Hertz A, Mokady R, Tenenbaum J, Aberman K, Pritch Y, Cohen-Or D.
  Prompt-to-prompt image editing with cross attention control. \emph{arXiv
  preprint arXiv:2208.01626}, 2022.

\bibitem{Ledits++_brack2024ledits++}
Brack M, Friedrich F, Kornmeier K, Tsaban L, Schramowski P, Kersting K, Passos
  A. Ledits++: Limitless image editing using text-to-image models. In
  \emph{Proceedings of the IEEE/CVF Conference on Computer Vision and Pattern
  Recognition}, 2024, 8861--8870.

\bibitem{Masactrl_cao2023masactrl}
Cao M, Wang X, Qi Z, Shan Y, Qie X, Zheng Y. Masactrl: Tuning-free mutual
  self-attention control for consistent image synthesis and editing. In
  \emph{Proceedings of the IEEE/CVF International Conference on Computer
  Vision}, 2023, 22560--22570.

\bibitem{photomaker_li2024photomaker}
Li Z, Cao M, Wang X, Qi Z, Cheng MM, Shan Y. Photomaker: Customizing realistic
  human photos via stacked id embedding. In \emph{Proceedings of the IEEE/CVF
  Conference on Computer Vision and Pattern Recognition}, 2024, 8640--8650.

\bibitem{Instantstyle_wang2024instantstyle}
Wang H, Spinelli M, Wang Q, Bai X, Qin Z, Chen A. Instantstyle: Free lunch
  towards style-preserving in text-to-image generation. \emph{arXiv preprint
  arXiv:2404.02733}, 2024.

\bibitem{TI_gal2022image}
Gal R, Alaluf Y, Atzmon Y, Patashnik O, Bermano AH, Chechik G, Cohen-Or D. An
  image is worth one word: Personalizing text-to-image generation using textual
  inversion. \emph{arXiv preprint arXiv:2208.01618}, 2022.

\bibitem{Dreambooth_ruiz2023dreambooth}
Ruiz N, Li Y, Jampani V, Pritch Y, Rubinstein M, Aberman K. Dreambooth: Fine
  tuning text-to-image diffusion models for subject-driven generation. In
  \emph{Proceedings of the IEEE/CVF conference on computer vision and pattern
  recognition}, 2023, 22500--22510.

\bibitem{ELITE_wei2023elite}
Wei Y, Zhang Y, Ji Z, Bai J, Zhang L, Zuo W. Elite: Encoding visual concepts
  into textual embeddings for customized text-to-image generation. In
  \emph{Proceedings of the IEEE/CVF International Conference on Computer
  Vision}, 2023, 15943--15953.

\bibitem{IPAdapter_ye2023ip}
Ye H, Zhang J, Liu S, Han X, Yang W. Ip-adapter: Text compatible image prompt
  adapter for text-to-image diffusion models. \emph{arXiv preprint
  arXiv:2308.06721}, 2023.

\bibitem{CD_kumari2023multi}
Kumari N, Zhang B, Zhang R, Shechtman E, Zhu JY. Multi-concept customization of
  text-to-image diffusion. In \emph{Proceedings of the IEEE/CVF Conference on
  Computer Vision and Pattern Recognition}, 2023, 1931--1941.

\bibitem{GVM_zhu2016generative}
Zhu JY, Kr{\"a}henb{\"u}hl P, Shechtman E, Efros AA. Generative visual
  manipulation on the natural image manifold. In \emph{Computer Vision--ECCV
  2016: 14th European Conference, Amsterdam, The Netherlands, October 11-14,
  2016, Proceedings, Part V 14}, 2016, 597--613.

\bibitem{psp_richardson2021encoding}
Richardson E, Alaluf Y, Patashnik O, Nitzan Y, Azar Y, Shapiro S, Cohen-Or D.
  Encoding in style: a stylegan encoder for image-to-image translation. In
  \emph{Proceedings of the IEEE/CVF conference on computer vision and pattern
  recognition}, 2021, 2287--2296.

\bibitem{e4e_tov2021designing}
Tov O, Alaluf Y, Nitzan Y, Patashnik O, Cohen-Or D. Designing an encoder for
  stylegan image manipulation. \emph{ACM Transactions on Graphics (TOG)}, 2021,
  40(4): 1--14.

\bibitem{liu2023survey}
Liu M, Wei Y, Wu X, Zuo W, Zhang L. Survey on leveraging pre-trained generative
  adversarial networks for image editing and restoration. \emph{Science China
  Information Sciences}, 2023, 66(5): 151101.

\bibitem{xia2022gan}
Xia W, Zhang Y, Yang Y, Xue JH, Zhou B, Yang MH. Gan inversion: A survey.
  \emph{IEEE transactions on pattern analysis and machine intelligence}, 2022,
  45(3): 3121--3138.

\bibitem{E4T_gal2023encoder}
Gal R, Arar M, Atzmon Y, Bermano AH, Chechik G, Cohen-Or D. Encoder-based
  domain tuning for fast personalization of text-to-image models. \emph{ACM
  Transactions on Graphics (TOG)}, 2023, 42(4): 1--13.

\bibitem{DAT_arar2023domain}
Arar M, Gal R, Atzmon Y, Chechik G, Cohen-Or D, Shamir A, H~Bermano A.
  Domain-agnostic tuning-encoder for fast personalization of text-to-image
  models. In \emph{SIGGRAPH Asia 2023 Conference Papers}, 2023, 1--10.

\bibitem{DALLE_ramesh2021zero}
Ramesh A, Pavlov M, Goh G, Gray S, Voss C, Radford A, Chen M, Sutskever I.
  Zero-shot text-to-image generation. In \emph{International conference on
  machine learning}, 2021, 8821--8831.

\bibitem{Cogview_ding2021cogview}
Ding M, Yang Z, Hong W, Zheng W, Zhou C, Yin D, Lin J, Zou X, Shao Z, Yang H,
  et~al.. Cogview: Mastering text-to-image generation via transformers.
  \emph{Advances in neural information processing systems}, 2021, 34:
  19822--19835.

\bibitem{Emu2_sun2024generative}
Sun Q, Cui Y, Zhang X, Zhang F, Yu Q, Wang Y, Rao Y, Liu J, Huang T, Wang X.
  Generative multimodal models are in-context learners. In \emph{Proceedings of
  the IEEE/CVF Conference on Computer Vision and Pattern Recognition}, 2024,
  14398--14409.

\bibitem{cao2024controllable}
Cao P, Zhou F, Song Q, Yang L. Controllable generation with text-to-image
  diffusion models: A survey. \emph{arXiv preprint arXiv:2403.04279}, 2024.

\bibitem{zhang2024text}
Zhang N, Tang H. Text-to-Image Synthesis: A Decade Survey. \emph{arXiv preprint
  arXiv:2411.16164}, 2024.

\bibitem{zhan2024conditional}
Zhan Z, Chen D, Mei JP, Zhao Z, Chen J, Chen C, Lyu S, Wang C. Conditional
  Image Synthesis with Diffusion Models: A Survey. \emph{arXiv preprint
  arXiv:2409.19365}, 2024.

\bibitem{shuai2024survey}
Shuai X, Ding H, Ma X, Tu R, Jiang YG, Tao D. A Survey of Multimodal-Guided
  Image Editing with Text-to-Image Diffusion Models. \emph{arXiv preprint
  arXiv:2406.14555}, 2024.

\bibitem{zhang2024survey}
Zhang X, Wei XY, Zhang W, Wu J, Zhang Z, Lei Z, Li Q. A Survey on Personalized
  Content Synthesis with Diffusion Models. \emph{arXiv preprint
  arXiv:2405.05538}, 2024.

\bibitem{katsumata2023revisiting}
Katsumata K, Vo DM, Liu B, Nakayama H. Revisiting Latent Space of GAN Inversion
  for Real Image Editing. \emph{arXiv preprint arXiv:2307.08995}, 2023.

\bibitem{Stylespace_wu2021stylespace}
Wu Z, Lischinski D, Shechtman E. Stylespace analysis: Disentangled controls for
  stylegan image generation. In \emph{Proceedings of the IEEE/CVF conference on
  computer vision and pattern recognition}, 2021, 12863--12872.

\bibitem{HFGI_wang2022high}
Wang T, Zhang Y, Fan Y, Wang J, Chen Q. High-fidelity gan inversion for image
  attribute editing. In \emph{Proceedings of the IEEE/CVF conference on
  computer vision and pattern recognition}, 2022, 11379--11388.

\bibitem{SAMInversion_parmar2022spatially}
Parmar G, Li Y, Lu J, Zhang R, Zhu JY, Singh KK. Spatially-adaptive multilayer
  selection for gan inversion and editing. In \emph{Proceedings of the IEEE/CVF
  conference on computer vision and pattern recognition}, 2022, 11399--11409.

\bibitem{StylePrompter_zhuang2023styleprompter}
Zhuang C, Gao P, Smolic A. StylePrompter: All Styles Need Is Attention. In
  \emph{Proceedings of the 31st ACM International Conference on Multimedia},
  2023, 2487--2497.

\bibitem{PTI_roich2022pivotal}
Roich D, Mokady R, Bermano AH, Cohen-Or D. Pivotal tuning for latent-based
  editing of real images. \emph{ACM Transactions on graphics (TOG)}, 2022,
  42(1): 1--13.

\bibitem{Hyperstyle_alaluf2022hyperstyle}
Alaluf Y, Tov O, Mokady R, Gal R, Bermano A. Hyperstyle: Stylegan inversion
  with hypernetworks for real image editing. In \emph{Proceedings of the
  IEEE/CVF conference on computer Vision and pattern recognition}, 2022,
  18511--18521.

\bibitem{InvertingGAN_creswell2018inverting}
Creswell A, Bharath AA. Inverting the generator of a generative adversarial
  network. \emph{IEEE transactions on neural networks and learning systems},
  2018, 30(7): 1967--1974.

\bibitem{StochasticClipping_lipton2017precise}
Lipton ZC, Tripathi S. Precise recovery of latent vectors from generative
  adversarial networks. \emph{arXiv preprint arXiv:1702.04782}, 2017.

\bibitem{PGD_shah2018solving}
Shah V, Hegde C. Solving linear inverse problems using gan priors: An algorithm
  with provable guarantees. In \emph{2018 IEEE international conference on
  acoustics, speech and signal processing (ICASSP)}, 2018, 4609--4613.

\bibitem{ma2018invertibility}
Ma F, Ayaz U, Karaman S. Invertibility of convolutional generative networks
  from partial measurements. \emph{Advances in Neural Information Processing
  Systems}, 2018, 31.

\bibitem{Image2stylegan_abdal2019image2stylegan}
Abdal R, Qin Y, Wonka P. Image2stylegan: How to embed images into the stylegan
  latent space? In \emph{Proceedings of the IEEE/CVF international conference
  on computer vision}, 2019, 4432--4441.

\bibitem{Image2stylegan++_abdal2020image2stylegan++}
Abdal R, Qin Y, Wonka P. Image2stylegan++: How to edit the embedded images? In
  \emph{Proceedings of the IEEE/CVF conference on computer vision and pattern
  recognition}, 2020, 8296--8305.

\bibitem{feng2022near}
Feng Q, Shah V, Gadde R, Perona P, Martinez A. Near perfect gan inversion.
  \emph{arXiv preprint arXiv:2202.11833}, 2022.

\bibitem{Hyperinverter_dinh2022hyperinverter}
Dinh TM, Tran AT, Nguyen R, Hua BS. Hyperinverter: Improving stylegan inversion
  via hypernetwork. In \emph{Proceedings of the IEEE/CVF conference on computer
  vision and pattern recognition}, 2022, 11389--11398.

\bibitem{Restyle_alaluf2021restyle}
Alaluf Y, Patashnik O, Cohen-Or D. Restyle: A residual-based stylegan encoder
  via iterative refinement. In \emph{Proceedings of the IEEE/CVF international
  conference on computer vision}, 2021, 6711--6720.

\bibitem{E2Style_wei2022e2style}
Wei T, Chen D, Zhou W, Liao J, Zhang W, Yuan L, Hua G, Yu N. E2Style: Improve
  the efficiency and effectiveness of StyleGAN inversion. \emph{IEEE
  Transactions on Image Processing}, 2022, 31: 3267--3280.

\bibitem{StyleTransformer_hu2022style}
Hu X, Huang Q, Shi Z, Li S, Gao C, Sun L, Li Q. Style transformer for image
  inversion and editing. In \emph{Proceedings of the IEEE/CVF Conference on
  Computer Vision and Pattern Recognition}, 2022, 11337--11346.

\bibitem{Interfacegan_shen2020interfacegan}
Shen Y, Yang C, Tang X, Zhou B. Interfacegan: Interpreting the disentangled
  face representation learned by gans. \emph{IEEE transactions on pattern
  analysis and machine intelligence}, 2020, 44(4): 2004--2018.

\bibitem{SeFa_shen2021closed}
Shen Y, Zhou B. Closed-form factorization of latent semantics in gans. In
  \emph{Proceedings of the IEEE/CVF conference on computer vision and pattern
  recognition}, 2021, 1532--1540.

\bibitem{HijackGAN_wang2021hijack}
Wang HP, Yu N, Fritz M. Hijack-gan: Unintended-use of pretrained, black-box
  gans. In \emph{Proceedings of the IEEE/CVF conference on computer vision and
  pattern recognition}, 2021, 7872--7881.

\bibitem{Stylegan2Distillation_viazovetskyi2020stylegan2}
Viazovetskyi Y, Ivashkin V, Kashin E. Stylegan2 distillation for feed-forward
  image manipulation. In \emph{Computer Vision--ECCV 2020: 16th European
  Conference, Glasgow, UK, August 23--28, 2020, Proceedings, Part XXII 16},
  2020, 170--186.

\bibitem{StyleFlow_abdal2021styleflow}
Abdal R, Zhu P, Mitra NJ, Wonka P. Styleflow: Attribute-conditioned exploration
  of stylegan-generated images using conditional continuous normalizing flows.
  \emph{ACM Transactions on Graphics (ToG)}, 2021, 40(3): 1--21.

\bibitem{parihar2023exploring}
Parihar R, Balaji P, Magazine R, Vora S, Karmali T, Jampani V, Babu RV.
  Exploring Attribute Variations in Style-based GANs using Diffusion Models.
  \emph{arXiv preprint arXiv:2311.16052}, 2023.

\bibitem{HessianPenalty_peebles2020hessian}
Peebles W, Peebles J, Zhu JY, Efros A, Torralba A. The hessian penalty: A weak
  prior for unsupervised disentanglement. In \emph{Computer Vision--ECCV 2020:
  16th European Conference, Glasgow, UK, August 23--28, 2020, Proceedings, Part
  VI 16}, 2020, 581--597.

\bibitem{SpectralRegularizer_ramesh2018spectral}
Ramesh A, Choi Y, LeCun Y. A spectral regularizer for unsupervised
  disentanglement. \emph{arXiv preprint arXiv:1812.01161}, 2018.

\bibitem{voynov2020unsupervised}
Voynov A, Babenko A. Unsupervised discovery of interpretable directions in the
  gan latent space. In \emph{International conference on machine learning},
  2020, 9786--9796.

\bibitem{WarpedGANSpace_tzelepis2021warpedganspace}
Tzelepis C, Tzimiropoulos G, Patras I. Warpedganspace: Finding non-linear rbf
  paths in gan latent space. In \emph{Proceedings of the IEEE/CVF international
  conference on computer vision}, 2021, 6393--6402.

\bibitem{Ganspace_harkonen2020ganspace}
H{\"a}rk{\"o}nen E, Hertzmann A, Lehtinen J, Paris S. Ganspace: Discovering
  interpretable gan controls. \emph{Advances in neural information processing
  systems}, 2020, 33: 9841--9850.

\bibitem{DragGAN_pan2023drag}
Pan X, Tewari A, Leimk{\"u}hler T, Liu L, Meka A, Theobalt C. Drag your gan:
  Interactive point-based manipulation on the generative image manifold. In
  \emph{ACM SIGGRAPH 2023 Conference Proceedings}, 2023, 1--11.

\bibitem{Styleclip_patashnik2021styleclip}
Patashnik O, Wu Z, Shechtman E, Cohen-Or D, Lischinski D. Styleclip:
  Text-driven manipulation of stylegan imagery. In \emph{Proceedings of the
  IEEE/CVF international conference on computer vision}, 2021, 2085--2094.

\bibitem{Stylemc_kocasari2022stylemc}
Kocasari U, Dirik A, Tiftikci M, Yanardag P. Stylemc: Multi-channel based fast
  text-guided image generation and manipulation. In \emph{Proceedings of the
  IEEE/CVF Winter Conference on Applications of Computer vision}, 2022,
  895--904.

\bibitem{PPE_xu2022predict}
Xu Z, Lin T, Tang H, Li F, He D, Sebe N, Timofte R, Van~Gool L, Ding E.
  Predict, prevent, and evaluate: Disentangled text-driven image manipulation
  empowered by pre-trained vision-language model. In \emph{Proceedings of the
  IEEE/CVF Conference on Computer Vision and Pattern Recognition}, 2022,
  18229--18238.

\bibitem{zheng2022bridging}
Zheng W, Li Q, Guo X, Wan P, Wang Z. Bridging clip and stylegan through latent
  alignment for image editing. \emph{arXiv preprint arXiv:2210.04506}, 2022.

\bibitem{FFCLIP_zhu2022one}
Zhu Y, Liu H, Song Y, Yuan Z, Han X, Yuan C, Chen Q, Wang J. One model to edit
  them all: Free-form text-driven image manipulation with semantic modulations.
  \emph{Advances in Neural Information Processing Systems}, 2022, 35:
  25146--25159.

\bibitem{Deltaedit_lyu2023deltaedit}
Lyu Y, Lin T, Li F, He D, Dong J, Tan T. Deltaedit: Exploring text-free
  training for text-driven image manipulation. \emph{arXiv preprint
  arXiv:2303.06285}, 2023.

\bibitem{Hairclip_wei2022hairclip}
Wei T, Chen D, Zhou W, Liao J, Tan Z, Yuan L, Zhang W, Yu N. Hairclip: Design
  your hair by text and reference image. In \emph{Proceedings of the IEEE/CVF
  Conference on Computer Vision and Pattern Recognition}, 2022, 18072--18081.

\bibitem{StyleganNada_gal2022stylegan}
Gal R, Patashnik O, Maron H, Bermano AH, Chechik G, Cohen-Or D. Stylegan-nada:
  Clip-guided domain adaptation of image generators. \emph{ACM Transactions on
  Graphics (TOG)}, 2022, 41(4): 1--13.

\bibitem{MindTheGap_zhu2021mind}
Zhu P, Abdal R, Femiani J, Wonka P. Mind the gap: Domain gap control for single
  shot domain adaptation for generative adversarial networks. \emph{arXiv
  preprint arXiv:2110.08398}, 2021.

\bibitem{DiFa_zhang2022towards}
Zhang Y, Yao M, Wei Y, Ji Z, Bai J, Zuo W, et~al.. Towards diverse and faithful
  one-shot adaption of generative adversarial networks. \emph{Advances in
  Neural Information Processing Systems}, 2022, 35: 37297--37308.

\bibitem{HyperGANCLIP_anees2024hypergan}
Anees AB, Baykal AC, Kizil MB, Ceylan D, Erdem E, Erdem A. HyperGAN-CLIP: A
  Unified Framework for Domain Adaptation, Image Synthesis and Manipulation. In
  \emph{SIGGRAPH Asia 2024 Conference Papers}, 2024, 1--12.

\bibitem{DDIM_song2020denoising}
Song J, Meng C, Ermon S. Denoising diffusion implicit models. \emph{arXiv
  preprint arXiv:2010.02502}, 2020.

\bibitem{NullInversion_mokady2023null}
Mokady R, Hertz A, Aberman K, Pritch Y, Cohen-Or D. Null-text inversion for
  editing real images using guided diffusion models. In \emph{Proceedings of
  the IEEE/CVF Conference on Computer Vision and Pattern Recognition}, 2023,
  6038--6047.

\bibitem{Catversion_zhao2023catversion}
Zhao R, Zhu M, Dong S, Wang N, Gao X. Catversion: Concatenating embeddings for
  diffusion-based text-to-image personalization. \emph{arXiv preprint
  arXiv:2311.14631}, 2023.

\bibitem{HiPer_han2023highly}
Han I, Yang S, Kwon T, Ye JC. Highly personalized text embedding for image
  manipulation by stable diffusion. \emph{arXiv preprint arXiv:2303.08767},
  2023.

\bibitem{P+_voynov2023p+}
Voynov A, Chu Q, Cohen-Or D, Aberman K. p+: Extended textual conditioning in
  text-to-image generation. \emph{arXiv preprint arXiv:2303.09522}, 2023.

\bibitem{NeTI_alaluf2023neural}
Alaluf Y, Richardson E, Metzer G, Cohen-Or D. A neural space-time
  representation for text-to-image personalization. \emph{ACM Transactions on
  Graphics (TOG)}, 2023, 42(6): 1--10.

\bibitem{CelebBasis_yuan2023inserting}
Yuan G, Cun X, Zhang Y, Li M, Qi C, Wang X, Shan Y, Zheng H. Inserting anybody
  in diffusion models via celeb basis. \emph{arXiv preprint arXiv:2306.00926},
  2023.

\bibitem{Instantbooth_shi2024instantbooth}
Shi J, Xiong W, Lin Z, Jung HJ. Instantbooth: Personalized text-to-image
  generation without test-time finetuning. In \emph{Proceedings of the IEEE/CVF
  Conference on Computer Vision and Pattern Recognition}, 2024, 8543--8552.

\bibitem{Cones_liu2023cones}
Liu Z, Feng R, Zhu K, Zhang Y, Zheng K, Liu Y, Zhao D, Zhou J, Cao Y. Cones:
  Concept neurons in diffusion models for customized generation. \emph{arXiv
  preprint arXiv:2303.05125}, 2023.

\bibitem{Perfusion_tewel2023key}
Tewel Y, Gal R, Chechik G, Atzmon Y. Key-locked rank one editing for
  text-to-image personalization. In \emph{ACM SIGGRAPH 2023 Conference
  Proceedings}, 2023, 1--11.

\bibitem{Lora_hu2021lora}
Hu EJ, Shen Y, Wallis P, Allen-Zhu Z, Li Y, Wang S, Wang L, Chen W. Lora:
  Low-rank adaptation of large language models. \emph{arXiv preprint
  arXiv:2106.09685}, 2021.

\bibitem{Ledits_tsaban2023ledits}
Tsaban L, Passos A. Ledits: Real image editing with ddpm inversion and semantic
  guidance. \emph{arXiv preprint arXiv:2307.00522}, 2023.

\bibitem{rout2024semantic}
Rout L, Chen Y, Ruiz N, Caramanis C, Shakkottai S, Chu WS. Semantic image
  inversion and editing using rectified stochastic differential equations.
  \emph{arXiv preprint arXiv:2410.10792}, 2024.

\bibitem{DSG_epstein2023diffusion}
Epstein D, Jabri A, Poole B, Efros A, Holynski A. Diffusion self-guidance for
  controllable image generation. \emph{Advances in Neural Information
  Processing Systems}, 2023, 36: 16222--16239.

\bibitem{PickandDraw_lv2024pick}
Lv H, Xiao J, Li L. Pick-and-draw: Training-free semantic guidance for
  text-to-image personalization. In \emph{Proceedings of the 32nd ACM
  International Conference on Multimedia}, 2024, 10535--10543.

\bibitem{ProFusion_zhou2023enhancing}
Zhou Y, Zhang R, Sun T, Xu J. Enhancing detail preservation for customized
  text-to-image generation: A regularization-free approach. \emph{arXiv
  preprint arXiv:2305.13579}, 2023.

\bibitem{MOMA_song2025moma}
Song K, Zhu Y, Liu B, Yan Q, Elgammal A, Yang X. Moma: Multimodal llm adapter
  for fast personalized image generation. In \emph{European Conference on
  Computer Vision}, 2025, 117--132.

\bibitem{SubjectDiffusion_ma2024subject}
Ma J, Liang J, Chen C, Lu H. Subject-diffusion: Open domain personalized
  text-to-image generation without test-time fine-tuning. In \emph{ACM SIGGRAPH
  2024 Conference Papers}, 2024, 1--12.

\bibitem{Fastcomposer_xiao2024fastcomposer}
Xiao G, Yin T, Freeman WT, Durand F, Han S. Fastcomposer: Tuning-free
  multi-subject image generation with localized attention. \emph{International
  Journal of Computer Vision}, 2024: 1--20.

\bibitem{BLIPDiffusion_li2024blip}
Li D, Li J, Hoi S. Blip-diffusion: Pre-trained subject representation for
  controllable text-to-image generation and editing. \emph{Advances in Neural
  Information Processing Systems}, 2024, 36.

\bibitem{HybridBooth_guan2025hybridbooth}
Guan S, Ge Y, Tai Y, Yang J, Li W, You M. HybridBooth: Hybrid Prompt Inversion
  for Efficient Subject-Driven Generation. In \emph{European Conference on
  Computer Vision}, 2025, 403--419.

\bibitem{ViCo_hao2023vico}
Hao S, Han K, Zhao S, Wong KYK. ViCo: Plug-and-play Visual Condition for
  Personalized Text-to-image Generation. \emph{arXiv preprint
  arXiv:2306.00971}, 2023.

\bibitem{HiFiTuner_wang2023hifi}
Wang Z, Wei W, Zhao Y, Xiao Z, Hasegawa-Johnson M, Shi H, Hou T. Hifi tuner:
  High-fidelity subject-driven fine-tuning for diffusion models. \emph{arXiv
  preprint arXiv:2312.00079}, 2023.

\bibitem{NestedAttention_patashnik2025nested}
Patashnik O, Gal R, Ostashev D, Tulyakov S, Aberman K, Cohen-Or D. Nested
  Attention: Semantic-aware Attention Values for Concept Personalization.
  \emph{arXiv preprint arXiv:2501.01407}, 2025.

\bibitem{OFT_qiu2023controlling}
Qiu Z, Liu W, Feng H, Xue Y, Feng Y, Liu Z, Zhang D, Weller A, Sch{\"o}lkopf B.
  Controlling text-to-image diffusion by orthogonal finetuning. \emph{Advances
  in Neural Information Processing Systems}, 2023, 36: 79320--79362.

\bibitem{BOFT_liu2023parameter}
Liu W, Qiu Z, Feng Y, Xiu Y, Xue Y, Yu L, Feng H, Liu Z, Heo J, Peng S, et~al..
  Parameter-efficient orthogonal finetuning via butterfly factorization.
  \emph{arXiv preprint arXiv:2311.06243}, 2023.

\bibitem{Disenbooth_chen2023disenbooth}
Chen H, Zhang Y, Wang X, Duan X, Zhou Y, Zhu W. Disenbooth: Disentangled
  parameter-efficient tuning for subject-driven text-to-image generation.
  \emph{arXiv preprint arXiv:2305.03374}, 2023, 3.

\bibitem{DETEX_cai2024decoupled}
Cai Y, Wei Y, Ji Z, Bai J, Han H, Zuo W. Decoupled textual embeddings for
  customized image generation. In \emph{Proceedings of the AAAI Conference on
  Artificial Intelligence}, 2024, 909--917.

\bibitem{Dreamtuner_hua2023dreamtuner}
Hua M, Liu J, Ding F, Liu W, Wu J, He Q. Dreamtuner: Single image is enough for
  subject-driven generation. \emph{arXiv preprint arXiv:2312.13691}, 2023.

\bibitem{BreakAScene_avrahami2023break}
Avrahami O, Aberman K, Fried O, Cohen-Or D, Lischinski D. Break-a-scene:
  Extracting multiple concepts from a single image. In \emph{SIGGRAPH Asia 2023
  Conference Papers}, 2023, 1--12.

\bibitem{Dreammatcher_nam2024dreammatcher}
Nam J, Kim H, Lee D, Jin S, Kim S, Chang S. Dreammatcher: Appearance matching
  self-attention for semantically-consistent text-to-image personalization. In
  \emph{Proceedings of the IEEE/CVF Conference on Computer Vision and Pattern
  Recognition}, 2024, 8100--8110.

\bibitem{ComFusion_hong2024comfusion}
Hong Y, Zhang J. ComFusion: Personalized Subject Generation in Multiple
  Specific Scenes From Single Image. \emph{arXiv preprint arXiv:2402.11849},
  2024.

\bibitem{Singleinsert_wu2023singleinsert}
Wu Z, Yu C, Zhu Z, Wang F, Bai X. Singleinsert: Inserting new concepts from a
  single image into text-to-image models for flexible editing. \emph{arXiv
  preprint arXiv:2310.08094}, 2023.

\bibitem{DCO_lee2024direct}
Lee K, Kwak S, Sohn K, Shin J. Direct consistency optimization for
  compositional text-to-image personalization. \emph{arXiv preprint
  arXiv:2402.12004}, 2024.

\bibitem{Instructbooth_chae2023instructbooth}
Chae D, Park N, Kim J, Lee K. Instructbooth: Instruction-following personalized
  text-to-image generation. \emph{arXiv preprint arXiv:2312.03011}, 2023.

\bibitem{CLIPconverter_ding2024clip}
Ding Y, Tian C, Ding H, Liu L. The CLIP model is secretly an image-to-prompt
  converter. \emph{Advances in Neural Information Processing Systems}, 2024,
  36.

\bibitem{FaceChainSuDe_qiao2024facechain}
Qiao P, Shang L, Liu C, Sun B, Ji X, Chen J. FaceChain-SuDe: Building Derived
  Class to Inherit Category Attributes for One-shot Subject-Driven Generation.
  In \emph{Proceedings of the IEEE/CVF Conference on Computer Vision and
  Pattern Recognition}, 2024, 7215--7224.

\bibitem{CoRe_wu2024core}
Wu F, Pang Y, Zhang J, Pang L, Yin J, Zhao B, Li Q, Mao X. CoRe:
  Context-Regularized Text Embedding Learning for Text-to-Image
  Personalization. \emph{arXiv preprint arXiv:2408.15914}, 2024.

\bibitem{DreamBlend_ram2025dreamblend}
Ram S, Neiman T, Feng Q, Stuart A, Tran S, Chilimbi T. DreamBlend: Advancing
  personalized fine-tuning of text-to-image diffusion models, 2025.

\bibitem{UMM_ma2023unified}
Ma Y, Yang H, Wang W, Fu J, Liu J. Unified multi-modal latent diffusion for
  joint subject and text conditional image generation. \emph{arXiv preprint
  arXiv:2303.09319}, 2023.

\bibitem{Bootpig_purushwalkam2024bootpig}
Purushwalkam S, Gokul A, Joty S, Naik N. Bootpig: Bootstrapping zero-shot
  personalized image generation capabilities in pretrained diffusion models.
  \emph{arXiv preprint arXiv:2401.13974}, 2024.

\bibitem{Taming_jia2023taming}
Jia X, Zhao Y, Chan KC, Li Y, Zhang H, Gong B, Hou T, Wang H, Su YC. Taming
  encoder for zero fine-tuning image customization with text-to-image diffusion
  models. \emph{arXiv preprint arXiv:2304.02642}, 2023.

\bibitem{CustomizationAssistant_zhou2024customization}
Zhou Y, Zhang R, Gu J, Sun T. Customization assistant for text-to-image
  generation. In \emph{Proceedings of the IEEE/CVF Conference on Computer
  Vision and Pattern Recognition}, 2024, 9182--9191.

\bibitem{SAG_chan2024improving}
Chan KC, Zhao Y, Jia X, Yang MH, Wang H. Improving Subject-Driven Image
  Synthesis with Subject-Agnostic Guidance. In \emph{Proceedings of the
  IEEE/CVF Conference on Computer Vision and Pattern Recognition}, 2024,
  6733--6742.

\bibitem{CustomContrast_chen2024customcontrast}
Chen N, Huang M, Chen Z, Zheng Y, Zhang L, Mao Z. CustomContrast: A Multilevel
  Contrastive Perspective For Subject-Driven Text-to-Image Customization.
  \emph{arXiv preprint arXiv:2409.05606}, 2024.

\bibitem{AnyDressing_li2024anydressing}
Li X, Sun Q, Zhang P, Ye F, Liao Z, Feng W, Zhao S, He Q. AnyDressing:
  Customizable Multi-Garment Virtual Dressing via Latent Diffusion Models.
  \emph{arXiv preprint arXiv:2412.04146}, 2024.

\bibitem{DisEnvisioner_he2024disenvisioner}
He J, Li H, Hu Y, Shen G, Cai Y, Qiu W, Chen YC. DisEnvisioner: Disentangled
  and Enriched Visual Prompt for Customized Image Generation. \emph{arXiv
  preprint arXiv:2410.02067}, 2024.

\bibitem{song2024harmonizing}
Song Y, Kim J, Park W, Shin W, Rhee W, Kwak N. Harmonizing Visual and Textual
  Embeddings for Zero-Shot Text-to-Image Customization. \emph{arXiv preprint
  arXiv:2403.14155}, 2024.

\bibitem{JeDi_zeng2024jedi}
Zeng Y, Patel VM, Wang H, Huang X, Wang TC, Liu MY, Balaji Y. JeDi: Joint-Image
  Diffusion Models for Finetuning-Free Personalized Text-to-Image Generation.
  In \emph{Proceedings of the IEEE/CVF Conference on Computer Vision and
  Pattern Recognition}, 2024, 6786--6795.

\bibitem{SSREncoder_zhang2024ssr}
Zhang Y, Song Y, Liu J, Wang R, Yu J, Tang H, Li H, Tang X, Hu Y, Pan H,
  et~al.. Ssr-encoder: Encoding selective subject representation for
  subject-driven generation. In \emph{Proceedings of the IEEE/CVF Conference on
  Computer Vision and Pattern Recognition}, 2024, 8069--8078.

\bibitem{PALP_arar2024palp}
Arar M, Voynov A, Hertz A, Avrahami O, Fruchter S, Pritch Y, Cohen-Or D, Shamir
  A. PALP: prompt aligned personalization of text-to-image models. In
  \emph{SIGGRAPH Asia 2024 Conference Papers}, 2024, 1--11.

\bibitem{cai2024diffusion}
Cai S, Chan E, Zhang Y, Guibas L, Wu J, Wetzstein G. Diffusion
  Self-Distillation for Zero-Shot Customized Image Generation. \emph{arXiv
  preprint arXiv:2411.18616}, 2024.

\bibitem{InstructImagen_hu2024instruct}
Hu H, Chan KC, Su YC, Chen W, Li Y, Sohn K, Zhao Y, Ben X, Gong B, Cohen W,
  et~al.. Instruct-Imagen: Image generation with multi-modal instruction. In
  \emph{Proceedings of the IEEE/CVF Conference on Computer Vision and Pattern
  Recognition}, 2024, 4754--4763.

\bibitem{OminiControl_tan2024ominicontrol}
Tan Z, Liu S, Yang X, Xue Q, Wang X. OminiControl: Minimal and Universal
  Control for Diffusion Transformer. \emph{arXiv preprint arXiv:2411.15098},
  2024.

\bibitem{Omnigen_xiao2024omnigen}
Xiao S, Wang Y, Zhou J, Yuan H, Xing X, Yan R, Wang S, Huang T, Liu Z. Omnigen:
  Unified image generation. \emph{arXiv preprint arXiv:2409.11340}, 2024.

\bibitem{GFTI_fei2023gradient}
Fei Z, Fan M, Huang J. Gradient-free textual inversion. In \emph{Proceedings of
  the 31st ACM International Conference on Multimedia}, 2023, 1364--1373.

\bibitem{PRISM_he2024automated}
He Y, Robey A, Murata N, Jiang Y, Williams J, Pappas GJ, Hassani H, Mitsufuji
  Y, Salakhutdinov R, Kolter JZ. Automated Black-box Prompt Engineering for
  Personalized Text-to-Image Generation. \emph{arXiv preprint
  arXiv:2403.19103}, 2024.

\bibitem{Dreamidentity_chen2023dreamidentity}
Chen Z, Fang S, Liu W, He Q, Huang M, Zhang Y, Mao Z. Dreamidentity: Improved
  editability for efficient face-identity preserved image generation.
  \emph{arXiv preprint arXiv:2307.00300}, 2023.

\bibitem{Stableidentity_wang2024stableidentity}
Wang Q, Jia X, Li X, Li T, Ma L, Zhuge Y, Lu H. Stableidentity: Inserting
  anybody into anywhere at first sight. \emph{arXiv preprint arXiv:2401.15975},
  2024.

\bibitem{ID-Booth_tomavsevicid}
Toma{\v{s}}evi{\'c} D, Boutros F, Damer N, Struc V, Peer P. ID-Booth:
  Identity-consistent image generation with diffusion models.

\bibitem{Difflora_wu2024difflora}
Wu Y, Shi Y, Wei J, Sun C, Zhou Y, Yang Y, Shen HT. Difflora: Generating
  personalized low-rank adaptation weights with diffusion. \emph{arXiv preprint
  arXiv:2408.06740}, 2024.

\bibitem{CrossInitialization_pang2024cross}
Pang L, Yin J, Xie H, Wang Q, Li Q, Mao X. Cross Initialization for Face
  Personalization of Text-to-Image Models. In \emph{Proceedings of the IEEE/CVF
  Conference on Computer Vision and Pattern Recognition}, 2024, 8393--8403.

\bibitem{Instantid_wang2024instantid}
Wang Q, Bai X, Wang H, Qin Z, Chen A, Li H, Tang X, Hu Y. Instantid: Zero-shot
  identity-preserving generation in seconds. \emph{arXiv preprint
  arXiv:2401.07519}, 2024.

\bibitem{Face2Diffusion_shiohara2024face2diffusion}
Shiohara K, Yamasaki T. Face2Diffusion for Fast and Editable Face
  Personalization. In \emph{Proceedings of the IEEE/CVF Conference on Computer
  Vision and Pattern Recognition}, 2024, 6850--6859.

\bibitem{Photoverse_chen2023photoverse}
Chen L, Zhao M, Liu Y, Ding M, Song Y, Wang S, Wang X, Yang H, Liu J, Du K,
  et~al.. Photoverse: Tuning-free image customization with text-to-image
  diffusion models. \emph{arXiv preprint arXiv:2309.05793}, 2023.

\bibitem{Pulid_guo2024pulid}
Guo Z, Wu Y, Chen Z, Chen L, Zhang P, He Q. Pulid: Pure and lightning id
  customization via contrastive alignment. \emph{arXiv preprint
  arXiv:2404.16022}, 2024.

\bibitem{Portraitbooth_peng2024portraitbooth}
Peng X, Zhu J, Jiang B, Tai Y, Luo D, Zhang J, Lin W, Jin T, Wang C, Ji R.
  Portraitbooth: A versatile portrait model for fast identity-preserved
  personalization. In \emph{Proceedings of the IEEE/CVF Conference on Computer
  Vision and Pattern Recognition}, 2024, 27080--27090.

\bibitem{Lcm-lookahead_gal2024lcm}
Gal R, Lichter O, Richardson E, Patashnik O, Bermano AH, Chechik G, Cohen-Or D.
  Lcm-lookahead for encoder-based text-to-image personalization. \emph{arXiv
  preprint arXiv:2404.03620}, 2024, 2(3): 4.

\bibitem{wplus_li2024stylegan}
Li X, Hou X, Loy CC. When stylegan meets stable diffusion: a w+ adapter for
  personalized image generation. In \emph{Proceedings of the IEEE/CVF
  Conference on Computer Vision and Pattern Recognition}, 2024, 2187--2196.

\bibitem{Precisecontrol_parihar2025precisecontrol}
Parihar R, Sachidanand V, Mani S, Karmali T, Venkatesh~Babu R. Precisecontrol:
  Enhancing text-to-image diffusion models with fine-grained attribute control.
  In \emph{European Conference on Computer Vision}, 2025, 469--487.

\bibitem{FlashFace_zhang2024flashface}
Zhang S, Huang L, Chen X, Zhang Y, Wu ZF, Feng Y, Wang W, Shen Y, Liu Y, Luo P.
  FlashFace: Human Image Personalization with High-fidelity Identity
  Preservation. \emph{arXiv preprint arXiv:2403.17008}, 2024.

\bibitem{Infinite-ID_wu2025infinite}
Wu Y, Li Z, Zheng H, Wang C, Li B. Infinite-ID: Identity-preserved
  Personalization via ID-semantics Decoupling Paradigm. In \emph{European
  Conference on Computer Vision}, 2025, 279--296.

\bibitem{MasterWeaver_wei2025masterweaver}
Wei Y, Ji Z, Bai J, Zhang H, Zhang L, Zuo W. MasterWeaver: Taming Editability
  and Face Identity for Personalized Text-to-Image Generation. In
  \emph{European Conference on Computer Vision}, 2025, 252--271.

\bibitem{Imagine-yourself_he2024imagine}
He Z, Sun B, Juefei-Xu F, Ma H, Ramchandani A, Cheung V, Shah S, Kalia A,
  Subramanyam H, Zareian A, et~al.. Imagine yourself: Tuning-free personalized
  image generation. \emph{arXiv preprint arXiv:2409.13346}, 2024.

\bibitem{Consistentid_huang2024consistentid}
Huang J, Dong X, Song W, Li H, Zhou J, Cheng Y, Liao S, Chen L, Yan Y, Liao S,
  et~al.. Consistentid: Portrait generation with multimodal fine-grained
  identity preserving. \emph{arXiv preprint arXiv:2404.16771}, 2024.

\bibitem{Caphuman_liang2024caphuman}
Liang C, Ma F, Zhu L, Deng Y, Yang Y. Caphuman: Capture your moments in
  parallel universes. In \emph{Proceedings of the IEEE/CVF Conference on
  Computer Vision and Pattern Recognition}, 2024, 6400--6409.

\bibitem{StoryMaker_zhou2024storymaker}
Zhou Z, Li J, Li H, Chen N, Tang X. StoryMaker: Towards Holistic Consistent
  Characters in Text-to-image Generation. \emph{arXiv preprint
  arXiv:2409.12576}, 2024.

\bibitem{Character-Adapter_ma2024character}
Ma Y, Xu W, Tang J, Jin Q, Zhang R, Zhao Z, Fan C, Hu Z. Character-Adapter:
  Prompt-Guided Region Control for High-Fidelity Character Customization.
  \emph{arXiv preprint arXiv:2406.16537}, 2024.

\bibitem{RETRIBOORU_tang2023retrieving}
Tang H, Zhou X, Deng J, Pan Z, Tian H, Chaudhari P. Retrieving conditions from
  reference images for diffusion models. \emph{arXiv preprint
  arXiv:2312.02521}, 2023.

\bibitem{SerialGen_xie2024serialgen}
Xie C, Zou H, Yu R, Zhang Y, Zhan Z. SerialGen: Personalized Image Generation
  by First Standardization Then Personalization. \emph{arXiv preprint
  arXiv:2412.01485}, 2024.

\bibitem{AnyStory_he2025anystory}
He J, Tuo Y, Chen B, Zhong C, Geng Y, Bo L. AnyStory: Towards Unified Single
  and Multiple Subject Personalization in Text-to-Image Generation. \emph{arXiv
  preprint arXiv:2501.09503}, 2025.

\bibitem{InST_zhang2023inversion}
Zhang Y, Huang N, Tang F, Huang H, Ma C, Dong W, Xu C. Inversion-based style
  transfer with diffusion models. In \emph{Proceedings of the IEEE/CVF
  conference on computer vision and pattern recognition}, 2023, 10146--10156.

\bibitem{CSGO_imagecsgo}
Image S. CSGO: Content-Style Composition in Text-to-Image Generation.

\bibitem{StyleBoost_park2023styleboost}
Park J, Ko B, Jang H. StyleBoost: A Study of Personalizing Text-to-Image
  Generation in Any Style using DreamBooth. In \emph{2023 14th International
  Conference on Information and Communication Technology Convergence (ICTC)},
  2023, 93--98.

\bibitem{StyleForge_park2024text}
Park J, Ko B, Jang H. Text-to-Image Synthesis for Any Artistic Styles:
  Advancements in Personalized Artistic Image Generation via Subdivision and
  Dual Binding. \emph{arXiv preprint arXiv:2404.05256}, 2024.

\bibitem{PairCustomization_jones2024customizing}
Jones M, Wang SY, Kumari N, Bau D, Zhu JY. Customizing text-to-image models
  with a single image pair. In \emph{SIGGRAPH Asia 2024 Conference Papers},
  2024, 1--13.

\bibitem{U-VAP_wu2024u}
Wu Y, Liu K, Mi X, Tang F, Cao J, Li J. U-VAP: User-specified Visual Appearance
  Personalization via Decoupled Self Augmentation. In \emph{Proceedings of the
  IEEE/CVF Conference on Computer Vision and Pattern Recognition}, 2024,
  9482--9491.

\bibitem{Ziplora_shah2025ziplora}
Shah V, Ruiz N, Cole F, Lu E, Lazebnik S, Li Y, Jampani V. Ziplora: Any subject
  in any style by effectively merging loras. In \emph{European Conference on
  Computer Vision}, 2025, 422--438.

\bibitem{UnZipLoRA_liu2024unziplora}
Liu C, Shah V, Cui A, Lazebnik S. UnZipLoRA: Separating Content and Style from
  a Single Image. \emph{arXiv preprint arXiv:2412.04465}, 2024.

\bibitem{StyleAdapter_wang2024styleadapter}
Wang Z, Wang X, Xie L, Qi Z, Shan Y, Wang W, Luo P. StyleAdapter: A Unified
  Stylized Image Generation Model. \emph{International Journal of Computer
  Vision}, 2024: 1--18.

\bibitem{ArtAdapter_chen2024artadapter}
Chen DY, Tennent H, Hsu CW. ArtAdapter: Text-to-Image Style Transfer using
  Multi-Level Style Encoder and Explicit Adaptation. In \emph{Proceedings of
  the IEEE/CVF Conference on Computer Vision and Pattern Recognition}, 2024,
  8619--8628.

\bibitem{FineStyle_zhangfinestyle}
Zhang G, Sohn K, Hahn M, Shi H, Essa I. FineStyle: Fine-grained Controllable
  Style Personalization for Text-to-image Models. In \emph{The Thirty-eighth
  Annual Conference on Neural Information Processing Systems}.

\bibitem{Artistic-Intelligence_yang2024artistic}
Yang W, Zhao Y. Artistic Intelligence: A Diffusion-Based Framework for
  High-Fidelity Landscape Painting Synthesis. \emph{IEEE Access}, 2024.

\bibitem{ReVersion_huang2024reversion}
Huang Z, Wu T, Jiang Y, Chan KC, Liu Z. ReVersion: Diffusion-based relation
  inversion from images. In \emph{SIGGRAPH Asia 2024 Conference Papers}, 2024,
  1--11.

\bibitem{Lego_motamed2023lego}
Motamed S, Paudel DP, Van~Gool L. Lego: Learning to disentangle and invert
  concepts beyond object appearance in text-to-image diffusion models.
  \emph{arXiv preprint arXiv:2311.13833}, 2023.

\bibitem{ADI_huang2024learning}
Huang S, Gong B, Feng Y, Chen X, Fu Y, Liu Y, Wang D. Learning disentangled
  identifiers for action-customized text-to-image generation. In
  \emph{Proceedings of the IEEE/CVF Conference on Computer Vision and Pattern
  Recognition}, 2024, 7797--7806.

\bibitem{ImPoster_kothandaraman2024imposter}
Kothandaraman D, Kulkarni K, Shekhar S, Srinivasan BV, Manocha D. ImPoster:
  Text and Frequency Guidance for Subject Driven Action Personalization using
  Diffusion Models. \emph{arXiv preprint arXiv:2409.15650}, 2024.

\bibitem{FreeEvent_wang2024event}
Wang Z, Jiang Y, Zheng D, Xiao J, Chen L. Event-Customized Image Generation.
  \emph{arXiv preprint arXiv:2410.02483}, 2024.

\bibitem{Svdiff_han2023svdiff}
Han L, Li Y, Zhang H, Milanfar P, Metaxas D, Yang F. Svdiff: Compact parameter
  space for diffusion fine-tuning. In \emph{Proceedings of the IEEE/CVF
  International Conference on Computer Vision}, 2023, 7323--7334.

\bibitem{Cones2_liu2024customizable}
Liu Z, Zhang Y, Shen Y, Zheng K, Zhu K, Feng R, Liu Y, Zhao D, Zhou J, Cao Y.
  Customizable image synthesis with multiple subjects. \emph{Advances in Neural
  Information Processing Systems}, 2024, 36.

\bibitem{MC2_jiang2024mc}
Jiang J, Zhang Y, Feng K, Wu X, Zuo W. MC$^2$: Multi-concept Guidance for
  Customized Multi-concept Generation. \emph{arXiv preprint arXiv:2404.05268},
  2024.

\bibitem{Omg_kong2025omg}
Kong Z, Zhang Y, Yang T, Wang T, Zhang K, Wu B, Chen G, Liu W, Luo W. Omg:
  Occlusion-friendly personalized multi-concept generation in diffusion models.
  In \emph{European Conference on Computer Vision}, 2025, 253--270.

\bibitem{ConceptWeaver_kwon2024concept}
Kwon G, Jenni S, Li D, Lee JY, Ye JC, Heilbron FC. Concept Weaver: Enabling
  Multi-Concept Fusion in Text-to-Image Models. In \emph{Proceedings of the
  IEEE/CVF Conference on Computer Vision and Pattern Recognition}, 2024,
  8880--8889.

\bibitem{CIDM_dong2024continually}
Dong J, Liang W, Li H, Zhang D, Cao M, Ding H, Khan S, Khan FS. How to
  Continually Adapt Text-to-Image Diffusion Models for Flexible Customization?
  \emph{arXiv preprint arXiv:2410.17594}, 2024.

\bibitem{SEGuidance_liu2024training}
Liu S, Wang B, Ma Y, Yang T, Cao X, Chen Q, Li H, Dong D, Jiang P.
  Training-free Subject-Enhanced Attention Guidance for Compositional
  Text-to-image Generation. \emph{arXiv preprint arXiv:2405.06948}, 2024.

\bibitem{ConceptConductor_yao2024concept}
Yao Z, Feng F, Li R, Wang X. Concept Conductor: Orchestrating Multiple
  Personalized Concepts in Text-to-Image Synthesis. \emph{arXiv preprint
  arXiv:2408.03632}, 2024.

\bibitem{MagicTailor_zhou2024magictailor}
Zhou D, Huang J, Bai J, Wang J, Chen H, Chen G, Hu X, Heng PA. MagicTailor:
  Component-Controllable Personalization in Text-to-Image Diffusion Models.
  \emph{arXiv preprint arXiv:2410.13370}, 2024.

\bibitem{GroundingBooth_xiong2024groundingbooth}
Xiong Z, Xiong W, Shi J, Zhang H, Song Y, Jacobs N. GroundingBooth: Grounding
  Text-to-Image Customization. \emph{arXiv preprint arXiv:2409.08520}, 2024.

\bibitem{MS-Diffusion_personalizationms}
Personalization Ss. MS-Diffusion: Multi-subject Zero-shot Image Personalization
  with Layout Guidance.

\bibitem{RelationBooth_shi2024relationbooth}
Shi Q, Qi L, Wu J, Bai J, Wang J, Tong Y, Li X, Yang MH. RelationBooth: Towards
  Relation-Aware Customized Object Generation. \emph{arXiv preprint
  arXiv:2410.23280}, 2024.

\bibitem{TokenVerse_garibi2025tokenVerse}
Garibi D, Yadin S, Paiss R, Tov O, Zada S, Ephrat A, Michaeli T, Mosseri I,
  Dekel T. TokenVerse: Versatile Multi-concept Personalization in Token
  Modulation Space. \emph{arXiv preprint arXiv:2501.12224}, 2025.

\bibitem{DDS_hertz2023delta}
Hertz A, Aberman K, Cohen-Or D. Delta denoising score. In \emph{Proceedings of
  the IEEE/CVF International Conference on Computer Vision}, 2023, 2328--2337.

\bibitem{StableFlow_avrahami2024stable}
Avrahami O, Patashnik O, Fried O, Nemchinov E, Aberman K, Lischinski D,
  Cohen-Or D. Stable Flow: Vital Layers for Training-Free Image Editing.
  \emph{arXiv preprint arXiv:2411.14430}, 2024.

\bibitem{HeadRouter_xu2024headrouter}
Xu Y, Tang F, Cao J, Zhang Y, Kong X, Li J, Deussen O, Lee TY. HeadRouter: A
  Training-free Image Editing Framework for MM-DiTs by Adaptively Routing
  Attention Heads. \emph{arXiv preprint arXiv:2411.15034}, 2024.

\bibitem{xu2024unveil}
Xu P, Jiang B, Hu X, Luo D, He Q, Zhang J, Wang C, Wu Y, Ling C, Wang B. Unveil
  Inversion and Invariance in Flow Transformer for Versatile Image Editing.
  \emph{arXiv preprint arXiv:2411.15843}, 2024.

\bibitem{Emu3_wang2024emu3}
Wang X, Zhang X, Luo Z, Sun Q, Cui Y, Wang J, Zhang F, Wang Y, Li Z, Yu Q,
  et~al.. Emu3: Next-token prediction is all you need. \emph{arXiv preprint
  arXiv:2409.18869}, 2024.

\bibitem{SeedX_ge2024seed}
Ge Y, Zhao S, Zhu J, Ge Y, Yi K, Song L, Li C, Ding X, Shan Y. Seed-x:
  Multimodal models with unified multi-granularity comprehension and
  generation. \emph{arXiv preprint arXiv:2404.14396}, 2024.

\bibitem{MetaMorph_tong2024metamorph}
Tong S, Fan D, Zhu J, Xiong Y, Chen X, Sinha K, Rabbat M, LeCun Y, Xie S, Liu
  Z. MetaMorph: Multimodal Understanding and Generation via Instruction Tuning.
  \emph{arXiv preprint arXiv:2412.14164}, 2024.

\bibitem{Chameleon_team2024chameleon}
Team C. Chameleon: Mixed-modal early-fusion foundation models. \emph{arXiv
  preprint arXiv:2405.09818}, 2024.

\bibitem{Puma_fang2024puma}
Fang R, Duan C, Wang K, Li H, Tian H, Zeng X, Zhao R, Dai J, Li H, Liu X. Puma:
  Empowering unified mllm with multi-granular visual generation. \emph{arXiv
  preprint arXiv:2410.13861}, 2024.

\bibitem{Liquid_wu2024liquid}
Wu J, Jiang Y, Ma C, Liu Y, Zhao H, Yuan Z, Bai S, Bai X. Liquid: Language
  Models are Scalable Multi-modal Generators. \emph{arXiv preprint
  arXiv:2412.04332}, 2024.

\bibitem{ILLUME_wang2024illume}
Wang C, Lu G, Yang J, Huang R, Han J, Hou L, Zhang W, Xu H. ILLUME:
  Illuminating Your LLMs to See, Draw, and Self-Enhance. \emph{arXiv preprint
  arXiv:2412.06673}, 2024.

\bibitem{X-Prompt_sun2024x}
Sun Z, Chu Z, Zhang P, Wu T, Dong X, Zang Y, Xiong Y, Lin D, Wang J. X-Prompt:
  Towards Universal In-Context Image Generation in Auto-Regressive Vision
  Language Foundation Models. \emph{arXiv preprint arXiv:2412.01824}, 2024.

\bibitem{DCGAN_radford2015unsupervised}
Radford A. Unsupervised representation learning with deep convolutional
  generative adversarial networks. \emph{arXiv preprint arXiv:1511.06434},
  2015.

\bibitem{StackGAN_zhang2017stackgan}
Zhang H, Xu T, Li H, Zhang S, Wang X, Huang X, Metaxas DN. Stackgan: Text to
  photo-realistic image synthesis with stacked generative adversarial networks.
  In \emph{Proceedings of the IEEE international conference on computer
  vision}, 2017, 5907--5915.

\bibitem{StyleGAN_karras2019style}
Karras T, Laine S, Aila T. A style-based generator architecture for generative
  adversarial networks. In \emph{Proceedings of the IEEE/CVF conference on
  computer vision and pattern recognition}, 2019, 4401--4410.

\bibitem{StyleGAN2_karras2020analyzing}
Karras T, Laine S, Aittala M, Hellsten J, Lehtinen J, Aila T. Analyzing and
  improving the image quality of stylegan. In \emph{Proceedings of the IEEE/CVF
  conference on computer vision and pattern recognition}, 2020, 8110--8119.

\bibitem{LSGAN_mao2017least}
Mao X, Li Q, Xie H, Lau RY, Wang Z, Paul~Smolley S. Least squares generative
  adversarial networks. In \emph{Proceedings of the IEEE international
  conference on computer vision}, 2017, 2794--2802.

\bibitem{WGAN_arjovsky2017wasserstein}
Arjovsky M, Chintala S, Bottou L. Wasserstein generative adversarial networks.
  In \emph{International conference on machine learning}, 2017, 214--223.

\bibitem{WGANGP_gulrajani2017improved}
Gulrajani I, Ahmed F, Arjovsky M, Dumoulin V, Courville AC. Improved training
  of wasserstein gans. \emph{Advances in neural information processing
  systems}, 2017, 30.

\bibitem{SNGAN_miyato2018spectral}
Miyato T, Kataoka T, Koyama M, Yoshida Y. Spectral normalization for generative
  adversarial networks. \emph{arXiv preprint arXiv:1802.05957}, 2018.

\bibitem{PGGAN_karras2017progressive}
Karras T. Progressive Growing of GANs for Improved Quality, Stability, and
  Variation. \emph{arXiv preprint arXiv:1710.10196}, 2017.

\bibitem{StyleGAN3_karras2021alias}
Karras T, Aittala M, Laine S, H{\"a}rk{\"o}nen E, Hellsten J, Lehtinen J, Aila
  T. Alias-free generative adversarial networks. \emph{Advances in neural
  information processing systems}, 2021, 34: 852--863.

\bibitem{BigGAN_brock2018large}
Brock A, Donahue J, Simonyan K. Large scale GAN training for high fidelity
  natural image synthesis. arXiv 2018. \emph{arXiv preprint arXiv:1809.11096},
  2018.

\bibitem{StyleGANT_sauer2023stylegan}
Sauer A, Karras T, Laine S, Geiger A, Aila T. Stylegan-t: Unlocking the power
  of gans for fast large-scale text-to-image synthesis. In \emph{International
  conference on machine learning}, 2023, 30105--30118.

\bibitem{GigaGAN_kang2023scaling}
Kang M, Zhu JY, Zhang R, Park J, Shechtman E, Paris S, Park T. Scaling up gans
  for text-to-image synthesis. In \emph{Proceedings of the IEEE/CVF Conference
  on Computer Vision and Pattern Recognition}, 2023, 10124--10134.

\bibitem{ediffi_balaji2022ediff}
Balaji Y, Nah S, Huang X, Vahdat A, Song J, Zhang Q, Kreis K, Aittala M, Aila
  T, Laine S, et~al.. ediff-i: Text-to-image diffusion models with an ensemble
  of expert denoisers. \emph{arXiv preprint arXiv:2211.01324}, 2022.

\bibitem{SD3_esser2024scaling}
Esser P, Kulal S, Blattmann A, Entezari R, M{\"u}ller J, Saini H, Levi Y,
  Lorenz D, Sauer A, Boesel F, et~al.. Scaling rectified flow transformers for
  high-resolution image synthesis. In \emph{Forty-first International
  Conference on Machine Learning}, 2024.

\bibitem{kolors}
Team K. Kolors: Effective Training of Diffusion Model for Photorealistic
  Text-to-Image Synthesis. \emph{arXiv preprint}, 2024.

\bibitem{CLIP_radford2021learning}
Radford A, Kim JW, Hallacy C, Ramesh A, Goh G, Agarwal S, Sastry G, Askell A,
  Mishkin P, Clark J, et~al.. Learning transferable visual models from natural
  language supervision. In \emph{International conference on machine learning},
  2021, 8748--8763.

\bibitem{T5_ni2021sentence}
Ni J, Abrego GH, Constant N, Ma J, Hall KB, Cer D, Yang Y. Sentence-t5:
  Scalable sentence encoders from pre-trained text-to-text models. \emph{arXiv
  preprint arXiv:2108.08877}, 2021.

\bibitem{Laion400M_schuhmann2021laion}
Schuhmann C, Vencu R, Beaumont R, Kaczmarczyk R, Mullis C, Katta A, Coombes T,
  Jitsev J, Komatsuzaki A. Laion-400m: Open dataset of clip-filtered 400
  million image-text pairs. \emph{arXiv preprint arXiv:2111.02114}, 2021.

\bibitem{Laion5b_schuhmann2022laion}
Schuhmann C, Beaumont R, Vencu R, Gordon C, Wightman R, Cherti M, Coombes T,
  Katta A, Mullis C, Wortsman M, et~al.. Laion-5b: An open large-scale dataset
  for training next generation image-text models. \emph{Advances in Neural
  Information Processing Systems}, 2022, 35: 25278--25294.

\bibitem{flux2023}
Labs BF. FLUX. \url{https://github.com/black-forest-labs/flux}, 2024.

\bibitem{DIT_peebles2023scalable}
Peebles W, Xie S. Scalable diffusion models with transformers. In
  \emph{Proceedings of the IEEE/CVF International Conference on Computer
  Vision}, 2023, 4195--4205.

\bibitem{lipman2022flow}
Lipman Y, Chen RT, Ben-Hamu H, Nickel M, Le M. Flow matching for generative
  modeling. \emph{arXiv preprint arXiv:2210.02747}, 2022.

\bibitem{radford2018improving}
Radford A. Improving language understanding by generative pre-training, 2018.

\bibitem{GPT2_radford2019language}
Radford A, Wu J, Child R, Luan D, Amodei D, Sutskever I, et~al.. Language
  models are unsupervised multitask learners. \emph{OpenAI blog}, 2019, 1(8):
  9.

\bibitem{GPT3_brown2020language}
Brown T, Mann B, Ryder N, Subbiah M, Kaplan JD, Dhariwal P, Neelakantan A,
  Shyam P, Sastry G, Askell A, et~al.. Language models are few-shot learners.
  \emph{Advances in neural information processing systems}, 2020, 33:
  1877--1901.

\bibitem{LlamaGen_sun2024autoregressive}
Sun P, Jiang Y, Chen S, Zhang S, Peng B, Luo P, Yuan Z. Autoregressive Model
  Beats Diffusion: Llama for Scalable Image Generation. \emph{arXiv preprint
  arXiv:2406.06525}, 2024.

\bibitem{NextGPT_wu2023next}
Wu S, Fei H, Qu L, Ji W, Chua TS. Next-gpt: Any-to-any multimodal llm.
  \emph{arXiv preprint arXiv:2309.05519}, 2023.

\bibitem{Cogview2_ding2022cogview2}
Ding M, Zheng W, Hong W, Tang J. Cogview2: Faster and better text-to-image
  generation via hierarchical transformers. \emph{Advances in Neural
  Information Processing Systems}, 2022, 35: 16890--16902.

\bibitem{Parti_yu2022scaling}
Yu J, Xu Y, Koh JY, Luong T, Baid G, Wang Z, Vasudevan V, Ku A, Yang Y, Ayan
  BK, et~al.. Scaling autoregressive models for content-rich text-to-image
  generation. \emph{arXiv preprint arXiv:2206.10789}, 2022, 2(3): 5.

\bibitem{VAR_tian2024visual}
Tian K, Jiang Y, Yuan Z, Peng B, Wang L. Visual autoregressive modeling:
  Scalable image generation via next-scale prediction. \emph{arXiv preprint
  arXiv:2404.02905}, 2024.

\bibitem{STAR_ma2024star}
Ma X, Zhou M, Liang T, Bai Y, Zhao T, Chen H, Jin Y. STAR: Scale-wise
  Text-to-image generation via Auto-Regressive representations. \emph{arXiv
  preprint arXiv:2406.10797}, 2024.

\bibitem{Varclip_zhang2024var}
Zhang Q, Dai X, Yang N, An X, Feng Z, Ren X. Var-clip: Text-to-image generator
  with visual auto-regressive modeling. \emph{arXiv preprint arXiv:2408.01181},
  2024.

\bibitem{liu2019stgan}
Liu M, Ding Y, Xia M, Liu X, Ding E, Zuo W, Wen S. Stgan: A unified selective
  transfer network for arbitrary image attribute editing. In \emph{Proceedings
  of the IEEE/CVF conference on computer vision and pattern recognition}, 2019,
  3673--3682.

\bibitem{StyleSpace1_liu2020style}
Liu Y, Li Q, Sun Z, Tan T. Style intervention: How to achieve spatial
  disentanglement with style-based generators? \emph{arXiv preprint
  arXiv:2011.09699}, 2020.

\bibitem{bau2019inverting}
Bau D, Zhu JY, Wulff J, Peebles W, Strobelt H, Zhou B, Torralba A. Inverting
  layers of a large generator. In \emph{ICLR workshop}, 2019, 4.

\bibitem{PSpace_zhu2020improved}
Zhu P, Abdal R, Qin Y, Femiani J, Wonka P. Improved stylegan embedding: Where
  are the good latents? \emph{arXiv preprint arXiv:2012.09036}, 2020.

\bibitem{BDInvert_kang2021gan}
Kang K, Kim S, Cho S. Gan inversion for out-of-range images with geometric
  transformations. In \emph{Proceedings of the IEEE/CVF international
  conference on computer vision}, 2021, 13941--13949.

\bibitem{Arcface_deng2019arcface}
Deng J, Guo J, Xue N, Zafeiriou S. Arcface: Additive angular margin loss for
  deep face recognition. In \emph{Proceedings of the IEEE/CVF conference on
  computer vision and pattern recognition}, 2019, 4690--4699.

\bibitem{MOCO_he2020momentum}
He K, Fan H, Wu Y, Xie S, Girshick R. Momentum contrast for unsupervised visual
  representation learning. In \emph{Proceedings of the IEEE/CVF conference on
  computer vision and pattern recognition}, 2020, 9729--9738.

\bibitem{Lsun_yu2015lsun}
Yu F, Seff A, Zhang Y, Song S, Funkhouser T, Xiao J. Lsun: Construction of a
  large-scale image dataset using deep learning with humans in the loop.
  \emph{arXiv preprint arXiv:1506.03365}, 2015.

\bibitem{IDInvert_zhu2020domain}
Zhu J, Shen Y, Zhao D, Zhou B. In-domain gan inversion for real image editing.
  In \emph{European conference on computer vision}, 2020, 592--608.

\bibitem{GANEnsembling_chai2021ensembling}
Chai L, Zhu JY, Shechtman E, Isola P, Zhang R. Ensembling with deep generative
  views. In \emph{Proceedings of the IEEE/CVF Conference on Computer Vision and
  Pattern Recognition}, 2021, 14997--15007.

\bibitem{HyperEditor_zhang2024hypereditor}
Zhang H, Wu C, Cao G, Wang H, Cao W. HyperEditor: Achieving Both Authenticity
  and Cross-Domain Capability in Image Editing via Hypernetworks. In
  \emph{Proceedings of the AAAI Conference on Artificial Intelligence}, 2024,
  7051--7059.

\bibitem{yildirim2024warping}
Yildirim AB, Pehlivan H, Dundar A. Warping the residuals for image editing with
  stylegan. \emph{International Journal of Computer Vision}, 2024: 1--16.

\bibitem{wang2021geometry}
Wang B, Ponce CR. The geometry of deep generative image models and its
  applications. \emph{arXiv preprint arXiv:2101.06006}, 2021.

\bibitem{nguyen2024edit}
Nguyen T, Ojha U, Li Y, Liu H, Lee YJ. Edit One for All: Interactive Batch
  Image Editing. In \emph{Proceedings of the IEEE/CVF Conference on Computer
  Vision and Pattern Recognition}, 2024, 8271--8280.

\bibitem{clip2latent_pinkney2022clip2latent}
Pinkney JN, Li C. clip2latent: Text driven sampling of a pre-trained stylegan
  using denoising diffusion and clip. \emph{arXiv preprint arXiv:2210.02347},
  2022.

\bibitem{CLIPPAE_zhou2023clip}
Zhou C, Zhong F, {\"O}ztireli C. CLIP-PAE: projection-augmentation embedding to
  extract relevant features for a disentangled, interpretable and controllable
  text-guided face manipulation. In \emph{ACM SIGGRAPH 2023 Conference
  Proceedings}, 2023, 1--9.

\bibitem{Contraclip_tzelepis2022contraclip}
Tzelepis C, Oldfield J, Tzimiropoulos G, Patras I. Contraclip: Interpretable
  gan generation driven by pairs of contrasting sentences. \emph{arXiv preprint
  arXiv:2206.02104}, 2022.

\bibitem{CLIPInverter_baykal2023clip}
Baykal AC, Anees AB, Ceylan D, Erdem E, Erdem A, Yuret D. CLIP-guided StyleGAN
  Inversion for Text-driven Real Image Editing. \emph{ACM Transactions on
  Graphics}, 2023, 42(5): 1--18.

\bibitem{DDPMInversion_huberman2024edit}
Huberman-Spiegelglas I, Kulikov V, Michaeli T. An edit friendly ddpm noise
  space: Inversion and manipulations. In \emph{Proceedings of the IEEE/CVF
  Conference on Computer Vision and Pattern Recognition}, 2024, 12469--12478.

\bibitem{CFG_ho2022classifier}
Ho J, Salimans T. Classifier-free diffusion guidance. \emph{arXiv preprint
  arXiv:2207.12598}, 2022.

\bibitem{he2023data}
He X, Cao Z, Kolkin N, Yu L, Wan K, Rhodin H, Kalarot R. A data perspective on
  enhanced identity preservation for diffusion personalization. \emph{arXiv
  preprint arXiv:2311.04315}, 2023.

\bibitem{Imagereward_xu2024imagereward}
Xu J, Liu X, Wu Y, Tong Y, Li Q, Ding M, Tang J, Dong Y. Imagereward: Learning
  and evaluating human preferences for text-to-image generation. \emph{Advances
  in Neural Information Processing Systems}, 2024, 36.

\bibitem{Anydoor_chen2024anydoor}
Chen X, Huang L, Liu Y, Shen Y, Zhao D, Zhao H. Anydoor: Zero-shot object-level
  image customization. In \emph{Proceedings of the IEEE/CVF Conference on
  Computer Vision and Pattern Recognition}, 2024, 6593--6602.

\bibitem{TFace_huang2020curricularface}
Huang Y, Wang Y, Tai Y, Liu X, Shen P, Li S, Li J, Huang F. Curricularface:
  adaptive curriculum learning loss for deep face recognition. In
  \emph{proceedings of the IEEE/CVF conference on computer vision and pattern
  recognition}, 2020, 5901--5910.

\bibitem{SUTI_chen2024subject}
Chen W, Hu H, Li Y, Ruiz N, Jia X, Chang MW, Cohen WW. Subject-driven
  text-to-image generation via apprenticeship learning. \emph{Advances in
  Neural Information Processing Systems}, 2024, 36.

\bibitem{DreamCache_aiello2024dreamcache}
Aiello E, Michieli U, Valsesia D, Ozay M, Magli E. DreamCache: Finetuning-Free
  Lightweight Personalized Image Generation via Feature Caching. \emph{arXiv
  preprint arXiv:2411.17786}, 2024.

\bibitem{Hyperdreambooth_ruiz2024hyperdreambooth}
Ruiz N, Li Y, Jampani V, Wei W, Hou T, Pritch Y, Wadhwa N, Rubinstein M,
  Aberman K. Hyperdreambooth: Hypernetworks for fast personalization of
  text-to-image models. In \emph{Proceedings of the IEEE/CVF Conference on
  Computer Vision and Pattern Recognition}, 2024, 6527--6536.

\bibitem{DCI-ICO_jin2025customized}
Jin J, Shen Y, Fu Z, Yang J. Customized Generation Reimagined: Fidelity and
  Editability Harmonized. In \emph{European Conference on Computer Vision},
  2025, 410--426.

\bibitem{Dreamartist_dong2022dreamartist}
Dong Z, Wei P, Lin L. Dreamartist: Towards controllable one-shot text-to-image
  generation via positive-negative prompt-tuning. \emph{arXiv preprint
  arXiv:2211.11337}, 2022.

\bibitem{TextBoost_park2024textboost}
Park N, Kim K, Shim H. TextBoost: Towards One-Shot Personalization of
  Text-to-Image Models via Fine-tuning Text Encoder. \emph{arXiv preprint
  arXiv:2409.08248}, 2024.

\bibitem{P3S-Diffusion_hu2024p3s}
Hu J, Gao S, Hong L, Wang Q, Zhao Y, Wang Y, Zhang W. P3S-Diffusion: A
  Selective Subject-driven Generation Framework via Point Supervision.
  \emph{arXiv preprint arXiv:2412.19533}, 2024.

\bibitem{PersonalizedResiduals_ham2024personalized}
Ham C, Fisher M, Hays J, Kolkin N, Liu Y, Zhang R, Hinz T. Personalized
  Residuals for Concept-Driven Text-to-Image Generation. In \emph{Proceedings
  of the IEEE/CVF Conference on Computer Vision and Pattern Recognition}, 2024,
  8186--8195.

\bibitem{CompositionalInversion_zhang2024compositional}
Zhang X, Wei XY, Wu J, Zhang T, Zhang Z, Lei Z, Li Q. Compositional inversion
  for stable diffusion models. In \emph{Proceedings of the AAAI Conference on
  Artificial Intelligence}, 2024, 7350--7358.

\bibitem{kim2024learning}
Kim T, Chen W, Qiu Q. Learning to Customize Text-to-Image Diffusion In Diverse
  Context. \emph{arXiv preprint arXiv:2410.10058}, 2024.

\bibitem{UNIMOG_li2024unimo}
Li W, Xu X, Liu J, Xiao X. UNIMO-G: Unified Image Generation through Multimodal
  Conditional Diffusion. \emph{arXiv preprint arXiv:2401.13388}, 2024.

\bibitem{lambdaECLIPSE_patel2024lambda}
Patel M, Jung S, Baral C, Yang Y. $\lambda$-ECLIPSE: Multi-Concept Personalized
  Text-to-Image Diffusion Models by Leveraging CLIP Latent Space. \emph{arXiv
  preprint arXiv:2402.05195}, 2024.

\bibitem{IPAdapterInstruct_rowles2024ipadapter}
Rowles C, Vainer S, De~Nigris D, Elizarov S, Kutsy K, Donn{\'e} S.
  IPAdapter-Instruct: Resolving Ambiguity in Image-based Conditioning using
  Instruct Prompts. \emph{arXiv preprint arXiv:2408.03209}, 2024.

\bibitem{le2024diffusiongenerate}
Le DH, Pham T, Lee S, Clark C, Kembhavi A, Mandt S, Krishna R, Lu J. One
  Diffusion to Generate Them All, 2024.

\bibitem{EasyPortrait_kapitanov2023easyportrait}
Kapitanov A, Kvanchiani K, Kirillova S. EasyPortrait-Face Parsing and Portrait
  Segmentation Dataset, 2023.

\bibitem{PFLD_guo2019pfld}
Guo X, Li S, Yu J, Zhang J, Ma J, Ma L, Liu W, Ling H. PFLD: A practical facial
  landmark detector. \emph{arXiv preprint arXiv:1902.10859}, 2019.

\bibitem{SeFi-IDE_li2024sefi}
Li Y, Yang S, Wang W, Dong J. SeFi-IDE: Semantic-Fidelity Identity Embedding
  for Personalized Diffusion-Based Generation. \emph{arXiv preprint
  arXiv:2402.00631}, 2024.

\bibitem{Dense-Face_guo2024dense}
Guo X, Tran M, Cheng J, Liu X. Dense-Face: Personalized Face Generation Model
  via Dense Annotation Prediction. \emph{arXiv preprint arXiv:2412.18149},
  2024.

\bibitem{PersonaMagic_li2024personamagic}
Li X, Zhan J, He S, Xu Y, Dong J, Zhang H, Du Y. PersonaMagic: Stage-Regulated
  High-Fidelity Face Customization with Tandem Equilibrium. \emph{arXiv
  preprint arXiv:2412.15674}, 2024.

\bibitem{Omni-ID_qian2024omni}
Qian G, Wang KC, Patashnik O, Heravi N, Ostashev D, Tulyakov S, Cohen-Or D,
  Aberman K. Omni-ID: Holistic Identity Representation Designed for Generative
  Tasks. \emph{arXiv preprint arXiv:2412.09694}, 2024.

\bibitem{RealisID_sun2024realisid}
Sun Z, Du F, Chen W, Wang F, Chen Y, Rong Y, Xiong S. RealisID: Scale-Robust
  and Fine-Controllable Identity Customization via Local and Global
  Complementation. \emph{arXiv preprint arXiv:2412.16832}, 2024.

\bibitem{PersonaHOI_hu2025personahoi}
Hu X, Wang H, Lenssen JE, Schiele B. PersonaHOI: Effortlessly Improving
  Personalized Face with Human-Object Interaction Generation. \emph{arXiv
  preprint arXiv:2501.05823}, 2025.

\bibitem{UniPortrait_he2024uniportrait}
He J, Geng Y, Bo L. UniPortrait: A Unified Framework for Identity-Preserving
  Single-and Multi-Human Image Personalization. \emph{arXiv preprint
  arXiv:2408.05939}, 2024.

\bibitem{IC-Portrait_yang2025ic}
Yang H, Simsar E, Anagnostidi S, Zang Y, Hofmann T, Liu Z. IC-Portrait:
  In-Context Matching for View-Consistent Personalized Portrait. \emph{arXiv
  preprint arXiv:2501.17159}, 2025.

\bibitem{CharacterFactory_wang2024characterfactory}
Wang Q, Li B, Li X, Cao B, Ma L, Lu H, Jia X. CharacterFactory: Sampling
  Consistent Characters with GANs for Diffusion Models. \emph{arXiv preprint
  arXiv:2404.15677}, 2024.

\bibitem{Face0_valevski2023face0}
Valevski D, Lumen D, Matias Y, Leviathan Y. Face0: Instantaneously conditioning
  a text-to-image model on a face. In \emph{SIGGRAPH Asia 2023 Conference
  Papers}, 2023, 1--10.

\bibitem{Facestudio_yan2023facestudio}
Yan Y, Zhang C, Wang R, Zhou Y, Zhang G, Cheng P, Yu G, Fu B. Facestudio: Put
  your face everywhere in seconds. \emph{arXiv preprint arXiv:2312.02663},
  2023.

\bibitem{FACT_yu2024facechain}
Yu C, Xie H, Shang L, Liu Y, Dan J, Bo L, Sun B. FaceChain-FACT: Face Adapter
  with Decoupled Training for Identity-preserved Personalization. \emph{arXiv
  preprint arXiv:2410.12312}, 2024.

\bibitem{Magicapture_hyung2024magicapture}
Hyung J, Shin J, Choo J. Magicapture: High-resolution multi-concept portrait
  customization. In \emph{Proceedings of the AAAI Conference on Artificial
  Intelligence}, 2024, 2445--2453.

\bibitem{ID-Aligner_chen2024id}
Chen W, Zhang J, Wu J, Wu H, Xiao X, Lin L. ID-Aligner: Enhancing
  Identity-Preserving Text-to-Image Generation with Reward Feedback Learning.
  \emph{arXiv preprint arXiv:2404.15449}, 2024.

\bibitem{IDAdapter_cui2024idadapter}
Cui S, Guo J, An X, Deng J, Zhao Y, Wei X, Feng Z. IDAdapter: Learning Mixed
  Features for Tuning-Free Personalization of Text-to-Image Models. In
  \emph{Proceedings of the IEEE/CVF Conference on Computer Vision and Pattern
  Recognition}, 2024, 950--959.

\bibitem{LCM_luo2023latent}
Luo S, Tan Y, Huang L, Li J, Zhao H. Latent consistency models: Synthesizing
  high-resolution images with few-step inference. \emph{arXiv preprint
  arXiv:2310.04378}, 2023.

\bibitem{Face-diffuser_wang2024high}
Wang Y, Zhang W, Zheng J, Jin C. High-fidelity Person-centric Subject-to-Image
  Synthesis. In \emph{Proceedings of the IEEE/CVF Conference on Computer Vision
  and Pattern Recognition}, 2024, 7675--7684.

\bibitem{FreeCure_cai2024foundation}
Cai Y, Jiang Z, Liu Y, Jiang C, Xue W, Luo W, Guo Y. Foundation Cures
  Personalization: Recovering Facial Personalized Models' Prompt Consistency.
  \emph{arXiv preprint arXiv:2411.15277}, 2024.

\bibitem{identity-expression_liu2024towards}
Liu R, Ma B, Zhang W, Hu Z, Fan C, Lv T, Ding Y, Cheng X. Towards a
  simultaneous and granular identity-expression control in personalized face
  generation. In \emph{Proceedings of the IEEE/CVF Conference on Computer
  Vision and Pattern Recognition}, 2024, 2114--2123.

\bibitem{EmojiDiff_jiang2024emojidiff}
Jiang L, Li R, Zhang Z, Fang S, Ma C. EmojiDiff: Advanced Facial Expression
  Control with High Identity Preservation in Portrait Generation. \emph{arXiv
  preprint arXiv:2412.01254}, 2024.

\bibitem{MagicID_deng2024magicid}
Deng Z, Liu W, Wang F, Zhang J, Chen F, Zhang M, Zhang W, Mi Z. MagicID:
  Flexible ID Fidelity Generation System. \emph{arXiv preprint
  arXiv:2408.09248}, 2024.

\bibitem{Diff-PC_xu2024diff}
Xu Y, Zhai B, Zhang C, Li M, Li Y, Du S. Diff-PC: Identity-preserving and
  3D-aware controllable diffusion for zero-shot portrait customization.
  \emph{Information Fusion}, 2024: 102869.

\bibitem{UniHuman_li2024unihuman}
Li N, Liu Q, Singh KK, Wang Y, Zhang J, Plummer BA, Lin Z. UniHuman: A Unified
  Model For Editing Human Images in the Wild. In \emph{Proceedings of the
  IEEE/CVF Conference on Computer Vision and Pattern Recognition}, 2024,
  2039--2048.

\bibitem{huang2024parts}
Huang Z, Fan H, Wang L, Sheng L. From Parts to Whole: A Unified Reference
  Framework for Controllable Human Image Generation. \emph{arXiv preprint
  arXiv:2404.15267}, 2024.

\bibitem{Prospect_zhang2023prospect}
Zhang Y, Dong W, Tang F, Huang N, Huang H, Ma C, Lee TY, Deussen O, Xu C.
  Prospect: Prompt spectrum for attribute-aware personalization of diffusion
  models. \emph{ACM Transactions on Graphics (TOG)}, 2023, 42(6): 1--14.

\bibitem{Style-friendly_choi2024style}
Choi J, Shin C, Oh Y, Kim H, Yoon S. Style-Friendly SNR Sampler for
  Style-Driven Generation. \emph{arXiv preprint arXiv:2411.14793}, 2024.

\bibitem{inspirationtree_vinker2023concept}
Vinker Y, Voynov A, Cohen-Or D, Shamir A. Concept decomposition for visual
  exploration and inspiration. \emph{ACM Transactions on Graphics (TOG)}, 2023,
  42(6): 1--13.

\bibitem{MATTE_agarwal2023image}
Agarwal A, Karanam S, Shukla T, Srinivasan BV. An image is worth multiple
  words: Multi-attribute inversion for constrained text-to-image synthesis.
  \emph{arXiv preprint arXiv:2311.11919}, 2023.

\bibitem{break-for-make_xu2024break}
Xu Y, Tang F, Cao J, Zhang Y, Deussen O, Dong W, Li J, Lee TY. Break-for-make:
  Modular low-rank adaptations for composable content-style customization.
  \emph{arXiv preprint arXiv:2403.19456}, 2024.

\bibitem{zhuoqi2024content}
Zhuoqi M, Yixuan Z, Zejun Y, Long T, Xiyang L. Content-style disentangled
  representation for controllable artistic image stylization and generation.
  \emph{arXiv preprint arXiv:2412.14496}, 2024.

\bibitem{BLoRA_frenkel2025implicit}
Frenkel Y, Vinker Y, Shamir A, Cohen-Or D. Implicit style-content separation
  using b-lora. In \emph{European Conference on Computer Vision}, 2025,
  181--198.

\bibitem{Aesthetic-Gradients_gallego2022personalizing}
Gallego V. Personalizing text-to-image generation via aesthetic gradients.
  \emph{arXiv preprint arXiv:2209.12330}, 2022.

\bibitem{Stylealigned_hertz2024style}
Hertz A, Voynov A, Fruchter S, Cohen-Or D. Style aligned image generation via
  shared attention. In \emph{Proceedings of the IEEE/CVF Conference on Computer
  Vision and Pattern Recognition}, 2024, 4775--4785.

\bibitem{SAG_pan2023towards}
Pan J, Yan H, Liew JH, Feng J, Tan VY. Towards accurate guided diffusion
  sampling through symplectic adjoint method. \emph{arXiv preprint
  arXiv:2312.12030}, 2023.

\bibitem{FreeTuner_xu2024freetuner}
Xu Y, Wang Z, Xiao J, Liu W, Chen L. FreeTuner: Any Subject in Any Style with
  Training-free Diffusion. \emph{arXiv preprint arXiv:2405.14201}, 2024.

\bibitem{RB-Modulation_rout2024rb}
Rout L, Chen Y, Ruiz N, Kumar A, Caramanis C, Shakkottai S, Chu WS.
  RB-Modulation: Training-Free Personalization of Diffusion Models using
  Stochastic Optimal Control. \emph{arXiv preprint arXiv:2405.17401}, 2024.

\bibitem{Diptych-Prompting_shin2024large}
Shin C, Choi J, Kim H, Yoon S. Large-Scale Text-to-Image Model with Inpainting
  is a Zero-Shot Subject-Driven Image Generator. \emph{arXiv preprint
  arXiv:2411.15466}, 2024.

\bibitem{Mix-of-show_gu2024mix}
Gu Y, Wang X, Wu JZ, Shi Y, Chen Y, Fan Z, Xiao W, Zhao R, Chang S, Wu W,
  et~al.. Mix-of-show: Decentralized low-rank adaptation for multi-concept
  customization of diffusion models. \emph{Advances in Neural Information
  Processing Systems}, 2024, 36.

\bibitem{Ortha_po2024orthogonal}
Po R, Yang G, Aberman K, Wetzstein G. Orthogonal adaptation for modular
  customization of diffusion models. In \emph{Proceedings of the IEEE/CVF
  Conference on Computer Vision and Pattern Recognition}, 2024, 7964--7973.

\bibitem{LoRA.rar_shenaj2024lora}
Shenaj D, Bohdal O, Ozay M, Zanuttigh P, Michieli U. LoRA. rar: Learning to
  Merge LoRAs via Hypernetworks for Subject-Style Conditioned Image Generation.
  \emph{arXiv preprint arXiv:2412.05148}, 2024.

\bibitem{Block-wise-LoRA_li2024block}
Li L, Zeng H, Yang C, Jia H, Xu D. Block-wise LoRA: Revisiting Fine-grained
  LoRA for Effective Personalization and Stylization in Text-to-Image
  Generation. \emph{arXiv preprint arXiv:2403.07500}, 2024.

\bibitem{MuDI_jang2024identity}
Jang S, Jo J, Lee K, Hwang SJ. Identity Decoupling for Multi-Subject
  Personalization of Text-to-Image Models. \emph{arXiv preprint
  arXiv:2404.04243}, 2024.

\bibitem{IR-Diffusion_he2024improving}
He H, Wang Q, Zhou Y, Cai Y, Chao H, Yin J, Yang H. Improving Multi-Subject
  Consistency in Open-Domain Image Generation with Isolation and Reposition
  Attention. \emph{arXiv preprint arXiv:2411.19261}, 2024.

\bibitem{DisenDiff_zhang2024attention}
Zhang Y, Yang M, Zhou Q, Wang Z. Attention Calibration for Disentangled
  Text-to-Image Personalization. In \emph{Proceedings of the IEEE/CVF
  Conference on Computer Vision and Pattern Recognition}, 2024, 4764--4774.

\bibitem{LoRAComposer_yang2024lora}
Yang Y, Wang W, Peng L, Song C, Chen Y, Li H, Yang X, Lu Q, Cai D, Wu B,
  et~al.. LoRA-Composer: Leveraging Low-Rank Adaptation for Multi-Concept
  Customization in Training-Free Diffusion Models. \emph{arXiv preprint
  arXiv:2403.11627}, 2024.

\bibitem{jain2024multi}
Jain A, Paliwal S, Sharma M, Jamwal V, Vig L. Multi-Subject Personalization.
  \emph{arXiv preprint arXiv:2405.12742}, 2024.

\bibitem{Autostory_wang2023autostory}
Wang W, Zhao C, Chen H, Chen Z, Zheng K, Shen C. Autostory: Generating diverse
  storytelling images with minimal human effort. \emph{arXiv preprint
  arXiv:2311.11243}, 2023.

\bibitem{parmar2025object}
Parmar G, Patashnik O, Wang KC, Ostashev D, Narasimhan S, Zhu JY, Cohen-Or D,
  Aberman K. Object-level Visual Prompts for Compositional Image Generation.
  \emph{arXiv preprint arXiv:2501.01424}, 2025.

\bibitem{GANTASTIC_dalva2024gantastic}
Dalva Y, Yesiltepe H, Yanardag P. GANTASTIC: GAN-based Transfer of
  Interpretable Directions for Disentangled Image Editing in Text-to-Image
  Diffusion Models. \emph{arXiv preprint arXiv:2403.19645}, 2024.

\bibitem{ARLDM_pan2024synthesizing}
Pan X, Qin P, Li Y, Xue H, Chen W. Synthesizing coherent story with
  auto-regressive latent diffusion models. In \emph{Proceedings of the IEEE/CVF
  Winter Conference on Applications of Computer Vision}, 2024, 2920--2930.

\bibitem{MCPL_jin2023image}
Jin C, Tanno R, Saseendran A, Diethe T, Teare P. An image is worth multiple
  words: Learning object level concepts using multi-concept prompt learning.
  \emph{arXiv preprint arXiv:2310.12274}, 2023.

\bibitem{lu2024object}
Lu J, Xie C, Guo H. Object-Driven One-Shot Fine-tuning of Text-to-Image
  Diffusion with Prototypical Embedding. \emph{arXiv preprint
  arXiv:2401.15708}, 2024.

\bibitem{ryu2024memory}
Ryu H, Lim S, Shim H. Memory-Efficient Personalization using Quantized
  Diffusion Model. \emph{arXiv preprint arXiv:2401.04339}, 2024.

\bibitem{CLIF_lin2024non}
Lin W, Chen J, Shi J, Zhu Y, Liang C, Miao J, Jin T, Zhao Z, Wu F, Yan S,
  et~al.. Non-confusing Generation of Customized Concepts in Diffusion Models.
  \emph{arXiv preprint arXiv:2405.06914}, 2024.

\bibitem{BRAT_baker2024brat}
Baker J. BRAT: Bonus oRthogonAl Token for Architecture Agnostic Textual
  Inversion. \emph{arXiv preprint arXiv:2408.04785}, 2024.

\bibitem{ArtiFade_yang2024artifade}
Yang S, Hao S, Cao Y, Wong KYK. ArtiFade: Learning to Generate High-quality
  Subject from Blemished Images. \emph{arXiv preprint arXiv:2409.03745}, 2024.

\bibitem{Cusconcept_xu2024cusconcept}
Xu Z, Hao S, Han K. Cusconcept: Customized visual concept decomposition with
  diffusion models. \emph{arXiv preprint arXiv:2410.00398}, 2024.

\bibitem{Customnet_yuan2023customnet}
Yuan Z, Cao M, Wang X, Qi Z, Yuan C, Shan Y. Customnet: Zero-shot object
  customization with variable-viewpoints in text-to-image diffusion models.
  \emph{arXiv preprint arXiv:2310.19784}, 2023.

\bibitem{LARGEN_pan2024locate}
Pan Y, Mao C, Jiang Z, Han Z, Zhang J. Locate, Assign, Refine: Taming
  Customized Image Inpainting with Text-Subject Guidance. \emph{arXiv preprint
  arXiv:2403.19534}, 2024.

\bibitem{RealCustom_huang2024realcustom}
Huang M, Mao Z, Liu M, He Q, Zhang Y. RealCustom: Narrowing Real Text Word for
  Real-Time Open-Domain Text-to-Image Customization. In \emph{Proceedings of
  the IEEE/CVF Conference on Computer Vision and Pattern Recognition}, 2024,
  7476--7485.

\bibitem{Moa_wang2024moa}
Wang KC, Ostashev D, Fang Y, Tulyakov S, Aberman K. Moa: Mixture-of-attention
  for subject-context disentanglement in personalized image generation. In
  \emph{SIGGRAPH Asia 2024 Conference Papers}, 2024, 1--12.

\bibitem{RealCustom++_mao2024realcustom++}
Mao Z, Huang M, Ding F, Liu M, He Q, Chang X, Zhang Y. RealCustom++:
  Representing Images as Real-Word for Real-Time Customization. \emph{arXiv
  preprint arXiv:2408.09744}, 2024.

\bibitem{EZIGen_duan2024ezigen}
Duan Z, Ding Y, Gou C, Zhou Z, Smith E, Liu L. EZIGen: Enhancing zero-shot
  subject-driven image generation with precise subject encoding and decoupled
  guidance. \emph{arXiv preprint arXiv:2409.08091}, 2024.

\bibitem{Fusion_mohamed2024fusion}
Mohamed S, Han D, Li Y. Fusion is all you need: Face Fusion for Customized
  Identity-Preserving Image Synthesis. \emph{arXiv preprint arXiv:2409.19111},
  2024.

\bibitem{ID-Patch_zhang2024id}
Zhang Y, Zhi T, Liu J, Sang S, Jiang L, Yan Q, Liu S, Luo L. ID-Patch: Robust
  ID Association for Group Photo Personalization. \emph{arXiv preprint
  arXiv:2411.13632}, 2024.

\bibitem{PersonaCraft_kim2024personacraft}
Kim G, Jeon SY, Lee S, Chun SY. PersonaCraft: Personalized Full-Body Image
  Synthesis for Multiple Identities from Single References Using
  3D-Model-Conditioned Diffusion. \emph{arXiv preprint arXiv:2411.18068}, 2024.

\bibitem{MagicNaming_zhao2024magicnaming}
Zhao J, Zheng H, Wang C, Lan L, Hunag W, Tang Y. MagicNaming: Consistent
  Identity Generation by Finding a" Name Space" in T2I Diffusion Models.
  \emph{arXiv preprint arXiv:2412.14902}, 2024.

\bibitem{StoryWeaver_zhang2024storyweaver}
Zhang J, Tang J, Zhang R, Lv T, Sun X. StoryWeaver: A Unified World Model for
  Knowledge-Enhanced Story Character Customization. \emph{arXiv preprint
  arXiv:2412.07375}, 2024.

\bibitem{peng2024dreambench++}
Peng Y, Cui Y, Tang H, Qi Z, Dong R, Bai J, Han C, Ge Z, Zhang X, Xia ST.
  Dreambench++: A human-aligned benchmark for personalized image generation.
  \emph{arXiv preprint arXiv:2406.16855}, 2024.

\bibitem{FID_heusel2017gans}
Heusel M, Ramsauer H, Unterthiner T, Nessler B, Hochreiter S. Gans trained by a
  two time-scale update rule converge to a local nash equilibrium.
  \emph{Advances in neural information processing systems}, 2017, 30.

\bibitem{Inception_szegedy2016rethinking}
Szegedy C, Vanhoucke V, Ioffe S, Shlens J, Wojna Z. Rethinking the inception
  architecture for computer vision. In \emph{Proceedings of the IEEE conference
  on computer vision and pattern recognition}, 2016, 2818--2826.

\bibitem{Dinov2_oquab2023dinov2}
Oquab M, Darcet T, Moutakanni T, Vo H, Szafraniec M, Khalidov V, Fernandez P,
  Haziza D, Massa F, El-Nouby A, et~al.. Dinov2: Learning robust visual
  features without supervision. \emph{arXiv preprint arXiv:2304.07193}, 2023.

\bibitem{CustomVideoX_she2025customvideox}
She D, Liu M, Pang J, Wang J, Yang Z, He W, Zhang G, Wang Y, Huang Q, Tang H,
  et~al.. CustomVideoX: 3D Reference Attention Driven Dynamic Adaptation for
  Zero-Shot Customized Video Diffusion Transformers. \emph{arXiv preprint
  arXiv:2502.06527}, 2025.

\bibitem{Pia_zhang2024pia}
Zhang Y, Xing Z, Zeng Y, Fang Y, Chen K. Pia: Your personalized image animator
  via plug-and-play modules in text-to-image models. In \emph{Proceedings of
  the IEEE/CVF Conference on Computer Vision and Pattern Recognition}, 2024,
  7747--7756.

\bibitem{MovieWeaver_liang2025movie}
Liang F, Ma H, He Z, Hou T, Hou J, Li K, Dai X, Juefei-Xu F, Azadi S, Sinha A,
  et~al.. Movie Weaver: Tuning-Free Multi-Concept Video Personalization with
  Anchored Prompts. \emph{arXiv preprint arXiv:2502.07802}, 2025.

\bibitem{Motrans_li2024motrans}
Li X, Jia X, Wang Q, Diao H, Ge M, Li P, He Y, Lu H. Motrans: Customized motion
  transfer with text-driven video diffusion models. In \emph{Proceedings of the
  32nd ACM International Conference on Multimedia}, 2024, 3421--3430.

\bibitem{PersonalVideo_li2024personalvideo}
Li H, Qiu H, Zhang S, Wang X, Wei Y, Li Z, Zhang Y, Wu B, Cai D. PersonalVideo:
  High ID-Fidelity Video Customization without Dynamic and Semantic
  Degradation. \emph{arXiv preprint arXiv:2411.17048}, 2024.

\bibitem{Customvideo_wang2024customvideo}
Wang Z, Li A, Zhu L, Guo Y, Dou Q, Li Z. Customvideo: Customizing text-to-video
  generation with multiple subjects. \emph{arXiv preprint arXiv:2401.09962},
  2024.

\bibitem{Videobooth_jiang2024videobooth}
Jiang Y, Wu T, Yang S, Si C, Lin D, Qiao Y, Loy CC, Liu Z. Videobooth:
  Diffusion-based video generation with image prompts. In \emph{Proceedings of
  the IEEE/CVF Conference on Computer Vision and Pattern Recognition}, 2024,
  6689--6700.

\bibitem{Dreambooth3d_raj2023dreambooth3d}
Raj A, Kaza S, Poole B, Niemeyer M, Ruiz N, Mildenhall B, Zada S, Aberman K,
  Rubinstein M, Barron J, et~al.. Dreambooth3d: Subject-driven text-to-3d
  generation. In \emph{Proceedings of the IEEE/CVF international conference on
  computer vision}, 2023, 2349--2359.

\bibitem{ouyang2023chasing}
Ouyang Y, Chai W, Ye J, Tao D, Zhan Y, Wang G. Chasing consistency in
  text-to-3d generation from a single image. \emph{arXiv preprint
  arXiv:2309.03599}, 2023.

\end{thebibliography}

\end{document}